\documentclass[journal]{IEEEtran}
\usepackage{graphicx}
\usepackage{amsmath}
\usepackage{amsmath,amssymb,amsfonts}
\usepackage{array}
\usepackage{booktabs}
\usepackage{geometry}
\usepackage{adjustbox}
\usepackage{longtable}
\usepackage{tikz}
\usepackage{hyperref}
\usepackage{algorithm}
\usepackage{algpseudocode}
\usepackage{comment}
\usepackage{caption}

\begin{document}

\title{Comprehensive Survey of Reinforcement Learning: From Algorithms to Practical Challenges}

\author{$\text{Majid Ghasemi}^{\dagger, 1}$, $\text{Amir Hossein Moosavi}^{1}$, 
and $\text{Dariush~Ebrahimi}^{1}$
\thanks{
$\dagger$Corresponding author\\ 
$^1$Department of Computer Science, Wilfrid Laurier University, Waterloo, Canada. Emails: \texttt{\{mghasemi, debrahimi\}@wlu.ca} \& \texttt{s.amirhosein.mn@gmail.com}
}
}

\maketitle

\begin{abstract}

Reinforcement Learning (RL) has emerged as a powerful paradigm in Artificial Intelligence (AI), enabling agents to learn optimal behaviors through interactions with their environments. Drawing from the foundations of trial and error, RL equips agents to make informed decisions through feedback in the form of rewards or penalties. This paper presents a comprehensive survey of RL, meticulously analyzing a wide range of algorithms, from foundational tabular methods to advanced Deep Reinforcement Learning (DRL) techniques. We categorize and evaluate these algorithms based on key criteria such as scalability, sample efficiency, and suitability.
We compare the methods in the form of their strengths and weaknesses in diverse settings.
Additionally, we offer practical insights into the selection and implementation of RL algorithms, addressing common challenges like convergence, stability, and the exploration-exploitation dilemma. This paper serves as a comprehensive reference for researchers and practitioners aiming to harness the full potential of RL in solving complex, real-world problems.

\end{abstract}


\begin{table*}[htbp]
\renewcommand{\arraystretch}{1.2} 
\centering
\caption{A list of notations \& symbols used in this survey}
\begin{tabular}{|>{\raggedright\arraybackslash}p{3cm}|>{\raggedright\arraybackslash}p{9.9cm}|}
\hline
\textbf{Symbols \& Notations} & \textbf{Definition} \\
\hline
\(S\) & A finite set of states \\
\hline
\(s\) & State \\
\hline
\(s'\) & Next State \\
\hline
\(A\) & A finite set of actions \\
\hline
\(a\) & Action \\
\hline
\(P\) & State transition probability function \\ 
\hline
\(P(s'|s, a)\) & Probability of transitioning to \(s'\) by taking action \(a\) \\
\hline
\(R\) & Reward function \\
\hline
\(R(s, a)\) & The immediate reward received after taking action \(a\) in state \(s\) \\   
\hline
\(\gamma\) & Discount Factor \\
\hline
\(\pi\) & (Target) Policy \\
\hline
\(\beta\) & Behaviour policy \\
\hline
\(\pi^*\) & Optimal Policy \\
\hline
\(\pi_\theta(a|s)\) & The probability of taking action \(a\) in the state \(s\) under policy \(\pi\) parameterized by \(\theta\) \\
\hline
\(G_t\) & Expected cumulative reward \\
\hline
\(V_{\pi}(s)\) & State-value function \\
\hline
\(Q_{\pi}(s, a)\) & Action-value function \\
\hline
\(\alpha\) & Learning Rate \\
\hline
\(\epsilon\) & Exploration rate \\
\hline
\(\delta\) & TD error \\
\hline
\(e(s)\) & Eligibility trace \\
\hline
\(q_*\) & Optimal action-value function \\
\hline
\(\theta^*\) & Optimal parameters \\
\hline
\(\theta_t\) & Weight vector \\
\hline
\(\theta_t^T\) & Transpose of the weight vector \\
\hline
\(\phi_t\) & Feature vector obtained through current state \(S_t\) \\
\hline
\(Q_{\pi_{\theta}}(s, a)\) & Action-value function under the current policy \\
\hline
\(J(\pi)\) & The expected discounted reward for a policy \(\pi\) \\
\hline
\(D_{KL}(\pi_{\theta}| \pi_{\theta'})\) & Kullback-Leibler (KL) divergence between the new policy \(\pi_{\theta'}\) and the old policy \(\pi_{\theta}\) \\
\hline
\(L(\theta)\) & Surrogate objective function \\
\hline
\(\hat{A}_{\theta_{\text{old}}}(s, a)\) & An estimate of the advantage function \\
\hline
\(A(s, a)\) & Advantage function \\
\hline
\(\mu_\theta\) & Deterministic policy \\
\hline
\(Q^\mu(s, a)\) & Action-value function under \(\mu_\theta\) \\
\hline
\(\rho^\mu\) & Discounted state visitation distribution under \(\mu_\theta\) \\
\hline
\(\tau < 1\) & Target update rate \\
\hline
\(Q_{\theta}\) & Critic network \\
\hline
\(Q_{\theta_i'}\) & Target Critic network \\
\hline
\(\mu_{\theta}\) & Actor network \\
\hline
\(\mu_{\theta'}\) & Target actor network \\
\hline
\end{tabular}
\label{tab:notations}
\end{table*}

\section{Introduction}

Reinforcement Learning (RL) is a subfield of Artificial Intelligence (AI) in which an agent learns to make decisions by interacting with an environment, aiming to maximize cumulative reward over time \cite{sutton2018reinforcement}. RL has rapidly evolved since the 1950s, when Richard Bellman's work on Dynamic Programming (DP) established foundational concepts that underpin current approaches \cite{bellman1954theory, ghasemi2024introduction}. The field gradually became more widespread by proposing more advanced approaches, such as Temporal-Difference (TD) Learning \cite{sutton1988learning, watkins1989learning} and suggesting solutions to exploration-exploitation dilemma \cite{kaelbling1996reinforcement, auer2002finite}. RL has further been evolving rapidly due to its integration with Deep Learning (DL), giving rise to Deep Reinforcement Learning (DRL). This advancement enables researchers to tackle more sophisticated and complex problems \cite{mnih2015human, silver2016mastering}. It has proven to be highly effective in solving sequential decision-making problems in a variety of fields, such as game playing (\cite{souchleris2023reinforcement, koyamada2024pgx, xu2023language, qu2023pursuit}), robotics (\cite{rana2023residual, kober2013reinforcement, balasubramanian2023intrinsically, abeyruwan2023sim2real}), and autonomous systems, particularly in Intelligent Transportation Systems (ITS) (\cite{zhu2023multi, yazdani2023intelligent, liu2023swarm, chen2023deep, ghosh2024maximizing}).


This survey examines the practical application of different RL approaches through various domains including but not limited to: robotics \cite{rana2023residual, aghaei2022real, yu2023obstacle}, optimization \cite{ de2005adaptive, li2023double}, energy efficiency and power management \cite{liu2018intelligent, jaiswal2020green, xuan2021power}, networks \cite{lassila2001efficient, oh2020reinforcement, kwon2022learning}, dynamic and partially observable environments \cite{de2018per, li2023motion, sutton2004temporal, zuters2010realizing}, video games \cite{moerland2018monte}, real-time systems and hardware implementations \cite{nasreen2022overview, lopes2018intelligent}, financial portfolios \cite{darapaneni2020automated}, ITS \cite{zhu2023multi, desai2017prioritized, santoso2020multiagent}, signal processing \cite{saha2002maximum}, benchmark tasks \cite{wu2023bias}, data management and processing \cite{yao2022improving, zhang2021image}, multi-agent and cloud-based systems \cite{suh2021sarsa, go2016reinforcement, zerbel2019multiagent, mangalampalli2024multi}. 
Moreover, our survey gives detailed explanations of RL algorithms, ranging from traditional tabular methods to state-of-the-art methods. 

\emph{Related Surveys}:
Several notable survey papers have examined different aspects of RL or attempted to teach various concepts to readers.
Foundational work, such as \cite{kaelbling1996reinforcement}, offers an in-depth computer science perspective on RL, encompassing both its historical roots and modern developments. In the realm of DRL, \cite{li2017deep, arulkumaran2017deep, wang2022deep, wang2020deep} offer comprehensive analyses of DRL's integration with DL, highlighting its applications and theoretical advancements.
Model-based RL is identified as a promising area in RL, with authors in \cite{moerland2023model, polydoros2017survey, luo2024survey} emphasizing its potential to enhance sample efficiency by simulating trial-and-error learning in predictive models, reducing the need for costly real-world interactions.

In \cite{sato2019model}, authors provided a survey of Model-free RL specifically within the financial portfolio management domain, exploring its applications and effectiveness. Meanwhile, \cite{ramirez2022model} analyzes the combination of Model-free RL and Imitation Learning (IL), a hybrid approach known as RL from Expert Demonstrations (RLED), which leverages expert data to enhance RL performance. These survey papers help to understand the various aspects of RL and its integration with other areas.

To our knowledge, no papers have analyzed the strengths and weaknesses of algorithms used in papers and provide a comprehensive analysis of an entire paper. The first motivation of this survey is to address this gap.

RL involves various challenges in choosing appropriate algorithms due to diverse factors, such as problem characteristics and environmental dynamics. In general, it depends on numerous factors based on the characteristics of the problem, including whether the state and action spaces are large, whether their values are discrete or continuous, or if the dynamics of the environment are stochastic or deterministic. Data availability and sample efficiency are other factors to consider. It is also important to consider the degree to which direct implementation can be achieved and how easily it can be debugged. In these respects, simpler algorithms may tend to be more user-friendly but cannot be applied to complex problems. Convergence and stability are important considerations, as certain algorithms provide better guarantees in specific circumstances. In conclusion, the decision-making process is influenced by exploration style, domain-specific requirements, and past research results. To determine which algorithm to use based on comparing the problem they are solving with similar existing research, researchers should have a comprehensive paper examining several papers in different domains thoroughly and accurately. Besides saving time and resources, this will also prevent the excess cost of going through a trial-and-error process to determine which solution to choose, which is the second motivation behind conducting this survey.

The rest of the paper is organized as follows: In \ref{sec:RL}, we provide a general overview of RL before diving into algorithms. Consequently, we undertake an examination of various algorithms within the domain of RL, inclusive of the associated papers, as well as an analysis of their respective merits and drawbacks. It is imperative to acknowledge that these algorithms fall into three overarching categories: Value-based Methods, Policy-based Methods, and Actor-Critic Methods. Section \ref{sec:VBM} initiates the discussion by focusing on Value-based Methods, delineated by its four core components: Dynamic Programming, Tabular Model-free, Approximation Model-free, and Tabular Model-based Methods. Section \ref{sec:PBM} subsequently talks about the Policy-based Methods. Furthermore, section \ref{sec:ACM} offers detailed insights into Actor-Critic Methods. In section \ref{sec:Final_Notes}, we give a summary of the paper and discuss the scope of it. Finally, section \ref{sec:conclusion} provides a synthesis of the paper through a review of key points and an exposition of future research directions.

\section{General Overview of RL}\label{sec:RL}
In this section, we give a general overview of RL, assuming readers have a basic knowledge of RL.
In the framework of RL, the learning process is defined by several key components. The fundamental concept of RL is to capture the crucial elements of a real problem faced by an \textbf{agent} that interacts with its \textbf{environment} to achieve a goal. It is evident that such an agent must be capable of sensing the state of the environment to some extent and must have the ability to take actions that influence the state \cite{sutton2018reinforcement}. In general, an \textbf{action} refers to any decision an agent will have to make, while a \textbf{state} refers to any factor that the agent will have to consider when making that decision. 

Beyond the agent and the environment, an RL system has four main sub-elements: a policy, a reward signal, a value function, and, optionally, a model of the environment.
The \textbf{reward signal} determines the agent’s behavior in the given environment. During each time step, the environment sends a single number, a reward, to the agent. Ultimately, the agent's sole objective is to maximize the total reward it receives. A \textbf{policy} defines how a learning agent should behave at a particular point in time. In simple terms, the agent's policy is the mapping from a possible state to a potential action. The \textbf{value function} corresponds to the agent's current mapping from the set of possible states to its estimates of the net long-term reward it can expect once it visits a state (or state-action pair) and continues to follow its current policy. As a final point, the \textbf{model of the environment} simulates the behavior of the environment or, more generally, can be used to infer how the environment will behave \cite{sutton2018reinforcement, eschmann2021reward, sutton1999reinforcement}.

There are two broad categories of RL methodology: Model-free and Model-based. Model-free methods do not assume knowledge of the environment’s dynamics and learn directly from interactions with the environment. On the other hand, Model-based methods involve building a model of the environment's dynamics and using this model to develop and improve policies \cite{kaelbling1996reinforcement, ghasemi2024introduction}. Each of the mentioned categories has its own advantages and disadvantages, that will be discussed later in the paper.


Following this, we will briefly examine two of the most crucial components of RL. First, we will explore the Markov Decision Process (MDP), the foundational framework that structures the learning environment and guides the agent's decision-making process. Then, we will discuss the exploration-exploitation dilemma, one of the most imperative characteristics of RL, which balances the need to gather new information with the goal of maximizing rewards.

\subsection{Markov Decision Process (MDP)}

The MDP is a sequential decision-making process in which the costs and transition functions are directly related to the current state and actions of the system. MDPs aim to provide the decision-maker with an optimal policy $\pi: S \rightarrow A$. The models have been applied to a wide range of subjects, including queueing, inventory control, and recommender systems \cite{PUTERMAN1990331, zanini2014markov, wei2017reinforcement}.

MDP is defined by a tuple $(S, A, P, R, \gamma)$, where:
\begin{itemize}
    \item $S$ is a finite set of states,
    \item $A$ is a finite set of actions,
    \item $P: S \times A \times S \rightarrow [0, 1]$ is the state transition probability function, where $P(s'|s, a)$ denotes the probability of transitioning to state $s'$ from state $s$ by taking action $a$,
    \item $R: S \times A \rightarrow \mathbb{R}$ is the reward function, where $R(s, a)$ denotes the immediate reward received after taking action $a$ in state $s$,
    \item $\gamma \in [0, 1]$ is the discount factor that determines the importance of future rewards.
\end{itemize}

The agent's behavior is defined by a policy $\pi: S \rightarrow A$, which maps states to actions. The goal of the agent is to find an optimal policy $\pi^*$ that maximizes the expected cumulative reward, often termed the return ($G_t$), which is defined as:
\begin{equation}
G_t = \sum_{k=0}^{\infty} \gamma^k R(s_{t+k}, a_{t+k})
\end{equation}

Central to solving an MDP are value functions, which estimate the expected return. The state-value function $V_{\pi}(s)$ under policy $\pi$ is the expected return starting from state $s$ and following policy $\pi$ thereafter \cite{ghasemi2024introduction, foster2023foundations, PUTERMAN1990331}:
\begin{equation}
\label{Eq_BE1}
V_{\pi}(s) = \mathbb{E}_{\pi} [G_t | s_t = s]
\end{equation}
Similarly, the action-value function $Q_{\pi}(s, a)$ is the expected return starting from state $s$, taking action $a$, and thereafter following policy $\pi$:
\begin{equation}
\label{Eq_BE2}
Q_{\pi}(s, a) = \mathbb{E}_{\pi} [G_t | s_t = s, a_t = a]
\end{equation}

Equations \ref{Eq_BE1} and \ref{Eq_BE2} are referred to Bellman Equations. To find the optimal policy $\pi^*$, RL algorithms iteratively update value functions based on experience \cite{watkins1989learning, rummery1994line}. In Q-learning (will be explained later in Alg.~\ref{alg:Q_learning}), the update rule for the action-value function is:
\begin{equation}
\begin{split}  
&Q(s_t, a_t) \leftarrow Q(s_t, a_t) + \alpha \bigg( R(s_t, a_t) + \\
&\gamma \max_{a'} Q(s_{t+1}, a') - Q(s_t, a_t) \bigg)
\end{split}
\end{equation}

where $\alpha$ is the learning rate.
For detailed explanations of MDPs and RL in general, readers are referred to \cite{sutton2018reinforcement, kaelbling1996reinforcement, wiering2012reinforcement, ernst2024introduction, wang2016learning, ding2020introduction, ghasemi2024introduction}

\subsection{Exploration vs Exploitation}
It is important to note that exploration and exploitation represent a fundamental trade-off in RL. The objective of exploration is to discover the effects of new actions, whereas the objective of exploitation is to select actions that have been shown to yield high rewards, to the best of the knowledge of the agent at that timestep.

Exploration techniques can be classified into two: undirected and directed exploration methods. Undirected exploration methods such as Semi-uniform ($\epsilon$-greedy) exploration and the Boltzmann exploration try to explore the whole state-action space by assigning positive probabilities to all possible actions \cite{sutton2018reinforcement, white1992role}. On the other hand, directed exploration methods like the $E^3$ algorithm and exploration bonus use the statistics obtained through past experiences to execute efficient exploration \cite{kearns1998near, sutton1990integrated}.

Balancing exploration and exploitation is a critical aspect of RL, and various strategies have been developed to manage this trade-off effectively. For instance, in  \cite{ishii2002control}, authors introduced a Model-based RL method that dynamically balances exploitation and exploration, particularly in changing environments. By using Bayesian inference with a forgetting effect, it estimates state-transition probabilities and adjusts the balance parameter based on action-outcome variations and environmental changes. Furthermore, \cite{schafer2021decoupling} proposed Decoupled RL (DeRL) which trains separate policies for exploration and exploitation and can be applied with on-policy and off-policy RL algorithms. Using decoupling policies, DeRL improves robustness and sample efficiency in sparse reward environments. More advanced methodologies have been used in \cite{wang2018exploration}, where the study investigated the trade-off between exploration and exploitation in continuous-time RL using an entropy-regularized reward function. As a result, the optimal exploration-exploitation balance was achieved through a Gaussian distribution for the control policy, where exploitation was captured by the mean and exploration by the variance. Moreover, various strategies such as $\epsilon$-c, Upper Confidence Bound (UCB), and Thompson Sampling are employed to balance this trade-off in simpler environments, like Bandits \cite{auer2002finite, sutton2018reinforcement, mnih2015human}.

In the subsequent sections, we will undertake an examination of various algorithms within the domain of RL, inclusive of the associated papers, as well as an analysis of their respective merits and drawbacks. It is imperative to acknowledge that these algorithms fall into three overarching categories: Value-based Methods, Policy-based Methods, and Actor-Critic Methods. Section \ref{sec:VBM} will initiate the discussion by focusing on Value-based Methods, delineated by its three core components: Tabular Model-free, Tabular Model-based, and Approximation Model-free Methods. Section \ref{sec:PBM} will subsequently address the sole method within the Policy-based Methods umbrella, namely Approximation Model-free. Furthermore, section \ref{sec:ACM} will offer detailed insights into Actor-Critic-based Methods. In section \ref{sec:Final_Notes}, we give a detailed explanation of how this survey is meant to be read to get the most out of it. Finally, section \ref{sec:conclusion} will provide a synthesis of the paper through a review of key points.

\section{Value-based Methods}\label{sec:VBM}

Value-based Methods in RL are techniques that focus on estimating the value of states or state-action pairs to guide decision-making. An essential component of the methodology is the learning of a \textbf{value function}, which quantifies the expected long-term reward for a given state under a given policy. Value-based Methods are ones that iteratively update their value estimates based on the observed rewards and transitions. Examples of Value-based methods algorithms include Q-learning and State-Action-Reward-State-Action (SARSA) (will be discussed briefly later in this chapter). These methods aim to derive an optimal policy by maximizing the value function, enabling the agent to choose actions that lead to the highest cumulative rewards.

Value-based Methods are divided into two broad categories: Tabular Model-based and Tabular Model-free methods. Model-based methods refer to a group of methods that rely on an explicit or learned model of the environment's dynamics. The model predicts how the environment will respond to an agent's actions (state transitions and rewards). On the other hand, Model-free methods do not rely on a model of the environment. Instead, they directly learn a policy or value function based on interactions with the environment.

We will begin by introducing the main part of Tabular Model-based methods, Dynamic Programming (DP). Next, we will explore both Tabular and Approximate Model-free algorithms. Finally, we will cover the advanced part of Tabular Model-based methods, Model-based Planning.

\subsection{Tabular Model-based Algorithms} 
\label{Sec_TMB}

In this subsection, we examine the first part of the Tabular Model-based methods in RL, Dynamic Programming methods. It must be noted that the advanced Tabular Model-based algorithms will be analyzed in section \ref{Sec_TMB2} along with various studies.

A Tabular Model-based algorithm is an example of an RL technique used for solving problems with a finite and discrete state and action space. These algorithms are explicitly based on the maintenance and updating of a table or matrix, which represents the dynamic nature of the environment and its reward. 'Model-based' algorithms are characterized by the fact that they involve the construction of an environmental model for use in decision-making.
Key characteristics of Tabular Model-based techniques include Model Representation, Planning and Policy Evaluation, and Value Iteration \cite{sutton2018reinforcement}. 

In Model Representation, the model depicts the dynamics in a tabular form with transition probabilities represented as $P(s'|s,a)$ and the reward function as $R(s, a)$.
These elements define the probability of transitioning to a new state $s'$ and the expected reward when taking action $a$ in state $s$. Planning and Policy Evaluation involves approximating the updating value function $V(s)$ iteratively using the Bellman equation until convergence. To calculate the optimal policy, the algorithm examines all possible future states and their associated actions. Value Iteration also approximates the updating value function $V(s)$ iteratively using the Bellman equation until convergence. Similar to Planning and Policy Evaluation, it analyzes all potential future states and their linked actions to determine the optimal policy.

Tabular Model-based algorithms can be divided into two general categories: DP and Model-based Planning, where this variant will be discussed in section \ref{Sec_MBP}.

\subsubsection{Dynamic Programming (DP)}
DP methods are fundamental techniques used to solve MDPs when a complete model of the environment is known. These methods are iterative and make use of the Bellman equations to compute the optimal policies. Two primary DP-based methods are \textbf{Policy Iteration} and \textbf{Value Iteration}.
We first start by examining Policy Iteration, then, we discuss Value Iteration.

\paragraph{Policy Iteration}
Policy Iteration is a method that iteratively improves the policy until it converges to the optimal policy. It consists of two main steps: policy evaluation and policy improvement \cite{sutton2018reinforcement}.
Policy Iteration consists of two different steps, Policy Evaluation and Improvement. \textbf{Policy Evaluation} calculates the value function \( V_\pi(s) \) for a given policy \( \pi \). This involves solving the Bellman expectation equation for the current policy:

\begin{equation}
    V_\pi(s) = \sum_{a \in A} \pi(a|s) \sum_{s' \in S} P(s'|s,a) [R(s,a,s') + \gamma V_\pi(s')]
\end{equation}

This step iteratively updates the value of each state under the current policy until the values converge \cite{ghasemi2024introduction}.
\textbf{Policy Improvement}, on the other hand, improves the policy by making it greedy with respect to the current value function:

\begin{equation}
    \pi'(s) = \arg\max_a \sum_{s'} P(s'|s,a) [R(s,a,s') + \gamma V_\pi(s')]
\end{equation}

This step updates the policy by selecting actions that maximize the expected value based on the current value function \cite{moerland2023model}. Alg. \ref{alg:Policy_Iteration} gives an overview of how policy iteration can be implemented.

Value Iteration is another approach in DP, which will be discussed in the next subsection.

\begin{algorithm}[t]
\caption{Policy Iteration}
\begin{algorithmic}[1]
\State Initialize policy \(\pi\) arbitrarily
\Repeat
    \State \parbox[t]{\dimexpr\linewidth-\algorithmicindent}{Perform policy evaluation to update the value function \(V_\pi\)\strut}
    \State \parbox[t]{\dimexpr\linewidth-\algorithmicindent}{Perform policy improvement to update the policy \(\pi\)\strut}
\Until{policy \(\pi\) converges}
\end{algorithmic}
\label{alg:Policy_Iteration}
\end{algorithm}

\paragraph{Value Iteration}
Value Iteration is another DP method that directly computes the optimal value function by iteratively updating the value of each state. It combines the steps of policy evaluation and policy improvement into a single step.
Value Iteration consists of two different steps, Value Update and Policy Extraction. \textbf{Value Update} updates the value function for each state based on the Bellman optimality equation:

\begin{equation}
    V(s) = \max_a \sum_{s'} P(s'|s,a) [R(s,a,s') + \gamma V(s')]
\end{equation}

This step involves iterating through all states and updating their values based on the maximum expected return of the possible actions.

\textbf{Policy Extraction}, happens once the value function has converged, the optimal policy can be extracted by selecting actions that maximize the expected value:

\begin{equation}
    \pi^*(s) = \arg\max_a \sum_{s'} P(s'|s,a) [R(s,a,s') + \gamma V(s')]
\end{equation}
Alg. \ref{alg:Value_Iteration} illustrates the implementation of Value Iteration.

\begin{algorithm}[t]
\caption{Value Iteration}
\begin{algorithmic}[1]
\State Initialize the value function \( V(s) \) arbitrarily
\Repeat
    \For{each state \( s \)}
        \State \parbox[t]{\dimexpr\linewidth-\algorithmicindent}{
    \raggedright Update \( V(s) \) using the Bellman optimality equation
\strut}
    \EndFor
\Until{the value function \( V(s) \) converges}
\State Extract the optimal policy \( \pi^* \) from the converged value function
\end{algorithmic}
\label{alg:Value_Iteration}
\end{algorithm}


Value Iteration is simpler to implement than Policy Iteration since it combines evaluation and improvement into a single process. Additionally, it is often faster in practice for many problems, as it does not require separate policy evaluation steps \cite{tamar2016value}. On the downside, Value Iteration may require a large number of iterations to converge, especially for problems with large state spaces. Furthermore, it can be less stable than Policy Iteration in some cases due to the combined update step \cite{sutton2018reinforcement}. A detailed comparison between Policy and Value Iterations is given in Table \ref{tab:PI_vs_VI}.

\begin{table*}[htb]
\renewcommand{\arraystretch}{1.2} 
\centering
\caption{Comparison of Policy Iteration and Value Iteration}
\begin{tabular}{|>{\raggedright\arraybackslash}p{3.5cm}|>{\centering\arraybackslash}p{6cm}|>{\centering\arraybackslash}p{6cm}|}
\hline
\textbf{Aspect} & \textbf{Policy Iteration} & \textbf{Value Iteration} \\
\hline
Convergence & Typically fewer iterations needed & May require more iterations \\
\hline
Complexity per Iteration & More complex (requires policy evaluation) & Simpler (single update step) \\
\hline
Stability & More stable due to separate steps & Can be less stable \\
\hline
Ease of Implementation & More complex & Simpler \\
\hline
Computational Cost & Higher per iteration & Lower per iteration \\
\hline
\end{tabular}
\label{tab:PI_vs_VI}
\end{table*}

Throughout the next subsection, we start exploring Tabular Model-free algorithms by introducing Monte Carlo (MC) methods first, then, Temporal Difference (TD) Learning methods later. 

\subsection{Tabular Model-free Algorithms}

Tabular Model-free algorithms are techniques suitable for problems with discrete state and action spaces that are typically small enough to be tabulated. Unlike Model-based algorithms, which require a model of the environment's dynamics, Model-free algorithms learn directly from interactions with the environment without understanding its underlying mechanics. In this context, we will explore various methods and algorithms that are categorized as Tabular Model-free algorithms, starting with MC Methods.

\subsubsection{Monte Carlo Methods}

In situations where complete knowledge of the environment is unavailable or undesirable, MC methods can be employed. 
MC methods rely on experience, using sample sequences of states, actions, and rewards obtained through real or simulated interactions with an environment \cite{sutton2018reinforcement}. Learning from real experience is notable because MC does not require prior knowledge of the environment's dynamics, yet it can still achieve optimal behavior. Similarly, learning from simulated experience is powerful. While a model is necessary, it only needs to generate sample transitions, not the complete probability distributions of all possible transitions as required in DP. 
To ensure well-defined returns, MC methods are used specifically for episodic tasks, where experiences are divided into episodes, and all episodes eventually terminate regardless of the selected actions. Changes to value estimates and policies occur only at the end of an episode (sample returns). As a result, MC methods show incremental behavior on an episode-by-episode basis instead of a step-by-step (online) fashion. While the term "Monte Carlo" is commonly used broadly for any estimation method involving a significant random component, here it specifically refers to methods based on averaging complete returns \cite{sutton2018reinforcement}.

In the following, we will explain how MC methods are used to learn the state-value function for a given policy $\pi$,where there are several ways of doing so.

\paragraph{Monte Carlo Estimation of State Values}

A straightforward method of estimating the value of a state based on experience is to average its observed returns after visits to the state. In terms of value, the state is defined as the anticipated cumulative discounted reward in the future. The average tends to converge to the expected value as more returns are observed (the law of large numbers), which is the basis for MC.

The value $v_\pi(s)$ of a state $s$ under policy $\pi$ can be estimated by considering a set of episodes obtained by following $\pi$ and passing through $s$. Each occurrence of state $s$ in an episode is known as a visit. When $s$ is visited multiple times within an episode, the first occurrence is referred to as the first visit to $s$. According to the First-visit MC method, $v_\pi(s)$ is calculated as the average of first visits to the visited states, while under the Every-visit MC method, the average is calculated based on all visits to $s$.

Consider a simple 3x3 grid world where an agent starts at the \textbf{top-left corner} (State $S_0$) and aims to reach the \textbf{bottom-right corner} (Goal State $G$). The agent can take one of four actions in each state: \textbf{up}, \textbf{down}, \textbf{left}, or \textbf{right}. Each action transitions the agent to an adjacent state unless it moves outside the grid boundaries, in which case the agent remains in its current state. The episode ends once the agent reaches the goal state, $G$.

\textbf{Key Rules:}
\begin{enumerate}
    \item States can be revisited during an episode.
    \item The return at the goal state is defined as the sum of rewards from the starting state to the goal state.
    \item Two approaches are considered for state-value estimation:
    \begin{itemize}
        \item \textbf{First-visit MC:} Only the first visit to each state in an episode is used for value estimation.
        \item \textbf{Every-visit MC:} All visits to each state in an episode are considered.
    \end{itemize}
\end{enumerate}

\textbf{Agent's Behavior Across Episodes:}

\textbf{Episode 1:}  
\begin{itemize}
    \item \textbf{Actions:} Right, Right, Down, Down (Reaches $G$).
    \item \textbf{Returns:} $G$ is visited once with a return of $5$.
\end{itemize}

\textbf{Episode 2:}  
\begin{itemize}
    \item \textbf{Actions:} Up, Right, Right, Down, Down (Reaches $G$).
    \item \textbf{Returns:} $G$ is visited twice, with returns of $8$ (first visit) and $8$ (second visit).
\end{itemize}

\textbf{Episode 3:}  
\begin{itemize}
    \item \textbf{Actions:} Right, Up, Right, Down, Down (Reaches $G$).
    \item \textbf{Returns:} $G$ is visited once with a return of $6$.
\end{itemize}

\textbf{State-Value Estimation for $G$:}

\begin{enumerate}
    \item \textbf{First-visit MC:}
    \begin{itemize}
        \item Considers only the \textbf{first visit} to $G$ in each episode.
        \item Returns for $G$:
        \begin{itemize}
            \item Episode 1: $5$
            \item Episode 2: $8$ (first visit only)
            \item Episode 3: $6$
        \end{itemize}
        \item \textbf{Average Return for $G$:} $\frac{5 + 8 + 6}{3} = 6.33$.
    \end{itemize}
    \item \textbf{Every-visit MC:}
    \begin{itemize}
        \item Considers \textbf{all visits} to $G$ in each episode.
        \item Returns for $G$:
        \begin{itemize}
            \item Episode 1: $5$
            \item Episode 2: $8, 8$
            \item Episode 3: $6$
        \end{itemize}
        \item \textbf{Average Return for $G$:} $\frac{5 + 8 + 8 + 6}{4} = 6.75$.
    \end{itemize}
\end{enumerate}

\textbf{General Insights:}
\begin{itemize}
    \item Using \textbf{First-visit MC}, each state value is updated using the average return from its first visits across episodes (e.g., $G$ in Episode 2 considers only the first return, $8$).
    \item Using \textbf{Every-visit MC}, the state value reflects the average of all returns across all visits to that state.
\end{itemize}

In MC, both the First-visit MC and the Every-visit MC methods converge toward $v_\pi(s)$ as the number of visits approaches infinity. Alg. \ref{alg:MC_Policy_Evaluation} examines the First-visit MC method for estimating $V_{\pi}$. The only difference between Every-visit and First-visit, as stated above, lies in line 8. For Every-visit MC, we should return following the every occurrence of state $s$.


\begin{algorithm}[t]
\caption{First-visit MC}
\begin{algorithmic}[1]
\State Initialize:
\State \(\pi \gets\) policy to be evaluated
\State \(V \gets\) an arbitrary state-value function
\State \(\text{Returns}(s) \gets\) an empty list, for all \(s \in \mathcal{S}\)
\Repeat
    \State \parbox[t]{\dimexpr\linewidth-\algorithmicindent}{
        \raggedright Generate an episode using \(\pi\)
    \strut}
    \For{each state \(s\) appearing in the episode}
        \State \parbox[t]{\dimexpr\linewidth-\algorithmicindent}{
            \raggedright \(G \gets\) return following the first occurrence of \\
            \(s\)
        \strut}
        \State \parbox[t]{\dimexpr\linewidth-\algorithmicindent}{
            \raggedright Append \(G\) to \(\text{Returns} (s)\)
        \strut}
        \State \parbox[t]{\dimexpr\linewidth-\algorithmicindent}{
            \raggedright \(V(s) \gets\) average(\(\text{Returns}(s)\))
        \strut}
    \EndFor
\Until{forever}
\end{algorithmic}
\label{alg:MC_Policy_Evaluation}
\end{algorithm}

\paragraph{MC Estimation of Action Values (with Exploration Starts)}

It becomes particularly advantageous to estimate action values (values associated with state-action pairs) rather than state values in the absence of an environment model. State values alone are adequate for determining a policy by examining one step ahead and selecting the action that leads to the optimal combination of reward and next state \cite{sutton2018reinforcement}. It is, however, insufficient to rely solely on state values without a model. An explicit estimate of each action's value is essential in order to provide meaningful guidance in the formulation of policy. MC methods are intended to accomplish this objective. As a first step, we address the issue of evaluating action values from a policy perspective in order to achieve this goal.

During policy evaluation for action values, we estimate $q_{\pi}(s,a)$, which represents the anticipated return when initiating in state $s$, taking action $a$, and following the policy $\pi$ \cite{fonteneau2010model}. In this task, MC methods are similar, except that visits to state-action pairs are used instead of states alone. When state $s$ is visited and action $a$ is taken during an episode, then the state-action pair $(s,a)$ is considered visited. Based on the average of returns following all visits to a state-action pair, the Every-visit MC method estimates its value. However, the First-visit MC method averages the returns after each state visit and action selection occurs. It is evident that both methods exhibit quadratic convergence as the number of visits to each state-action pair approaches infinity \cite{sutton2018reinforcement}.

A deterministic policy $\pi$ presents a significant challenge in that numerous state-action pairs may never be visited. When following $\pi$, only one action is observed for each state. As a consequence, MC estimates for the remaining actions do not improve with experience, presenting a significant problem. As a result of learning action values, one can choose among all available actions in each state more easily. It is crucial to estimate the value of all actions from each state and not only the one that is at present favored for the purpose of making informed comparisons.

It may be beneficial to explore starts in some situations, but they cannot be relied upon in all cases, especially when learning directly from the environment. Therefore, the initial conditions are less likely to be favorable in such situations. Alternatively, stochastic policies that select all actions in each state with a non-zero probability may be considered in order to ensure that all state-action pairs are encountered. As a first step, we will continue to assume that starts will be explored and conclude with a comprehensive MC simulation approach.

To begin, we consider an MC adaptation of classical Policy Iteration. We use this approach to iteratively evaluate and improve policy starting with an arbitrary policy $\pi_0$ and ending with an optimal policy and optimal action-value function:

\begin{equation}
\begin{split}
&\pi_0 \xrightarrow{\text{Evaluation}} q_{\pi_0} \xrightarrow{\text{Improvement}} \pi_1 \xrightarrow{\text{Evaluation}} \\ 
&q_{\pi_1} \xrightarrow{\text{Improvement}} \pi_2 \xrightarrow{\text{Evaluation}} \dots \xrightarrow{\text{Improvement}} \\ 
&\pi_* \xrightarrow{\text{Improvement}} q_{\pi_*}
\end{split}
\end{equation}

As the approximate action-value function approaches the true function asymptotically, many episodes occur. We will assume that there are an infinite number of episodes we observe and that these episodes are generated with explorations starts for now. Exploration starts referring to an assumption we make. We assume that all the states have a non-zero probability of starting. This approach encourages exploration as well though not practical as this assumption does not hold true in many real-world applications.
For any arbitrary policy $\pi_k$ under these assumptions, MC methods will compute $q_{\pi_k}$ accurately.

In order to improve the policy, it is necessary to make the policy greedy regarding the current value function. As a result, an action-value function is used to construct the greedy policy without requiring a model. Whenever an action-value function $q$ is there, a greedy policy is defined as one that selects the action with the maximal action-value for each state $s \in S$.

\begin{equation}
    \pi(s) \equiv \arg\max_a q(s,a)
\end{equation}

Policy Improvement can be executed by formulating each $\pi_{k+1}$ as the greedy policy with respect to $q_{\pi_k}$. The Policy Improvement theorem, as discussed earlier, is then applicable to $\pi_k$ and $\pi_{k+1}$ because, for all $s \in S$, there is:

\begin{equation}
\begin{split}
&q_{\pi_k}(s, \pi_{k+1}(s)) = q_{\pi_k}(s, \arg\max_a q_{\pi_k}(s,a)) = \\ &\max_a q_{\pi_k}(s,a) \geq q_{\pi_k}(s, \pi_k(s)) = v_{\pi_k}(s)
\end{split}
\end{equation}

Based on the General Policy Theorem \cite{sutton2018reinforcement}, each $\pi_{k+1}$ is uniformly superior to $\pi_k$, or equally optimal, which makes both policies optimal. As a result, the overall process converges to an optimal value function and policy. Using MC methods, optimal policies can be determined solely based on sample episodes, without the need for additional information about the environment's dynamics.

Despite the convergence guarantee for MC simulations, two key assumptions must be addressed to create a practical algorithm: the presence of exploration starts in episodes and the need for an infinite number of episodes for policy evaluation.
Our focus is on removing the assumption of an infinite number of episodes for policy evaluation. This can be achieved by approximating \( q_{\pi_k} \) during each evaluation, using measurements and assumptions to minimize error. Although this method could theoretically ensure correct convergence, it often requires an impractically large number of episodes, especially for complex problems. Alternatively, we can avoid relying on infinite episodes by not fully completing the evaluation before improving the policy. Instead, the value function is adjusted towards \( q_{\pi_k} \) incrementally across multiple steps, as seen in Generalized Policy Iteration (GPI). In Value Iteration, this approach is evident, where only one policy evaluation occurs between policy improvements.
MC policy iteration, by its nature, alternates between evaluation and improvement after each episode. The returns observed in an episode are used for evaluation, followed by policy improvement for all visited states.
Over the next subsection, we analyze the integration of Importance Sampling \cite{tokdar2010importance}, a well-known concept in Statistics, with MC methods.

\paragraph{Off-Policy Prediction via Importance Sampling}

The use of RL as a control method is confronted with a fundamental dilemma, namely the need to learn the value of actions based on the assumption of subsequent optimal behavior, contrasting with the need for non-optimal behavior to explore all possible actions in order to find the optimal action. This conundrum leaves us with the question of how we can learn about optimal policies while operating under exploratory policies. A Policy-based approach serves as a compromise in this situation. As part of this strategy, we seek to learn the action values for a policy that, while it is not optimal, is close to it and incorporates mechanisms for exploration. Due to its exploratory nature, it does not directly address the issue of learning the optimal policy action values.

Off-policy learning is an effective method of addressing this challenge by utilizing two distinct policies: a target policy ($\pi$) whose objective is to become the optimal policy, and a behavior policy ($b$) whose purpose is to generate behavior. The dual-policy framework allows exploration to occur independently of learning about the optimal policy, with learning occurring from data generated outside the target policy by behavior policy. Off-policy methods are more versatile and powerful than on-policy methods. On-policy methods can also be incorporated as a special case when both target and behavior policies are the same. 

On the other hand, off-policy methods introduce additional complexity, which requires the use of more sophisticated concepts and notations. Off-policy learning involves using data from a different policy, resulting in higher variance and slower convergence than on-policy learning. On-policy methods provide simplicity and direct learning from the agent's exploratory actions, while off-policy methods provide a robust framework for learning optimal policies indirectly through exploration guided by a separate behavior policy. Essentially, the dichotomy between on-policy and off-policy learning represents the exploration-exploitation trade-off that underlies RL, which enables agents to learn and adapt to complex environments in a variety of ways \cite{ghasemi2024introduction}. Alg. \ref{alg:Off_Policy_MC_Control} provides a general overview of this algorithm.


\begin{algorithm}[t]
\caption{Off-Policy Prediction (via Importance Sampling)}
\begin{algorithmic}[1]
\State Initialize, for all \(s \in \mathcal{S}\), \(a \in \mathcal{A}(s)\):
\State \(\quad Q(s, a) \in \mathbb{R}\) (arbitrarily)
\State \(\quad C(s, a) \gets 0\)
\State \(\quad \pi(s) \gets \arg\max_a Q(s, a)\) (ties broken randomly)
\Repeat
    \State \(b \gets\) any soft policy
    \State \parbox[t]{\dimexpr\linewidth-\algorithmicindent}{
        Generate an episode using \(b\): \\ 
        \((S_0, A_0, R_1, \ldots, S_{T-1}, A_{T-1}, R_T)\)\strut}
    \State \(G \gets 0\)
    \State \(W \gets 1\)
    \For{each step of episode, \(t = T-1, \ldots, 0\):}
        \State \(G \gets \gamma G + R_{t+1}\)
        \State \(C(S_t, A_t) \gets C(S_t, A_t) + W\)
        \State \parbox[t]{\dimexpr\linewidth-
        \algorithmicindent}{
        \(Q(S_t, A_t) \gets Q(S_t, A_t) + \frac{W}{C(S_t, A_t)} \\ 
        \big[G - Q(S_t, A_t)\big]\)\strut}
        \State \parbox[t]{\dimexpr\linewidth-\algorithmicindent}{
            \(\pi(S_t) \gets \arg\max_a Q(S_t, a)\) \\ (with ties broken consistently)\strut}
        \If{\(A_t \neq \pi(S_t)\)}
            \State \textbf{break} \Comment{Proceed to next episode}
        \EndIf
        \State \parbox[t]{\dimexpr\linewidth-\algorithmicindent}{
            \(W \gets W \cdot \frac{1}{b(A_t|S_t)}\)\strut}
    \EndFor
\Until{forever}
\end{algorithmic}
\label{alg:Off_Policy_MC_Control}
\end{algorithm}




Throughout the next few paragraphs, we will analyze selected research studies that have employed MC methods and its mentioned variations, analyzing their rationale and addressing their specific challenges. In analyzing these papers in depth, we intend to demonstrate that MC methods are versatile and effective for solving a wide range of problems while emphasizing the decision-making processes leading to their selection.

Based on MC with historical data, \cite{liu2018intelligent} proposed an intelligent train control approach enhancing energy efficiency and punctuality. It offered a Model-free approach, achieving 6.31\% energy savings and improving punctuality. However, the method's success depended heavily on the availability and quality of historical data, and its scalability to larger, more complex networks remained a consideration.

In \cite{subramanian2019renewal}, a Renewal MC (RMC) algorithm was developed to reduce variance, avoid delays in updating, and achieve quicker convergence to locally optimal policies. It worked well with continuous state and action spaces and was applicable across various fields, such as robotics and game theory. The method also introduced an approximate version for faster convergence with bounded errors. However, the performance of RMC was dependent on the chosen renewal set and its size.

In \cite{peters2005monte} authors introduced an MC off-policy strategy augmented by rough set theory, providing a novel approach. This integration offered new insights and methodologies in the field. However, the approach's complexity might challenge broader applicability, and the study's focus on theoretical formulations necessitated further empirical research for validation.

Authors in \cite{aghaei2022real} introduced a Bayesian Model-free Markov Chain Monte Carlo (MCMC) algorithm for policy search, specifically applied to a 2-DoF robotic manipulator. The algorithm demonstrated practicality and effectiveness in real implementations, adopting a gradient-free strategy that simplified the process and excelled in mastering complex trajectory control tasks within a limited number of iterations. However, its applicability might be confined to specific scenarios, and the high variance in the estimator presented challenges.

The research on MC Bayesian RL (MCBRL) by \cite{wang2012monte} introduced an innovative method that streamlined Bayesian RL (BRL). By sampling a limited set of hypotheses, it constructed a discrete, Partially Observable Markov Decision Process (POMDP), eliminating the need for conjugate distributions and facilitating the application of point-based approximation algorithms. This method was adaptable to fully or partially observable environments, showing reliable performance across diverse domains. However, the efficacy of the sampling process was contingent on the choice of prior distribution, and insufficient sample sizes might have affected performance.

In \cite{wu2017monte}, authors introduced a Factored MC Bayesian RL (FMCBRL) approach to solve the BRL problem online. It leveraged factored representations to reduce the size of learning parameters and applied partially observable Monte-Carlo planning as an online solver. This approach managed the complexity of BRL, enhancing scalability and efficiency in large-scale domains.

Researchers in \cite{baykal2019reinforcement} discussed developing self-learning agents for the Batak card game using MC methods for state-value estimation and artificial neural networks for function approximation. The approach handled the large state space effectively, enabling agents to improve gameplay over time. However, the study's focus on a specific card game might have limited the direct applicability of its findings to other domains.

Next, we will discuss another variant of MC, MC with Importance Sampling, which is an off-policy algorithm.

\paragraph{MC with Importance Sampling}

MC Importance Sampling is a method used to improve the efficiency of MC simulations when estimating expected values. It samples from the proposal distribution rather than directly from the original distribution. As demonstrated in Alg. \ref{alg:MC_Weighted_Importance_Sampling}, re-weighting the samples is based on the ratio of the original distribution to the proposal distribution. It is through this re-weighting that the bias introduced by sampling from the alternative distribution is corrected, allowing for a more accurate and efficient estimation of the data \cite{siegmund1976importance}.


\begin{algorithm}[t]
\caption{MC with Importance Sampling}
\begin{algorithmic}[1]
\State Initialize, for all \(s \in \mathcal{S}, a \in \mathcal{A}(s)\):
\State \hspace{1em} \(Q(s, a) \gets\) arbitrary
\State \hspace{1em} \(C(s, a) \gets 0\)
\State \hspace{1em} \(\mu(a|s) \gets\) an arbitrary soft behavior policy
\State \hspace{1em} \(\pi(a|s) \gets\) an arbitrary target policy
\Repeat \Comment{For ever}
    \State \parbox[t]{\dimexpr\linewidth-\algorithmicindent}{
        Generate an episode using \(\mu\): \\ 
        \((S_0, A_0, R_1, \dots, S_T, A_T, R_T, S_T)\)\strut}
    \State \(G \gets 0\)
    \State \(W \gets 1\)
    \For{\(t = T-1, T-2, \dots, 0\) down to 0}
        \State \(G \gets \gamma G + R_{t+1}\)
        \State \(C(S_t, A_t) \gets C(S_t, A_t) + W\)
        \State \parbox[t]{\dimexpr\linewidth-\algorithmicindent}{
            \(Q(S_t, A_t) \gets Q(S_t, A_t) + \frac{W}{C(S_t, A_t)} \\ 
            \big[G - Q(S_t, A_t)\big]\)\strut}
        \State \parbox[t]{\dimexpr\linewidth-\algorithmicindent}{
            \(W \gets W \cdot \frac{\pi(A_t|S_t)}{\mu(A_t|S_t)}\)\strut}
        \If{\(W = 0\)}
            \State \textbf{break} \Comment{Exit inner loop}
        \EndIf
    \EndFor
\Until{convergence or a stopping criterion is met}
\end{algorithmic}
\label{alg:MC_Weighted_Importance_Sampling}
\end{algorithm}

Let us analyze several papers regarding MC Importance Sampling.
In \cite{saha2002maximum}, authors presented a non-iterative approach for estimating the parameters of superimposed chirp signals in noise using an MC Importance Sampling method. The primary method utilized maximum likelihood to optimize the estimation process efficiently. This approach, which differed from traditional grid-search methods, focused on estimating chirp rates and frequencies, providing a practical solution to multidimensional problems. The technique was non-iterative, reducing computational complexity, and it achieved precise parameter estimation even under challenging conditions. Additionally, the method remained scalable for multiple signal scenarios without significant increases in computational load. However, performance might have been affected below certain signal-to-noise ratios, and the selection of MC samples and other parameters required problem-specific tuning.

In \cite{lassila2001efficient}, a method to estimate blocking probabilities in multi-cast loss systems through simulation was introduced, improving upon static MC methods with Importance Sampling. The technique divided the complex problem into simpler sub-problems focused on blocking probability contributions from individual links, using a distribution tailored to the blocking states of each link. An inverse convolution method for sample generation, coupled with a dynamic control algorithm, achieved significant variance reduction. Although this method efficiently allocated samples and reduced variance, its complexity and setup requirements might have made it less accessible compared to simpler approaches.

Authors in \cite{de2005adaptive} introduced a method for improving simulation efficiency in rare-event phenomena through Importance Sampling. Initially developed for Markovian random walks, this method was expanded to include non-Markovian scenarios and molecular dynamics simulations. It increased simulation efficiency by optimizing the sampling of successful transition paths and introduced a method for identifying an optimal importance function to maximize computational efficiency. However, optimizing the importance function could be challenging, particularly in systems with many states or those extending beyond Markovian frameworks.

\cite{bulteau2002new} presented an advanced MC Importance Sampling approach for assessing the reliability of stochastic flow networks. This method leveraged a recursive state-space decomposition to efficiently target critical network segments for sampling, reducing computational demands and enhancing the precision of reliability estimates. However, its complexity and dependence on specific network attributes might have limited its applicability in very large or intricate networks. 
Lastly, to wrap up MC methods, we shall examine another variation, On-Policy Monte Carlo (without Exploration Starts). 

\paragraph{On-Policy Monte Carlo (MC without Exploration Starts)}

MC with On-Policy Starts, also known as MC without Exploration Starts, is a method in which the agent learns based on episodes generated by following its current policy. In contrast to methods that use exploratory starts, where the agent can begin from any state-action pair, On-Policy MC uses only the actual experiences generated by following the policy in the environment. As a result of this approach, the agent estimates the value of the policy by averaging the cumulative rewards from all episodes beginning with the start of the policy and ending with the termination of the policy. It is important to ensure that sufficient exploration is conducted in order to learn accurate estimates of value since the agent's experiences are limited to what is dictated by the current policy. The agent may explore different state-action pairs over time by making policies stochastic or incorporating soft exploration strategies. The method is particularly useful in scenarios where exploration starts are not feasible or where policy must be discovered through direct interaction with the environment \cite{sutton2018reinforcement, wang2020convergence}.

We would like to mention two studies in this section, acknowledging the fact that there are more papers in the literature.
In \cite{peters2007approximation}, researchers introduced a novel off-policy MC learning method that integrated approximation spaces and rough set theory. This approach refined the RL process in dynamic environments by using weighted sampling based on observed behavior patterns, enabling a nuanced estimation of action values and improving the adaptability and efficiency of learning strategies. The method's effectiveness relied on accurately identifying and applying behavior patterns, particularly in complex environments. Further exploration and comparison with other advanced RL techniques could have provided a more comprehensive understanding of its efficacy and scalability across various applications.

In \cite{peters2005monte}, researchers presented a method that advanced RL by applying rough set theory and approximation spaces to evaluate state values with MC without Exploration Starts. Tested on simulated zebra danio fish behavior, this approach integrated observed patterns for a refined estimation of action values, showing promise for modeling complex systems. While the technique was innovative, it also introduced computational complexity, highlighting a balance between innovation and practical application. The research underscored the potential of pattern-based evaluation, with considerations for scalability and broader implementation.

\begin{table}[t]
\centering
\renewcommand{\arraystretch}{1.2} 
\caption{MC Papers Review}
\begin{tabular}{|>{\raggedright\arraybackslash}p{4cm}|>{\raggedright\arraybackslash}p{3cm}|}
\hline
\textbf{Application Domain} & \textbf{References} \\
\hline
Train Control and Energy Efficiency & \cite{liu2018intelligent} \\
\hline
Algorithmic RL (Renewal Theory, Rough Set Theory) & \cite{subramanian2019renewal}, \cite{peters2005monte} \\
\hline
Robotics and Trajectory Control & \cite{aghaei2022real} \\
\hline
Bayesian RL & \cite{wang2012monte}, \cite{wu2017monte} \\
\hline
Game Strategy and Card Games & \cite{baykal2019reinforcement} \\
\hline
Signal Processing and Parameter Estimation & \cite{saha2002maximum} \\
\hline
Network Reliability and Blocking Probabilities & \cite{lassila2001efficient} \\
\hline
Simulation Efficiency in Rare-Event Phenomena & \cite{de2005adaptive} \\
\hline
Rough Set Theory and Approximation Spaces & \cite{bulteau2002new} \\
\hline
\end{tabular}
\label{tab:Monte_Carlo_Paper}
\end{table}

Table \ref{tab:Monte_Carlo_Paper} gives a summary of articles that utilized MC methods
Over the next subsection, we start analyzing another main category of Tabular Model-free methods, TD Learning, as one of the most fundamental concepts within RL.

\subsubsection{Temporal Difference (TD) Learning}

TD learning is undoubtedly the most fundamental and innovative concept. A combination of MC methods and DP is used in this method. On one hand, similar to MC approaches, TD learning can be used to acquire knowledge from unprocessed experience without the need for a model that describes the dynamics of the environment. On the other hand, TD algorithms are also similar to DP in that they refine predictions using previously learned estimates instead of requiring a definitive outcome in order to proceed (known as bootstrapping).

It is important to recognize that the expression within brackets in the TD(0) update represents a type of error. This error measures the discrepancy between the estimated value of $S_t$ and a more refined estimate, $R_{t+1} + \gamma V(S_{t+1})$. This discrepancy is known as the TD error.
TD(0) is an essential concept for understanding other TD learning algorithms, such as Q-learning, SARSA, and Double Q-learning, among others, which we will explore in subsequent subsections. The following sections introduce and analyze various TD-based papers in order to gain a general understanding of the mentioned algorithms before delving into their details. It will be started with TD(0)-Replay and will be finished with N-step SARSA.

Alg. \ref{alg:TD_Policy_Evaluation} describes tabular TD(0) for estimating $V_\pi$. While MC methods require waiting until the conclusion of an episode to calculate the update to $V(S_t)$ as $G_t$, TD methods only need to wait until the subsequent timestep. At time $t+1$, TD methods swiftly establish a target and perform an effective update using the observed reward $R_{t+1}$ and the estimated value $V(S_{t+1})$. The most basic form of the TD method executes this update as follows (line 7 in Alg. \ref{alg:TD_Policy_Evaluation}):

\begin{equation}
V(S_t) \leftarrow V(S_t) + \alpha \left[ R_{t+1} + \gamma V(S_{t+1}) - V(S_t) \right]
\end{equation}

Upon transitioning to $S_{t+1}$ and receiving $R_{t+1}$, TD methods implement the update immediately. In contrast, the target for a MC update is $G_t$, while the target for the TD update is $R_{t+1} + \gamma V(S_{t+1})$. This TD method is called TD(0), or one-step TD. TD(0) is a bootstrapping method, like DP, as its update is in part on an existing estimate \cite{sutton2018reinforcement}.

\begin{algorithm}[t]
\caption{Tabular TD(0)}
\begin{algorithmic}[1]
\State Initialize the value function \( V(s) \) arbitrarily (e.g., \( V(s) = 0 \) for all \( s \in S^+ \))
\Repeat \Comment{For each episode}
    \State Initialize state \( S \)
    \Repeat \Comment{For each step of the episode}
        \State \( A \gets \) action given by policy \( \pi \) for \( S \)
        \State \parbox[t]{\dimexpr\linewidth-\algorithmicindent}{ Take action \( A \); observe reward \( R \) and next \\
        state \( S' \)\strut}
        \State \( V(S) \gets V(S) + \alpha [R + \gamma V(S') - V(S)] \)
        \State \( S \gets S' \)
    \Until{\( S \) is terminal}
\Until{convergence or for a specified number of episodes}
\end{algorithmic}
\label{alg:TD_Policy_Evaluation}
\end{algorithm}

\paragraph{TD(0)-Replay}

TD(0)-Replay algorithm (Alg.\ref{alg:TD0_Replay}), introduced in \cite{altahhan2018td}, enhances policy learning efficiency by leveraging full experience replay. Utilizing the agent's entire history of interactions allows for comprehensive updates to the value function and policy at each step. This approach accelerates convergence to optimal policies, particularly in complex environments where acquiring new experiences is costly or challenging.
TD(0)-Replay efficiently uses full replay of past experiences, crucial in environments where re-experiencing events is costly or impossible. It promotes rapid optimization of the agent's value function, leading to quicker learning and adaptation in dynamic environments, offering significant advantages over methods without replay mechanisms. The algorithm's adaptability is highlighted through dynamic updates of weight parameters based on historical data, ensuring effectiveness as environmental conditions evolve. Its broad applicability across various domains underscores its versatility, making it suitable for both theoretical research and practical applications. 


\begin{algorithm}[t]
\caption{TD(0)-Replay}
\begin{algorithmic}[1]
\State \textbf{Input:} \(\alpha, \gamma, \theta_{\text{init}}\)
\State \(\theta \gets \theta_{\text{init}}\)
\Loop \Comment{Over episodes}
    \State Obtain initial state \(S\), features \(\phi\)
    \State \(\hat{e} \gets I_{n \times n}\), \(e \gets \mathbf{0}_{n \times 1}\)
    \While{(terminal state has not been reached)}
        \State Act according to the policy
        \State \parbox[t]{\dimexpr\linewidth-\algorithmicindent}{
            Observe next reward \(R = R_{t+1}\), next state \\
            \(\hat{S} = S_{t+1}\), \\ 
            and its features \(\hat{\phi} = \phi_{t+1}\)\strut}
        \State \(\alpha \gets \ell(\alpha)\)
        \State \parbox[t]{\dimexpr\linewidth-\algorithmicindent}{
            \(e \gets e + \alpha \phi \left( \left( \gamma \hat{\phi} - \phi \right)^\top e + R \right)\)\strut}
        \State \parbox[t]{\dimexpr\linewidth-\algorithmicindent}{
            \(\hat{e} \gets \hat{e} + \alpha \phi \left[ \left( \gamma \hat{\phi} - \phi \right)^\top \hat{e} \right]\)\strut}
        \State \parbox[t]{\dimexpr\linewidth-\algorithmicindent}{
            \(\theta \gets \theta + \hat{e} + e\)\strut}
        \State \(\phi \gets \hat{\phi}\)
    \EndWhile
\EndLoop
\end{algorithmic}
\label{alg:TD0_Replay}
\end{algorithm}

\(\delta_t\) refers to the the TD error, \(R_{t+1}\) is the reward signal, \(\gamma\) is a discount factor, \(\phi_t^T\) is the transpose of feature vector \(\phi_t\) obtained through current state \(S_t\), \(\theta_t^T\) is the transpose of the weight vector \(\theta_t\), and \(\alpha_t\) is a learning step; all varies according to time step \(t\).
Over the next paragraphs, we analyze another variation of TD methods, TD($\lambda$).

\paragraph{TD($\lambda$)}

TD($\lambda$) is a powerful and flexible algorithm introduced in \cite{sutton1988learning}. TD($\lambda$) generalizes the simpler TD(0) method by introducing a parameter $\lambda$ (lambda), which controls the weighting of n-step returns, blending one-step updates and MC methods. This allows the algorithm to consider the entire trajectory of experiences rather than just the immediate next step. It introduces the concept of eligibility traces, which keep a record of states and how eligible they are for learning updates. It can be seen as a credit assignment problem to answer the question of how many steps before or after what we would like to update are contributing. Alg. \ref{alg:TD_lambda} comprehensively represents the TD($\lambda$) algorithm. An eligibility trace (lines 9-11) is a temporary record of the occurrence of an event, such as the visiting of a state, and these traces decay over time, controlled by the parameter $\lambda$, allowing the algorithm to attribute credit for rewards to prior states in a temporally distributed manner.

The value update in TD($\lambda$) is given by (line 13):
\begin{equation}
V(s) \leftarrow V(s) + \alpha \delta e(s)
\end{equation}
where $\alpha$ is the learning rate, $\delta$ is the TD error ($\delta = R + \gamma V(s') - V(s)$), and $e(s)$ is the eligibility trace for state $s$. TD($\lambda$) provides a continuum of methods from TD(0) (pure TD) to MC (full return) methods, often learning more efficiently than either extreme by balancing the bias-variance trade-off. It can quickly propagate information about value estimates through the state space, improving learning efficiency \cite{sutton1988learning, sutton2018reinforcement}.

\begin{algorithm}[t]
\caption{TD(\(\lambda\))}
\begin{algorithmic}[1]
\State Initialize \(V(s)\) arbitrarily (but set to 0 if \(s\) is terminal)
\Repeat \text{(for each episode)}
    \State Initialize \(E(s) = 0\), for all \(s \in S\)
    \State Initialize \(S\)
    \Repeat \Comment{(for each step of episode)}
        \State \(A \gets\) action given by \(\pi\) for \(S\)
        \State \parbox[t]{\dimexpr\linewidth-\algorithmicindent}{
        Take action \(A\), observe reward \(R\), and \\ next state \(S'\)}
        \State \(\delta \gets R + \gamma V(S') - V(S)\)
        \State \(E(S) \gets E(S) + 1\) \Comment{(accumulating traces)}
        \State or \(E(S) \gets (1 - \alpha)E(S) + 1\) \\ \Comment{(Dutch traces)}
        \State or \(E(S) \gets 1\) \Comment{(replacing traces)}
        \For{all \(s \in S\)}
            \State \parbox[t]{\dimexpr\linewidth-\algorithmicindent}{
                \(V(s) \gets V(s) + \alpha \delta E(s)\)\strut}
            \State \parbox[t]{\dimexpr\linewidth-\algorithmicindent}{
                \(E(s) \gets \gamma \lambda E(s)\)\strut}
        \EndFor
        \State \(S \gets S'\)
    \Until \(S\) is terminal
\Until{convergence or a stopping criterion is met}
\end{algorithmic}
\label{alg:TD_lambda}
\end{algorithm}

Now, we need to cover various papers that employed a form of TD($\lambda$) to analyze the practicality of it.
In \cite{baxter1999knightcap} authors presented KnightCap, a chess program that learned its evaluation function using TDLeaf($\lambda$), a variation of the TD($\lambda$) algorithm integrated with game-tree search. KnightCap improved its rating from 1650 to 2150 in just 308 games over three days by playing on the Free Internet Chess Server. The use of TDLeaf($\lambda$) in conjunction with game-tree search marked a significant advancement in integrating RL with traditional AI methods in chess. The approach leveraged the strengths of both deep evaluation through game-tree search and the learning capabilities of TD methods. KnightCap's rapid improvement demonstrated the effectiveness of online learning against diverse opponents, emphasizing the importance of playing against varied strategies. The adaptation of TD($\lambda$) to TDLeaf($\lambda$) for deep minimax search allowed dynamic improvement of the evaluation function based on real-game outcomes. The success of the algorithm highlighted its potential in environments that required learning from interactions, although starting with intelligent initial parameters was beneficial. While further validation in varied game scenarios and against stronger AI opponents would provide a more comprehensive assessment, the results were promising.

A convergence theorem for TD learning by generalizing convergence theorem to handle arbitrary time steps, crucial for predicting future outcomes based on past states was implemented in \cite{dayan1992convergence}. This work advanced the understanding of TD($\lambda$), ensuring convergence even with linearly dependent state representations, and bridged the gap between TD learning and DP. The extension of Q-learning convergence proof to TD(0) reinforced the strong convergence properties of TD methods, making them applicable in various RL scenarios. While the paper was primarily theoretical, the results provided strong guarantees about the stability and reliability of TD learning. Further empirical validation in diverse, dynamic environments would offer a complete assessment of the proposed methods.

Authors in \cite{dayan1994td}, strengthened the theoretical foundations of TD learning by proving that TD($\lambda$) algorithms converge with probability 1 under certain conditions. Building on earlier work by Sutton and Watkins, the authors provided a rigorous proof of convergence for TD($\lambda$) in the general case, ensuring reliable and consistent predictions. This theoretical advancement was crucial for the broader adoption of TD learning, providing strong assurances about the stability and reliability of the learning process. The detailed analysis and proof techniques contributed to a deeper understanding of TD learning dynamics, informing future research and development in the field. Although the paper was theoretical, it provided a significant contribution by proving the convergence of TD($\lambda$) with probability 1.

In \cite{wiering2007two}, researchers introduced two algorithms, QV($\lambda$)-learning and the Actor-Critic Learning Automaton (ACLA), built upon TD($\lambda$) methods. QV($\lambda$)-learning combined value function learning with a form of Q-learning, while ACLA used a learning automaton-like update rule for the actor component. These algorithms were tested across various environments, demonstrating their robustness and generalizability. QV($\lambda$)-learning combined the strengths of traditional Q-learning and Actor-Critic methods, leading to more stable and efficient learning. ACLA introduced a flexible mechanism for updating action preferences and adapting to various environments. The empirical results showed that these algorithms outperformed conventional methods such as Q($\lambda$)-learning, SARSA($\lambda$), and Actor-Critic, particularly in learning speed and final performance. While further validation in diverse and dynamic real-world scenarios was needed, the results highlighted the potential for broader applications in complex and partially observable domains. 
%
%
%
In the next subsection, we start analyzing another form of the TD method, N-step Bootstrapping.

\paragraph{N-step Bootstrapping}

Having covered TD(0)-Replay and TD($\lambda$), we now have a solid base to explore bootstrapping methods. N-step bootstrapping is a key technique that extends the concept of updating estimated value functions across multiple steps rather than just based on immediate rewards or the value of the next state. This method provides a middle ground between MC methods, which delay updating value estimates until the end of an episode, and one-step TD methods that update values immediately based on the subsequent state. In N-step bootstrapping, the update of the value function is based on n subsequent rewards and the estimated value of the state that follows these n steps. The primary advantage of this approach is that it can lead to faster learning and reduced variance in the updates compared to one-step methods \cite{de2018per, sutton2018reinforcement, sutton1988learning}. The generic update rule for the N-step method can be expressed as:

\begin{equation}
\begin{aligned}
V(S_t) &\leftarrow V(S_t) + \alpha \Bigg[ \sum_{k=0}^{n-1} \gamma^k R_{t+k+1} \\
&+ \gamma^n V(S_{t+n}) - V(S_t) \Bigg]
\end{aligned}
\end{equation}
where \( R_{t+k+1} \) are the rewards received after taking action \( A_t \) from state \( S_t \), \( \gamma \) is the discount factor, and \( \alpha \) is the learning rate.

\paragraph{N-step TD Prediction} 

As previously explored, N-step bootstrapping forms the basis for various adaptations like N-step TD, and n-step SARSA, among others. N-step methods (where \( n \neq 1 \)) distinguish themselves by looking ahead multiple steps, which can be adjusted according to specific requirements. This approach contrasts with MC-based methods, which update the value of each state based on the complete sequence of observed rewards from that state until the episode concludes. It also differs from one-step TD methods, which focus solely on the next reward and use the value of the state one step later as a surrogate for the subsequent rewards \cite{de2018multi}.
This gap can be bridged by performing updates based on an intermediate number of rewards - more than one, but fewer than the total number following until the end of the episode. A two-step update, for example, would utilize the first two rewards and the estimated value of the state two steps ahead. Additionally, this can be extended to three-step updates, four-step updates, and beyond, each incorporating an increasing number of rewards and subsequent state values.

Alg. \ref{alg:n_step_TD} gives a general overview of the N-step TD Prediction algorithm. For a given state \( S_t \) at time \( t \), the update rule for the value function \( V \) in N-step TD prediction is (line 21):
\begin{equation}
V(S_t) \leftarrow V(S_t) + \alpha (G_t^{(n)} - V(S_t))
\end{equation}
where \( \alpha \) is the learning rate, and \( G_t^{(n)} \) is derived as follows (line 18):
\begin{equation}
G_t^{(n)} = R_{t+1} + \gamma R_{t+2} + \cdots + \gamma^{n-1} R_{t+n} + \gamma^n V(S_{t+n})
\end{equation}


%

\begin{algorithm}[t]
\caption{N-step TD for Estimating \( V \approx v_\pi \)}
\begin{algorithmic}[1]
\State Input: a policy \( \pi \)
\State Algorithm parameters: step size \( \alpha \in (0, 1] \), a positive integer \( n \)
\State Initialize \( V(s) \) arbitrarily for all \( s \in S \)
\State All store and access operations (for \( S_t \) and \( R_t \)) use index \( \text{mod } n+1 \)
\Repeat \Comment{Loop for each episode}
    \State Initialize and store \( S_0 \) such that \( S_0 \neq \text{terminal} \)
    \State \( T \gets \infty \)
    \For{\( t = 0, 1, 2, \ldots \)}
        \If{\( t < T \)}
            \State Take an action according to \( \pi(\cdot | S_t) \)
            \State \parbox[t]{\dimexpr\linewidth-\algorithmicindent}{
                Observe and store the next reward as \\
                \( R_{t+1} \)
                and the next state as \( S_{t+1} \)\strut}
            \If{\( S_{t+1} \) is terminal}
                \State \( T \gets t + 1 \)
            \EndIf
        \EndIf
        \State \parbox[t]{\dimexpr\linewidth-\algorithmicindent}{
    \( \tau \gets t - n + 1 \) \\ 
    \Comment{\( \tau \) is the time whose state's estimate \\ is being updated}\strut}
        \If{\( \tau \geq 0 \)}
            \State \parbox[t]{\dimexpr\linewidth-\algorithmicindent}{
                \( G \gets \sum_{i=\tau+1}^{\min(\tau+n, T)} \gamma^{i-\tau-1} R_i \)\strut}
            \If{\( \tau + n < T \)}
                \State \parbox[t]{\dimexpr\linewidth-\algorithmicindent}{
                    \( G \gets G + \gamma^n V(S_{\tau+n}) \)\strut}
            \EndIf
            \State \parbox[t]{\dimexpr\linewidth-\algorithmicindent}{
                \( V(S_\tau) \gets V(S_\tau) + \alpha \left[G - V(S_\tau)\right] \)\strut}
        \EndIf
    \EndFor
\Until{\( \tau = T - 1 \)}
\end{algorithmic}
\label{alg:n_step_TD}
\end{algorithm}

Let us now examine research papers that have implemented these methods and assess their advantages and disadvantages.
Authors in \cite{sutton2004temporal} introduced an advanced generalization of TD learning by incorporating networks of interrelated predictions. This method expanded traditional TD techniques by connecting multiple predictions over time, allowing for a more detailed approach to learning and representing predictions. TD networks facilitated the learning of interconnected predictions, offering a broader range of representable and learnable predictions. This advancement was particularly useful in solving problems that traditional TD methods could not, as demonstrated by experiments on the random-walk problem and predictive state representations. Experimental results showed that TD networks could learn complex prediction tasks more efficiently than MC methods, particularly in terms of data efficiency and learning speed. The practical benefits of TD networks were evident in scenarios requiring specific sequences of actions. While the paper focused on small-scale experiments, the results suggested that TD networks had significant potential for broader applicability and effectiveness in various domains.

In \cite{de2018per}, an advanced multi-step TD learning technique was introduced, emphasizing the use of per-decision control variates to reduce variance in updates. This method significantly improved the performance of multi-step TD algorithms, particularly in off-policy learning contexts, by enhancing stability and convergence speed. Empirical results from tasks like the 5x5 Grid World and Mountain Car showed that the n-step SARSA method outperformed standard n-step Expected SARSA in reducing root-mean-square error and improving learning efficiency. The approach was compatible with function approximation, underscoring its practical applicability in complex environments.

Researchers in \cite{chen2021lyapunov} developed a unified framework to study finite-sample convergence guarantees of various Value-based asynchronous RL algorithms. By reformulating these RL algorithms as Markovian Stochastic Approximation algorithms and employing a Lyapunov analysis, the authors derived mean-square error bounds on the convergence of these algorithms. This framework provided a systematic approach to analyzing the convergence properties of algorithms like Q-learning, n-step TD, TD(\(\lambda\)), and off-policy TD algorithms such as V-trace. The paper effectively addressed the challenges of handling asynchronous updates and offered robust convergence guarantees through detailed finite-sample mean-square convergence bounds.

In \cite{lee2024analysis}, the "deadly triad" in RL—off-policy learning, bootstrapping, and function approximation—was addressed through an in-depth theoretical analysis of multi-step TD learning. The paper provided a comprehensive theoretical foundation for understanding the behavior of these algorithms in off-policy settings with linear function approximation. A notable contribution was the introduction of Model-based deterministic counterparts to multi-step TD learning algorithms, enhancing the robustness of the findings. The results demonstrated that multi-step TD learning algorithms could converge to meaningful solutions when the sampling horizon \( n \) was sufficiently large, addressing a critical issue of divergence in certain conditions.

In \cite{de2018unified}, researchers delved into various facets of TD learning, particularly focusing on multi-step methods. The thesis introduced the innovative use of control variates in multi-step TD learning to reduce variance in return estimates, thereby enhancing both learning speed and accuracy. The work also presented a unified framework for multi-step TD methods, extending the n-step Q(\(\sigma\)) algorithm and proposing the n-step CV Q(\(\sigma\)) and Q(\(\sigma, \lambda\)) algorithms. This unification clarified the relationships between different multi-step TD algorithms and their variants. Additionally, the thesis extended predictive knowledge representation into the frequency domain, allowing TD learning agents to detect periodic structures in return, providing a more comprehensive representation of the environment.

"Undelayed N-step TD prediction" (TD-P), developed in \cite{zuters2010realizing}, integrated techniques like eligibility traces, value function approximators, and environmental models. This method employed Neural Networks to predict future steps in a learning episode, combining RL techniques to enhance learning efficiency and performance. By using a forward-looking mechanism, the TD-P method sought to gather additional information that traditional backward-looking eligibility traces might miss, leading to more accurate value function updates and better decision-making. The TD-P method was particularly designed for partially observable environments, utilizing Neural Networks to handle complex and continuous state-action spaces effectively.
To further solidify our knowledge of TD methods, we will analyze the off-policy version of N-step Learning over the following paragraphs.

\paragraph{N-step Off-policy Learning}

N-step off-policy learning is an advanced method that combines the concepts of N-step returns with off-policy updates. This approach leverages the benefits of multi-step returns to improve the stability and performance of learning algorithms while allowing the use of data generated by a different policy (the behavior policy $b$) than the one being improved (the target policy $\pi$) \cite{sutton2018reinforcement}.
Multi-step returns differ from one-step methods by considering cumulative rewards over multiple steps, providing a more comprehensive view of future rewards and leading to more accurate value estimates. In off-policy learning, the policy used to generate behavior (behavior policy) is different from the policy being optimized (target policy), allowing for the reuse of past experiences and improving sample efficiency by enabling learning from demonstrations or historical data. Importance sampling is used to correct the discrepancy between the behavior policy and the target policy in N-step off-policy learning, where Importance Sampling ratios adjust the updates to account for differences in action probabilities under the two policies.
The N-step off-policy return, \( G_t^{(n)} \), can be calculated in the following way:

\begin{equation}
G_t^{(n)} = R_{t+1} + \gamma R_{t+2} + \cdots + \gamma^{n-1} R_{t+n} + \gamma^n V(S_{t+n})
\end{equation}

To ensure the update is off-policy, Importance Sampling ratios are incorporated as follows:

\begin{equation}
\begin{aligned}
G_t^{(n)} = &\sum_{k=0}^{n-1} \gamma^k \left( \prod_{i=0}^{k-1} \rho_{t+i} \right) R_{t+k+1} \\
&+ \left( \prod_{i=0}^{n-1} \rho_{t+i} \right) \gamma^n V(S_{t+n})
\end{aligned}
\end{equation}
where \( \rho_{t+i} = \frac{\pi(A_{t+i} \mid S_{t+i})}{b(A_{t+i} \mid S_{t+i})} \) is the Importance Sampling ratio, with \( \pi \) as the target policy and \( b \) the behavior policy \cite{szepesvari2022algorithms}.

Now that we have acquired the fundamental knowledge, we can examine papers based on the N-step Off Policy.
\cite{wang2020greedy} introduced the Greedy Multi-step Value Iteration (GM-VI) algorithm, which approximated the optimal value function using a novel multi-step bootstrapping technique. The method dynamically adjusted the step size along each trajectory based on a greedy principle, effectively balancing information propagation and estimation accuracy. GM-VI's adaptive step size mechanism is adjusted according to the quality of trajectory data, enhancing learning efficiency and robustness. The method could safely learn from arbitrary behavior policies without needing off-policy corrections, simplifying the algorithm and reducing variance. Theoretical analysis showed that GM-VI converged to the optimal value function faster than traditional one-step Bellman optimality operators. Empirical results demonstrated state-of-the-art performance on standard benchmarks, with GM-VI showing superior sample efficiency and reward performance compared to classical algorithms like Mountain Car and Acrobot.

The core strength of \cite{mahmood2017multi} lies in its innovative approach to eliminating the use of Importance Sampling ratios in multi-step TD learning. By removing these ratios, the method reduced estimation variance, enhancing stability and efficiency in off-policy learning. The introduction of action-dependent bootstrapping parameters allowed the algorithm to adapt flexibly to different state-action pairs, further reducing variance in updates. Empirical validation on challenging off-policy tasks demonstrated the algorithm's stability and superior performance compared to state-of-the-art counterparts, highlighting its practical applicability. 

\cite{wu2023bias} introduced innovative bias management techniques in multi-step Goal-Conditioned RL (GCRL) by categorizing and addressing shooting and shifting biases. This approach allowed for larger step sizes, enhancing learning efficiency and performance. The proposed methods were validated across various tasks, showing superior performance compared to baseline multi-step GCRL benchmarks. The use of quantile regression to manage biases effectively demonstrated the practical applicability of the approach. The introduction of resilient strategies for bias management ensured robust improvement in learning efficiency, with empirical results indicating effectiveness in diverse scenarios.
Table \ref{tab:TD_Papers_Review} gives an overview of the TD-based papers.

\begin{table}[t]
\centering
\renewcommand{\arraystretch}{1.2} 
 \caption{TD (and its variations) Papers Review}
\begin{tabular}{|>{\raggedright\arraybackslash}p{4cm}|>{\raggedright\arraybackslash}p{3cm}|}
\hline
\textbf{Application Domain} & \textbf{References} \\
\hline
General RL (Policy learning, raw experience) & \cite{altahhan2018td}, \cite{sutton1988learning} \\
\hline
Games (Chess) & \cite{baxter1999knightcap} \\
\hline
Theoretical Research (Convergence, stability) & \cite{dayan1992convergence}, \cite{dayan1994td}, \cite{chen2021lyapunov}, \cite{lee2024analysis}, \cite{de2018unified}, \cite{wang2020greedy} \\
\hline
Dynamic Environments (Mazes, Mountain Car, Atari) & \cite{de2018per}  \\
\hline
Partially Observable Environments (Predictions) &  \cite{wiering2007two},\cite{sutton2004temporal}, \cite{zuters2010realizing}
 \\
\hline
Benchmark Tasks (Mountain Car, Acrobot, GCRL) & \cite{ watkins1992q}, \cite{wu2023bias} \\
\hline
\end{tabular}
\label{tab:TD_Papers_Review}
\end{table}
Now, it is time to study and analyze one of the most widely used algorithms in RL, Q-learning.

\paragraph{Q-learning}

Moving on from TD(0), a significant breakthrough was made by \cite{watkins1989learning} with the introduction of Q-learning, a Model-free algorithm considered as off-policy TD control. Q-learning enables an agent to learn the value of an action in a particular state through experience, without requiring a model of the environment. It operates on the principle of learning an action-value function that gives the expected utility of taking a given action in each state and following a fixed policy thereafter. 

A general overview of Q-learning is demonstrated in Alg. \ref{alg:Q_learning}. The core of the Q-learning algorithm involves updating the Q-values (action-value pairs), where the learned action-value function, denoted as \( Q \), approximates \( q_* \), the optimal action-value function, regardless of the policy being followed. This significantly simplifies the algorithm's analysis and has facilitated early proofs of convergence. However, the policy still influences the process by determining which state-action pairs are visited and subsequently updated (lines 4-9).

\begin{equation}
\begin{aligned}
Q(S_t, A_t) \leftarrow Q(S_t, A_t) 
+ \alpha \big[ & R_{t+1} + \gamma \max_a Q(S_{t+1}, a) \\
& - Q(S_t, A_t) \big]
\end{aligned}
\end{equation}


\begin{algorithm}[t]
\caption{Q-learning}
\begin{algorithmic}[1]
\State \parbox[t]{\dimexpr\linewidth-\algorithmicindent}{
    Initialize \(Q(s, a)\), \(\forall s \in S, a \in A(s)\), arbitrarily, and \\ 
    \( Q(\text{terminal-state}, \cdot) = 0 \)\strut}
\Repeat \Comment{(for each episode)}
    \State Initialize \(S\)
    \Repeat \Comment{(for each step of episode)}
        \State \parbox[t]{\dimexpr\linewidth-\algorithmicindent}{ Choose \(A\) from \(S\) using policy derived from \\
        \(Q\) (e.g., \(\epsilon\)-greedy)\strut}
        \State Take action \(A\), observe \(R\), \(S'\)
        \State \parbox[t]{\dimexpr\linewidth-\algorithmicindent}{
            \( Q(S, A) \gets Q(S, A) + \\
            \alpha \big[ R + \gamma \max_a Q(S', a) - Q(S, A) \big] \)\strut}
        \State \(S \gets S'\)
    \Until \(S\) is terminal
\Until{convergence or a stopping criterion is met}
\end{algorithmic}
\label{alg:Q_learning}
\end{algorithm}

Now that we have established a foundational understanding of Q-learning, it is appropriate to explore the specifics of research papers that have utilized this algorithm. Q-learning, being fundamental and relatively straightforward, has been extensively applied across numerous studies. Here, we will briefly touch upon a variety of notable studies.

Researchers in \cite{bi2023comparative} investigated the performance differences between deterministic and stochastic policies within a grid-world problem using Q-learning. The authors developed a flexible agent capable of operating under both policy types to determine which parameters maximized cumulative reward. Their results indicated the superiority of deterministic policies in achieving higher rewards in a structured task environment. The study's strength lies in its clear methodological execution, systematically exploring the impact of policy variations on learning outcomes. However, the study was confined to a simulated grid world, which may not fully capture the complexities of real-world environments, potentially reducing the generalizability of the findings. Additionally, the paper focused on policy optimization without significant consideration of the computational costs associated with each policy type.

In \cite{spano2019efficient}, the authors explored the development of a hardware architecture optimized for Q-learning, focusing on real-time applications. Key innovations included low power usage, high throughput, and minimal use of hardware resources. The implementation, tested on an Evaluation Kit, demonstrated improved performance metrics such as speed, power consumption, and hardware resource use compared to existing Q-learning hardware accelerators. While the study presented a comprehensive approach to optimizing Q-learning for hardware implementation, making it suitable for real-time and Internet of Things (IoT) applications, it was limited by the specific hardware used for testing and the types of environments evaluated. The generalizability of the findings to other hardware or more complex real-world applications was not fully explored.

The application of Q-learning to improve power allocation in Wireless Body Area Networks (WBANs) was explored in \cite{nasreen2022overview}. The focus was on enhancing energy efficiency while maintaining effective communication within the network, particularly through controlling transmission power, reducing interference, and optimizing routing paths. The paper addressed critical aspects of WBANs that impact their practical deployment, especially in healthcare settings. While it presented a robust approach to managing interference and power consumption, the paper did not discuss the computational overhead introduced by the Q-learning algorithm and game theory applications, which was crucial for feasibility in devices with limited computational capabilities.

In \cite{huang2022application}, authors explored improvements to the Q-learning algorithm's efficiency using massively parallel computing techniques, such as multi-threading and Graphics Processing Unit (GPU) computing. The paper addressed the issue of slow convergence in traditional Q-learning, particularly in complex, real-world system control problems that demanded quick adaptation to dynamic environments. By leveraging GPUs and multi-threading, the authors significantly decreased convergence time, which was vital for real-time applications like robotics and industrial control. However, the paper did not thoroughly examine how parallel processing affected the Q-learning algorithm's integrity and reliability, particularly under different computational loads.

Researchers in \cite{darapaneni2020automated} investigated the use of Q-learning to automate portfolio re-balancing. The research applied basic Q-learning agents to discern trading patterns across 15 Indian financial assets using technical indicators. The paper innovatively harnessed Q-learning to improve portfolio re-balancing, a domain where traditional rule-based systems might not perform optimally. However, the simplicity of the Q-learning models used might not have fully captured the intricate dependencies and dynamics of highly fluctuating markets.

Authors in \cite{daswani2013q} extended traditional Q-learning to handle POMDPs by incorporating \( L_0 \) regularization to manage the complexity of the state representation derived from the agent's history. This innovative approach transformed non-Markov and partially observable environments into a history-based RL framework, allowing Q-learning to be applicable in more complex scenarios. While the method showed improved computational and memory efficiency, the reliance on \( L_0 \) regularization could have made the learning process highly dependent on the initial choice of features, potentially limiting adaptability in dynamic environments.

An innovative approach to enhancing the performance of Proportional Integral Derivative (PID) controllers for magnetic levitation train systems through the integration of Q-learning was introduced in \cite{shou2023design}. This method adapted PID parameters in real-time, maintaining optimal levitation control despite non-linearities and uncertainties. While it showed improved performance over traditional PID controllers, the requirement for continuous learning and adjustment could be computationally intensive.

A novel application of Deep Q-learning (discussed in subsection \ref{Sec_DQN}) to optimize parameter settings in Hadoop, improving its efficiency by iteratively adjusting configurations based on feedback from performance metrics introduced in \cite{akshay2023enhancing}. This approach reduced the manual effort and expertise required for parameter tuning and demonstrated marked improvements in processing speeds. However, the model's simplification of the parameter space might have overlooked interactions that could achieve further optimization, and the scalability and adaptability in varying operational environments remained somewhat uncertain.

Researchers in \cite{yao2022improving} presented a novel approach to enhancing the accuracy of nuclei segmentation in pathological images using Q-learning and Deep Q-Network (DQN) (will be discussed in \ref{Sec_DQN}) algorithms. The study reported improvements in Intersection over Union for segmentation, critical in cancer diagnosis. While the approach adapted the segmentation threshold dynamically, leading to better outcomes, the reliance on RL introduced computational complexity and required significant training data.

The sampling efficiency of Q-learning was discussed in \cite{jin2018q}, exploring the feasibility of making Model-free algorithms as sample-efficient as Model-based counterparts. The research introduced a variant of Q-learning that incorporated UCB exploration, achieving a regret bound that approached the best possible by Model-based methods. This paper marked a significant theoretical advance by providing rigorous proof that Q-learning could achieve sublinear regret in episodic MDPs without requiring a simulator. However, the analysis relied on assumptions such as an episodic structure and precise model knowledge, which might not have translated well to more uncertain environments.

Researchers in \cite{da2018parallel} examined the deployment of Q-learning on Field Programmable Gate Arrays (FPGAs) to enhance processing speed by leveraging parallel computing capabilities. The study demonstrated substantial acceleration of the Q-learning process, making it suitable for real-time scenarios. However, the focus on a particular FPGA model might have restricted the broader applicability of the findings to different hardware platforms.

The challenge of managing frequent handovers in high-speed railway systems using a Q-learning-based approach was addressed in \cite{wang2023q}. The authors proposed a scheme that minimized unnecessary handovers and enhanced network performance by dynamically adjusting to changes in the network environment. While the approach showed promise, its complexity, computational demands, and need for real-time processing presented challenges for practical implementation.

In \cite{wang2021taxi} authors explored optimizing taxi dispatching and routing using Q-learning and MDPs to enhance taxi service efficiency in urban environments. The method enabled dynamic adjustments based on real-time conditions, potentially alleviating urban traffic congestion. However, the model's effectiveness heavily depended on the accuracy and comprehensiveness of input data, and the computational complexity might have posed challenges for real-time deployment.

The study \cite{xiao2022random} addressed the optimization of network routing within Optical Transport Networks using a Q-learning-based algorithm under the Software-Defined Networking framework. The approach improved network capacity and efficiency by managing routes based on learned network conditions. While the method outperformed traditional routing strategies, its scalability and computational complexity in real-world scenarios remained areas for further exploration.

In \cite{qu2011visual}, authors proposed a Q-learning algorithm motivated by internal stimuli, specifically visual novelty, to enhance autonomous learning in robots. The method blended Q-learning with a cognitive developmental approach, making the robot's learning process more dynamic and responsive to new stimuli. While the approach reduced computational costs and enhanced adaptability, its scalability in complex environments was not thoroughly explored.

The use of Q-learning and Double Q-learning (discussed in the next subsection) to manage the duty cycles of IoT devices in environmental monitoring was analyzed in \cite{paterova2022robustness}. The study demonstrated significant advancements in optimizing IoT device operations, enhancing energy efficiency and operational effectiveness. However, the performance heavily depended on environmental parameters, and the increased memory requirements for Double Q-learning might have posed challenges for memory-constrained IoT devices.

The study \cite{nassar2019reinforcement} presented a robust implementation of RL to address resource allocation challenges in Fog Radio Access Networks (Fog RAN) for IoTs. The use of Q-learning effectively adapted to changes in network conditions, improving network performance and reducing latency. While the approach showed promise, the reliance on Q-learning might not have fully captured the complexities of larger-scale IoT networks, and the exploration-exploitation trade-off could have led to suboptimal performance in highly dynamic environments.

The overview of the reviewed papers categorized by their respective domains is presented in Table \ref{tab:Q_Learning_Papers}. 
Q-learning has an overestimation bias primarily because of the max operator used in the Q-value update rule, which tends to favor overestimated action values. To address this, researchers introduced a new architecture to tackle this issue, which we will discuss in the next subsection.

\begin{table}[t]
\centering
\renewcommand{\arraystretch}{1.2} 
\caption{Q-learning Papers Review }
\begin{tabular}{|>{\raggedright\arraybackslash}p{4cm}|>{\raggedright\arraybackslash}p{3cm}|}
\hline
\textbf{Application Domain} & \textbf{References} \\
\hline
General RL (Policy learning, raw experience) & \cite{bi2023comparative}, \cite{jin2018q}  \\
\hline
Games and Simulations (Chess, StarCraft) & \cite{daswani2013q}, \cite{wender2012applying} \\
\hline
Theoretical Research (Convergence, Stability) & \cite{dayan1992convergence}, \cite{watkins1989learning} \\
\hline
Dynamic and Complex Environments (Mazes, Mountain Car, Atari) & \cite{suh2021sarsa}, \cite{wiering2007two}  \\
\hline
Partially Observable Environments (Predictions, POMDPs) &  \cite{iima2008swarm} \\
\hline
Real-time Systems and Hardware Implementations (FPGA, Real-time applications) & \cite{spano2019efficient}, \cite{nasreen2022overview}, \cite{da2018parallel} \\
\hline
Energy Efficiency and Power Management (IoT, WBAN, PID controllers) & \cite{paterova2022robustness} \\
\hline
Financial Applications (Portfolio Rebalancing) & \cite{darapaneni2020automated} \\
\hline
Data Management and Processing (Hadoop, Pathological Images) & \cite{akshay2023enhancing}, \cite{yao2022improving} \\
\hline
Network Optimization (Optical Transport Networks, Fog RAN) & \cite{xiao2022random}, \cite{nassar2019reinforcement} \\
\hline
Transportation Systems (Railway, Taxi, Electric Vehicles) & \cite{shou2023design}, \cite{wang2023q}\\
\hline
Autonomous Systems and Robotics (Learning, Routing) & \cite{wang2021taxi}, \cite{qu2011visual} \\
\hline
\end{tabular}
\label{tab:Q_Learning_Papers}
\end{table}

\paragraph{Double Q-learning}

Double Q-learning was introduced in \cite{hasselt2010double} as an enhancement to traditional Q-learning to reduce the overestimation of action values, which can be problematic in environments with stochastic rewards. Traditional Q-learning tends to overestimate because it uses the maximum action value as an approximation for the maximum expected action value. To address this, as shown in Alg. \ref{alg:Double_Q_learning}, Double Q-learning maintains two separate estimators (Q-tables), $Q_A$ and $Q_B$ (line 1). Each estimator is updated independently using the maximum value from the other estimator, reducing the overestimation bias typically seen in standard Q-learning \cite{hasselt2010double} (lines 3-10).

The core update rule for Double Q-learning is as follows:
Selecting an action $a$ based on the average of $Q_A$ and $Q_B$. Updating one of the Q-functions randomly with a probability of 0.5, for example, $Q_A$, using (line 7):

\begin{equation}
\begin{aligned}
Q_A(s,a) \leftarrow &\, Q_A(s,a) \\
&+ \alpha \Big( r + \gamma Q_B \left( s', \arg\max_a Q_A(s',a) \right) \\
&\quad - Q_A(s,a) \Big)
\end{aligned}
\end{equation}
where $s'$ is the next state in the environment. Alternatively, we shall update $Q_B$ similarly using (line 10):

\begin{equation}
\begin{aligned}
Q_B(s,a) \leftarrow &\, Q_B(s,a) \\
&+ \alpha \Big( r + \gamma Q_A \left( s', \arg\max_a Q_B(s',a) \right) \\
&\quad - Q_B(s,a) \Big)
\end{aligned}
\end{equation}

In essence, the approach alternates updates between $Q_A$ and $Q_B$ using the maximum action value from the other table, which is believed to provide a more unbiased estimate of the underlying value function. This method helps to mitigate the positive bias seen in traditional Q-learning by occasionally underestimating the maximum expected values, aiming to strike a balance closer to true expectations. 
Over the following paragraphs, we will examine a handful of research papers that utilized Double Q-learning.


\begin{algorithm}[t]
\caption{Double Q-learning}
\begin{algorithmic}[1]
\State Initialize \(Q^A\), \(Q^B\), \(s\)
\Repeat
    \State \parbox[t]{\dimexpr\linewidth-\algorithmicindent}{ Choose \(a\) based on \(Q^A(s, \cdot)\) and \(Q^B(s, \cdot)\), observe \(r\), \(s'\)\strut}
    \State \parbox[t]{\dimexpr\linewidth-\algorithmicindent}{ Choose (e.g., random) either UPDATE(A) or UPDATE(B)\strut}
    \If{UPDATE(A)}
        \State Define \(a^* = \arg\max_a Q^A(s', a)\)
        \State \parbox[t]{\dimexpr\linewidth-\algorithmicindent}{
            \(Q^A(s, a) \gets Q^A(s, a) \\
            + \alpha(s, a) \left( r + \gamma Q^B(s', a^*) - Q^A(s, a) \right)\)\strut}
    \ElsIf{UPDATE(B)}
        \State Define \(b^* = \arg\max_a Q^B(s', a)\)
        \State \parbox[t]{\dimexpr\linewidth-\algorithmicindent}{
            \(Q^B(s, a) \gets Q^B(s, a) \\
            + \alpha(s, a) \left( r + \gamma Q^A(s', b^*) - Q^B(s, a) \right)\)\strut}
    \EndIf
    \State \(s \gets s'\)
\Until{end}
\end{algorithmic}
\label{alg:Double_Q_learning}
\end{algorithm}

The focus of \cite{ben2023efficient} was on the FPGA-based implementation of the Double Q-learning algorithm, emphasizing its efficiency in stochastic environments and its superiority over standard Q-learning due to reduced overestimation biases. This implementation of FPGA allowed for the parallel processing of actions, significantly speeding up the learning process. This was particularly beneficial for applications requiring real-time decision-making. Additionally, the use of asynchronous reading and synchronous writing memory architectures optimized data exchange and reduced the hardware footprint, which was a critical advancement in hardware implementations of RL algorithms. However, while the hardware implementation was efficient for the stated range of states (from 8 to 256 states), the paper did not extensively discuss scalability beyond this range, which might be a limitation for environments with a much larger state space. The paper focused on the hardware aspect without a detailed discussion of how the implementation could be adapted to different RL scenarios or environments, which may have limited its applicability without additional modifications or tuning.

The integration of A* pathfinding with Double Q-learning to optimize route planning in autonomous driving systems was investigated in \cite{jamshidi2021autonomous}. This innovative approach aimed to enhance route efficiency and safety by minimizing the common problem of action value overestimations found in standard Q-learning. The paper introduced a novel combination of A* pathfinding with RL, providing a dual strategy that leveraged the deterministic benefits of A* for initial path planning and the adaptive strengths of the proposed method for real-time adjustments to dynamic conditions, such as unexpected obstacles. This synergy allowed for a balanced approach to navigating real-world driving scenarios efficiently. By employing Double Q-learning, the system addressed and mitigated the issue of action value overestimation, which was prevalent in RL. This enhancement was crucial for autonomous driving applications where decisions had to be both accurate and dependable to ensure safety and operational reliability. On the other hand, this combination, while robust, introduced a significant computational demand that might have impacted the system's performance in real-time scenarios. Quick decision-making was essential in dynamic driving environments, and the increased computational load could have hindered the system's ability to respond promptly. Additionally, the promising method introduced lacked a detailed exploration of how this hybrid model performed across different environmental conditions or traffic scenarios. This limitation might have affected the model's effectiveness in diverse settings without further adaptation or refinement.

Researchers in \cite{paterova2021data} delved into optimizing IoT devices' power management by employing a Double Q-learning based controller. This Double Data-Driven Self-Learning (DDDSL) controller dynamically adjusted operational duty cycles, leveraging predictive data analytics to enhance power efficiency significantly. A notable strength of the paper was the improved operational efficiency introduced by the Double Q-learning, which effectively handled the overestimation issues found in standard Q-learning within stochastic environments. This led to more precise power management decisions, crucial for prolonging battery life and minimizing energy usage in IoT devices. Furthermore, the DDDSL controller showed a marked performance enhancement, outperforming traditional fixed duty cycle controllers by 42--50\%, and the previous Data-Driven Self-Learning (DDSL) model by 2--12\%. However, while the performance improvements were compelling, they were obtained under specific conditions, which might have limited the broader applicability of the findings without further adaptations or validations for different operational environments or IoT device configurations.

A Double Q-learning based routing protocol for optimizing maritime network routing, crucial for effective communication in maritime search and rescue operations, was developed in \cite{fan2023double}. The use of Double Q-learning in this context aimed to tackle the overestimation problems inherent in Q-learning protocols, which was a significant improvement in maintaining stability in the model's predictions and actions. This approach not only enhanced the routing efficiency but also incorporated a trust management system to ensure the reliability of data transfers and safeguard against packet-dropping attacks. The protocol demonstrated robust performance in various simulated attack scenarios with efficient energy consumption and minimal resource footprint, crucial for the resource-constrained environments in which maritime operations occurred. While the proposed method showed promising results in simulations, the complexity of real-world application scenarios could have posed challenges. Maritime environments were highly dynamic with numerous unpredictable elements, which might have affected the consistency of the performance gains observed in controlled simulations. Additionally, the scalability of this approach when applied to very large-scale networks or under extreme conditions typical of maritime emergencies could have required further validation. The integration of such sophisticated systems also raised concerns about the computational overhead and the practical deployment in existing maritime communication infrastructures.

Authors in \cite{konecny2022double} explored the application of Double Q-learning to manage adaptive wavelet compression in environmental sensor networks. This method aimed to optimize data transmission efficiency by dynamically adjusting compression levels based on real-time communication bandwidth availability. A key advantage of this approach was its high adaptability, which ensured efficient bandwidth utilization and minimized data loss even under fluctuating network conditions, critical for remote environmental monitoring stations where connectivity might have been inconsistent. Furthermore, this integration helped significantly reduce the risk of overestimating action values, a common problem in Q-learning, which could have led to suboptimal compression settings. However, the complexity of this implementation posed a notable challenge, particularly in environments where computational resources were limited. The necessity for managing two separate Q-values for each action increased the computational demand, potentially impacting the system's real-time response capabilities. Additionally, the performance of the algorithm heavily depended on the accuracy of the network condition assessments, with inaccuracies potentially leading to inefficient compression and data transmission.

In \cite{huang2017double}, authors explored the application of Double Q-learning to enhance dynamic voltage and frequency scaling in multi-core real-time systems. Their approach addressed the inherent overestimation bias found in Q-learning, aiming to provide more accurate and reliable power management. A major strength of their work lies in the innovative use of Double Q-learning to mitigate the overestimation issue common in single-estimator Q-learning methods. By utilizing two estimators, the system could potentially make more informed decisions that better-balanced power consumption against system performance needs. Additionally, their thorough simulation-based evaluation indicated that this method could outperform traditional methods, offering substantial energy savings across various system conditions. Nevertheless, managing two Q-value estimators increased the computational load, which could have challenged the limited resources available in real-time systems where rapid processing was paramount. Furthermore, while the simulation results were decent, translating these outcomes to real-world scenarios required additional adjustments to accommodate the diverse nature and unpredictability of real system workloads.
Table \ref{tab:Double_Q_Papers} summarizes the examined papers by their domain.

\begin{table}[t]
\centering
\renewcommand{\arraystretch}{1.2} 
\caption{Double Q-learning Papers Review}
\begin{tabular}{|>{\raggedright\arraybackslash}p{4cm}|>{\raggedright\arraybackslash}p{3cm}|}
\hline
\textbf{Application Domain} & \textbf{References} \\
\hline
Hardware Implementations (FPGA, Real-Time Systems) & \cite{ben2023efficient}, \cite{huang2017double} \\
\hline
IoT and Power Management (IoT Devices, Maritime Network Routing) & \cite{paterova2021data}, \cite{fan2023double} \\
\hline
Data Transmission and Compression (Environmental Monitoring) & \cite{konecny2022double} \\
\hline
\end{tabular}
\label{tab:Double_Q_Papers}
\end{table}

Transitioning from one off-policy TD algorithm, Q-learning, and its variation, Double Q-learning, which utilizes two Q functions, we now turn our attention to another TD-based algorithm, SARSA, in the next subsection.

\paragraph{State-Action-Reward-State-Action (SARSA)}

SARSA algorithm is an on-policy method that updates the action-value function (Q-function) incrementally and in an online manner. Originating from the work in \cite{rummery1994line}, SARSA is distinguished by its approach of learning the Q-values directly from the policy being executed. SARSA is characterized by its on-policy learning approach, where the policy used to make decisions is the same as the policy being evaluated and improved. This contrasts with off-policy methods, such as Q-learning, which may learn about the optimal policy independently from the agent's actions.

The core of the SARSA algorithm lies in its method for updating the Q-values. The updates occur according to the following rule \cite{sutton2018reinforcement} (Alg. \ref{alg:Sarsa}, line 8):

\begin{equation}
Q(s,a) \leftarrow Q(s,a) + \alpha [r + \gamma Q(s',a') - Q(s,a)]
\end{equation}
The 
The SARSA and Q-learning algorithms update their estimates of the action-value function based on the information gleaned from actions taken. In Q-learning, the update is based on the maximum estimated future reward, represented by the maximum Q-value of the next state, regardless of the action taken. In contrast, SARSA updates its Q-values using the action actually taken in the next state, not necessarily the best possible action. In other words, SARSA does not wait till the agent takes the next action and calculates the maximum of it. It simply takes the next available possible action. A fundamental algorithm of the SARSA is presented in Alg.~\ref{alg:Sarsa}.
Several research studies that used SARSA are examined below.


\begin{algorithm}[t]
\caption{SARSA}
\begin{algorithmic}[1]
\State \parbox[t]{\dimexpr\linewidth-\algorithmicindent}{
    Initialize \(Q(s, a)\), \(\forall s \in S, a \in A(s)\), arbitrarily, and \\ 
    \(Q(\text{terminal-state}, \cdot) = 0\)\strut}
\Repeat \Comment{(for each episode)}
    \State Initialize \(S\)
    \State \parbox[t]{\dimexpr\linewidth-\algorithmicindent}{ Choose \(A\) from \(S\) using policy derived from \(Q\) (e.g., \(\epsilon\)-greedy)\strut}
    \Repeat \Comment{(for each step of the episode)}
        \State Take action \(A\), observe \(R\), \(S'\)
        \State \parbox[t]{\dimexpr\linewidth-\algorithmicindent}{ Choose \(A'\) from \(S'\) using policy derived \\ from
        \(Q\) (e.g., \(\epsilon\)-greedy)\strut}
        \State \parbox[t]{\dimexpr\linewidth-\algorithmicindent}{
            \(Q(S, A) \gets Q(S, A) \\
            + \alpha \left[ R + \gamma Q(S', A') - Q(S, A) \right]\)\strut}
        \State \(S \gets S'\); \(A \gets A'\)
    \Until \(S\) is terminal
\Until{convergence or a stopping criterion is met}
\end{algorithmic}
\label{alg:Sarsa}
\end{algorithm}

The application of the SARSA algorithm in a simulated shepherding scenario where a dog herds sheep towards a target was investigated in \cite{go2016reinforcement}. This RL model incorporated a discretized state and action space and designed a specific reward system to facilitate learning. The application of SARSA to a complex, dynamic task like shepherding demonstrated the algorithm's versatility and its potential in environments involving multiple agents and stochastic elements. The model successfully taught a dog to herd sheep by learning to reach sub-goals, which simplified the learning process and improved the manageability of the task. However, the discretization of the state and action spaces might have limited the dog's movement and decision-making capabilities, potentially leading to less optimal performance in more realistic or varied environments. The stochastic nature of sheep movement and the complexity of the shepherding task led to a significant learning time, with notable success only after many episodes, indicating potential efficiency issues in more demanding or time-sensitive applications.

Authors in \cite{wender2012applying} primarily focused on the implementation of Q-learning and SARSA algorithms, incorporating eligibility traces to improve handling delayed rewards. The research demonstrated the effectiveness of RL algorithms in navigating dynamic tasks like micro-managing combat units in real-time strategy games. Adding eligibility traces significantly boosted the algorithms’ learning from sequences of interdependent actions, essential in fast-paced, chaotic environments. Using a commercial game like StarCraft as a testbed introduced real-world complexity often absent in simulated settings. This method confirmed the practicality of RL algorithms in real scenarios and highlighted their potential to adapt to commercial applications. The algorithms showed promise in small-scale combat, but scaling to larger, more complex battles in StarCraft or similar games remained uncertain. Concerns arose about computational demands and the efficiency of learning optimal strategies without extensive prior training. While effective within StarCraft, their broader applicability to other real-time strategy games or applications remained untested. The specialized design of state and action spaces for StarCraft could have hindered transferring these methods to different domains without significant modifications.

In \cite{jaiswal2020green}, authors studied the enhancement of energy efficiency in IoT networks through strategic power and channel resource allocation using a SARSA-based algorithm. This approach addressed the unpredictable nature of energy from renewable sources and the variability of wireless channels in real-time settings. A major contribution of the study lay in its innovative use of a Model-free, on-policy RL method to manage energy distribution across IoT nodes. This method efficiently handled the stochastic nature of energy harvesting and channel conditions, optimizing network performance in terms of energy efficiency and longevity. Integrating SARSA with linear function approximation helped refine solutions to continuous state and action spaces, enhancing the practicality in real-world IoT applications. However, relying on linear function approximation introduced limitations in capturing the full complexity of interactions in dynamic, multi-dimensional state spaces typical of IoT environments. While the proposed SARSA algorithm demonstrated network efficiency improvements, implementing such an RL system in real IoT networks posed challenges. These included the need for continual learning and adaptation to changing conditions, impacting the deployment's feasibility and scalability.

Authors in \cite{wang2020peer} explored applying SARSA to optimize peer-to-peer (P2P) electricity transactions among small-scale users in smart energy systems. This research aimed to enhance economic efficiency and reliability in decentralized energy markets. A significant strength lies in its innovative application of SARSA to a complex, dynamic energy trading system. Researchers modeled P2P electricity transactions as an MDP, allowing the system to handle uncertainties in small-scale energy trading, like fluctuating prices and varying demands. This modeling enabled the system to learn optimal transaction strategies over time, potentially enhancing both efficiency and profitability in energy trading. However, the approach had limitations. SARSA, while effective in learning optimal policies through trial and error, required extensive interaction with the environment to achieve satisfactory performance. This data requirement could have been a drawback in real-world applications where immediate decisions were necessary, and historical transaction data was limited. Moreover, implementing such a system in a live environment, where real-time decision-making was crucial, posed additional challenges, including the need for robust computational resources to handle continuous state and action space calculations.

In \cite{suh2021sarsa}, they examined integrating SARSA(0) with Fully Homomorphic Encryption (FHE) for cloud-based control systems to ensure data confidentiality while performing RL computations on encrypted data. A significant strength was preserving privacy in cloud environments where sensitive control data could have been vulnerable to breaches. Using FHE allowed the RL algorithm to execute without decrypting the data, providing a robust method for maintaining confidentiality. The paper successfully demonstrated this method on a classical pole-balancing problem, showing it was theoretically sound and practically feasible. However, implementing SARSA(0) over FHE introduced challenges related to computational overhead and latency due to encryption operations. These factors could have impacted the efficiency and scalability of the RL system, especially in environments requiring real-time decision-making. Additionally, encryption-induced delays and managing encrypted computations might have limited this method's application to scenarios where control tasks could tolerate such delays.

Authors in \cite{iima2008swarm} presented a novel application of SARSA within a swarm RL framework to solve optimization problems more efficiently, particularly those involving large negative rewards. The authors incorporated individual learning and cooperative information exchange among multiple agents, aiming to speed up learning and enhance decision-making efficiency. This approach significantly leveraged swarm intelligence and RL, particularly SARSA’s ability to handle tasks with substantial negative rewards. By enabling agents to learn from individual experiences and shared insights, the system could converge to optimal policies more swiftly than traditional single-agent or non-cooperative multi-agent systems. Implementing the shortest path problem demonstrated the method’s practicality and effectiveness, showcasing improved learning speeds and robustness against pitfalls with significant penalties. Nonetheless, managing multiple agents and their interactions increased the computational overhead and complexity of the learning process. Moreover, the approach heavily relied on designing the information-sharing protocol among agents, which, if not optimized, could have led to inefficiencies or suboptimal learning outcomes. The generalized application of this method across different environments or RL tasks remained to be thoroughly tested, suggesting potential limitations in adaptability.

In \cite{aljohani2022real}, authors explored the application of SARSA to optimize routing for Electric Vehicles (EVs) to minimize energy consumption. This approach adapted to real-time driving conditions to reduce on-road energy needs by selecting routes with lower energy requirements. The research’s strength lies in its real-time application and use of SARSA to learn and predict the most energy-efficient routes under various conditions, like traffic and road type. By utilizing Markov chain models to estimate energy requirements based on actual driving data, the framework aimed to extend the driving range of EVs, crucial for reducing range anxiety. However, the reliance on real-time data and the inherent variability of driving conditions presented challenges. The accuracy of the SARSA model’s predictions heavily depended on the data quality and immediacy, which could have been compromised by limitations in data transmission or processing delays. Implementing this system in a real-world environment could have posed scalability and adaptability challenges, particularly concerning integration with existing vehicle navigation systems and the computational demands on the vehicle’s hardware.

A variant of SARSA which incorporates expectations over all possible next actions instead of relying solely on the sampled next action is introduced in the next subsection.

\paragraph{Expected SARSA}

The Expected SARSA algorithm, proposed in \cite{van2009theoretical}, extended the classic SARSA algorithm. Expected SARSA differed from standard SARSA by incorporating expectations over all possible next actions instead of relying solely on the sampled next action. This modification reduced the update rule's variance, potentially allowing for faster and more stable convergence in learning tasks. The algorithm operated under the premise that by averaging all possible actions from the next state (weighted by their probability under the current policy), it could achieve a more stable estimate of state-action values. Expected SARSA’s convergence was guaranteed under conditions similar to those required by SARSA, such as all state-action pairs being visited infinitely often. \cite{van2009theoretical} provided proof that Expected SARSA converged to the optimal action-value function under typical RL assumptions, like finite state and action spaces, and a policy that became greedy in the limit with infinite exploration. Empirically, Expected SARSA outperformed both SARSA and Q-learning in various domains, especially in tasks where certain actions could lead to significant negative consequences. By incorporating the expectation over all possible next actions, the algorithm effectively smoothed out learning updates, which helped in environments where certain decisions or state transitions led to high variability in rewards \cite{van2009theoretical, sutton2018reinforcement}. In summary, Expected SARSA provides a robust approach to learning in stochastic environments by reducing the variance inherent in updates of state-action values. This leads to more reliable learning performance, particularly in complex environments where action outcomes are highly uncertain. Expected SARSA update rule is:

\begin{equation}
\begin{aligned}
Q(s_t,a_t) \leftarrow Q(s_t,a_t) + \alpha \Bigg[ r_{t+1} \\ + \gamma \sum_{a} \pi(a|s_{t+1}) Q(s_{t+1},a)
- Q(s_t,a_t) \Bigg]
\end{aligned}
\end{equation}

Before discussing the selected studies that have utilized Expected SARSA, a general overview of the algorithm can be found in Alg. \ref{alg:Expected_Sarsa}.


\begin{algorithm}[t]
\caption{Expected SARSA}
\begin{algorithmic}[1]
\State \textbf{Input:} policy \(\pi\), positive integer \(num\_episodes\), small positive fraction \(\alpha\), GLIE \(\{\epsilon_i\}\)
\State \textbf{Output:} value function \(Q\) (\(q_\pi\) if \(num\_episodes\) is large enough)
\State Initialize \(Q\) arbitrarily (e.g., \(Q(s, a) = 0\) for all \(s \in S\) and \(a \in A(s)\), and \(Q(\text{terminal-state}, \cdot) = 0\))
\For{\(i \gets 1\) to \(num\_episodes\) do}
    \State \(\epsilon \gets \epsilon_i\)
    \State Observe \(S_0\)
    \State \(t \gets 0\)
    \Repeat
        \State \parbox[t]{\dimexpr\linewidth-\algorithmicindent}{ Choose action \(A_t\) using policy derived from \\
        \(Q\) (e.g., \(\epsilon\)-greedy)\strut}
        \State Take action \(A_t\) and observe \(R_{t+1}, S_{t+1}\)
        \State \parbox[t]{\dimexpr\linewidth-\algorithmicindent}{
            \resizebox{0.4\textwidth}{!}{\(Q(S_t, A_t) \gets Q(S_t, A_t) \\
            + \alpha \left( R_{t+1} + \gamma \sum_a \pi(a|S_{t+1}) Q(S_{t+1}, a)
            - Q(S_t, A_t) \right)\)\strut}}
        \State \(t \gets t + 1\)
    \Until \(S_t\) is terminal
\EndFor
\State \textbf{return} \(Q\)
\end{algorithmic}
\label{alg:Expected_Sarsa}
\end{algorithm}

Authors in \cite{moradimaryamnegari2022model} designed a comprehensive exploration of combining Model Predictive Control (MPC) with the Expected SARSA algorithm to tune MPC models' parameters. This integration aimed to enhance the robustness and efficiency of control systems, particularly in applications where the system model's parameters were not fully known or subject to change. A key advantage was the innovative approach to speeding up learning by directly integrating RL with MPC, reducing the episodes needed for effective training. This efficiency was achieved using the Expected SARSA algorithm, which offered smoother convergence and better performance due to its average-based update rule compared to the more common Q-learning method, which focused on maximum expected rewards and might lead to higher variance in updates. However, the approach's complexity and the need for precise tuning of MPC model parameters represented significant challenges. The computational demands increased due to the dual needs of continuous adaptation by the RL algorithm and the rigorous constraints enforced by MPC. While the framework showed potential in simulations, its real-world applicability, especially in highly dynamic and unpredictable environments, required further validation.

Authors in \cite{gonzalez2023comparison} focused on improving Deterministic and Synchronous Multichannel Extension (DSME) networks' resilience against WiFi interference using the Expected SARSA algorithm. It evaluated channel adaptation and hopping strategies to mitigate interference in industrial environments, providing a detailed analysis with Expected SARSA. The study stood out for its rigorous simulation-based evaluation of interference mitigation strategies in a controlled DSME network. By using Expected SARSA, the research effectively reduced uncertainty in channel quality assessment, leading to more reliable network performance under interference conditions, crucial for industrial applications where reliable data transmission was vital for operational efficiency and safety. However, the complexity of RL implementation and its dependency on accurate real-time data posed challenges. The computational demands of running Expected SARSA in real-time environments could limit the practical deployment of this strategy in resource-constrained settings.

In \cite{muduli2023application}, researchers investigated using the Expected SARSA learning algorithm to manage Load Frequency Control in multi-area power systems. The study focused on enhancing power systems' stability integrated with Distributed Feed-in Generation based wind power, using a Model-free RL approach to adjust to variable power supply conditions without a predefined system model. The primary strength lies in its innovative approach to addressing challenges in integrating renewable energy sources into the power grid. Expected SARSA, an on-policy RL algorithm, demonstrated how to manage and stabilize frequency variations due to unpredictable renewable energy outputs and fluctuating demand adaptively. This was crucial for maintaining reliability and efficiency in power grids increasingly incorporating variable renewable energy sources. However, the approach's downside was related to the RL algorithm's computational demands and the need for extensive simulation and testing to fine-tune system parameters. Implementing such a system in a real-world setting could be constrained by these factors, particularly in terms of real-time computation capabilities and the scalability of the solution to larger, more complex grid systems.

Authors in \cite{oh2020reinforcement} utilized the Expected SARSA algorithm to optimize Energy Storage Systems (ESS) operation in managing uncertainties in wind power generation forecasts. The study modeled the problem as MDP where the state and action spaces were defined by the ESS's operational constraints. Expected SARSA’s primary result in this context was its superior performance compared to conventional Q-learning-based methods. The algorithm effectively handled the wide variance in wind power forecasts, crucial for optimizing ESS's charging and discharging actions to reduce forecast errors. The strategy's effectiveness was underscored by its near-optimal performance, closely approximating the optimal solution with complete future information. Simulation results demonstrated that the Expected SARSA-based strategy could manage wind power forecast uncertainty more effectively by adapting to varying conditions. This adaptability was enhanced by including frequency-domain data clustering, which refined the learning process and reduced input data variability, further improving the RL model’s performance. 
The last variant of SARSA is N-step SARSA. We will cover it in the next subsection before delving deeper into Approximation Model-free algorithms.

\paragraph{N-step SARSA}

With foundational concepts of n-step TD and bootstrapping established, we can expand on these ideas and explore research papers utilizing them. Let’s begin with N-step SARSA, an enhancement of the conventional one-step SARSA within the on-policy learning framework. In N-step SARSA, as shown in Alg. \ref{alg:n_step_Sarsa}, the update to the value (or action-value) is not limited to just the next state and action, as seen in one-step SARSA, but instead incorporates a sequence of n actions and rewards. The value function in the equation is replaced with \(Q(S_{t+n},A_{t+n})\):

\begin{equation}
\begin{aligned}
Q(S_t,A_t) \leftarrow &\, Q(S_t,A_t)
\alpha \Bigg[ \sum_{k=0}^{n-1} \gamma^k R_{t+k+1} \\
&\quad + \gamma^n Q(S_{t+n},A_{t+n}) - Q(S_t,A_t) \Bigg]
\end{aligned}
\end{equation}

N-step SARSA maintained the on-policy characteristic of SARSA, meaning the policy generating the behavior was the same as the one being evaluated and improved. This allowed it to effectively integrate the benefits of TD learning and the broader horizon considered in MC methods, balancing the bias-variance trade-off by adjusting the number of steps \(n\) looked ahead. N-step methods, including N-step SARSA, enhance learning by providing more robust estimates that incorporate multiple future outcomes rather than relying solely on the immediate next state and action. This leads to faster learning and improves policy performance, especially in complex environments where future rewards are significantly affected by actions taken over multiple steps \cite{sutton2018reinforcement, de2018multi}.

\begin{figure}[htb]
\begin{algorithm}[H]
\caption{N-step SARSA}
\begin{algorithmic}[1]
\State Initialize \(Q(s, a)\) arbitrarily, for all \(s \in S, a \in A\)
\State Initialize \(\pi\) to be \(\epsilon\)-greedy with respect to \(Q\), or to a fixed given policy
\State Algorithm parameters: step size \(\alpha \in (0, 1]\), small \(\epsilon > 0\), a positive integer \(n\)
\State All store and access operations (for \(S_t, A_t, R_t\)) can take their index mod \(n + 1\)
\Repeat \Comment{(for each episode)}
    \State Initialize and store \(S_0 \neq\) terminal
    \State Select and store an action \(A_0 \sim \pi(\cdot|S_0)\)
    \State \(T \gets \infty\)
    \For{\(t = 0, 1, 2, \dots\)}
        \If{\(t < T\):}
            \State Take action \(A_t\)
            \State \parbox[t]{\dimexpr\linewidth-\algorithmicindent}{ Observe and store the next reward as\\
            \(R_{t+1}\)
            and the next state as \(S_{t+1}\)\strut}
            \If{\(S_{t+1}\) is terminal }
                \State \(T \gets t + 1\)
            \Else
                \State \parbox[t]{\dimexpr\linewidth-\algorithmicindent}{ Select and store an action \(A_{t+1} \\
                \sim \pi(\cdot|S_{t+1})\)\strut}
            \EndIf
        \EndIf
        \State \parbox[t]{\dimexpr\linewidth-\algorithmicindent}{ \(\tau \gets t - n + 1\) \(\quad (\tau\) is the time whose \\
        estimate is being updated\() \)\strut}
        \If{\(\tau \geq 0\):}
            \State \parbox[t]{\dimexpr\linewidth-\algorithmicindent}{%
                \(G \gets \sum_{i = \tau + 1}^{\min(\tau + n, T)} \gamma^{i - \tau - 1} R_i\)\strut}
            \If{\(\tau + n < T\):}
                \State \parbox[t]{\dimexpr\linewidth-\algorithmicindent}{%
                    \(G \gets G + \gamma^n Q(S_{\tau + n}, A_{\tau + n})\) \\
                    \(\quad (G_{\tau:\tau+n})\)\strut}
                \EndIf
            \State \parbox[t]{\dimexpr\linewidth-\algorithmicindent}{%
                \(Q(S_\tau, A_\tau) \gets Q(S_\tau, A_\tau) \\
                + \alpha \left[ G - Q(S_\tau, A_\tau) \right]\)\strut}
            \If{\(\pi\) is being learned}
                \State \parbox[t]{\dimexpr\linewidth-\algorithmicindent}{ Ensure that \(\pi(\cdot|S_\tau)\) is \(\epsilon\)-greedy \\
                with respect to \(Q\)\strut}
            \EndIf
        \EndIf
    \EndFor
\Until{\(\tau = T - 1\)}
\end{algorithmic}
\label{alg:n_step_Sarsa}
\end{algorithm}
\end{figure}

Starting papers analysis, \cite{kekuda2021reinforcement} examined the application of the n-step SARSA RL algorithm to optimize traffic signal control. This method dynamically adjusted traffic signals based on real-time conditions, aiming to reduce congestion more effectively than traditional methods like Static Signaling (SS) and Longest Queue First. The paper innovatively applied the n-step SARSA, incorporating multiple future steps into decision-making. This enabled more strategic planning and could lead to better handling of complex traffic situations. The use of the Simulation of Urban MObility (SUMO) traffic simulator to model real-world traffic in Texas provided a solid base for testing and validating the algorithm. A detailed comparative analysis with existing methods showed potential improvements in managing traffic flow and reducing congestion. The research also tackled scalability by employing a centralized control agent, mitigating rapid growth in state-action space seen in decentralized systems, and enhancing feasibility for large urban areas. Despite benefits, real-world implementation could present challenges, including high computational demands and the need for real-time data processing and communication infrastructure. While simulations were crucial for preliminary tests, there was a risk that the model might be over-fitted to these conditions, potentially impairing the algorithm’s effectiveness in real settings. Additionally, the study focused on a particular urban environment and did not extensively investigate the algorithm's performance across diverse traffic patterns, different urban layouts, or during unusual events like accidents or road closures.

In \cite{kuchibhotla2020n}, authors introduced a hybrid method that melded on-policy traits of SARSA with off-policy features of Q-learning, integrating an N-step look-ahead capability to enhance foresight in decision-making. The algorithm's flexibility to adjust the number of look-ahead steps (N) dynamically allowed it to suit different learning stages, providing an effective balance between exploration and exploitation. This adaptability could facilitate more efficient learning than traditional RL methods. By combining on-policy and off-policy updates, the algorithm capitalized on the strengths of both SARSA and Q-learning, potentially decreasing the overestimation bias seen in Q-learning while still targeting optimal policies. The approach's ability to adjust the N parameter based on learning phases or specific conditions suggested customization for a broad array of applications, from simple gaming scenarios to intricate decision-making tasks in real-world contexts. However, the algorithm's requirement to finely adjust various hyperparameters, like maximum and minimum values of N and the breakpoint for decrementing N, added complexity to its configuration and optimization. This could pose a challenge, particularly for users with limited RL experience. Additionally, maintaining and updating a combination of policies over multiple steps before reaching a decision resulted in higher computational demands, potentially restricting the algorithm’s applicability in settings with limited resources or where rapid response times were essential.
The categorization of the examined papers by their domain is given in Table \ref{tab:SARSA_Papers}.

\begin{table}[t]
\centering
\renewcommand{\arraystretch}{1.2} 
\caption{SARSA Papers Review }
\begin{tabular}{|>{\raggedright\arraybackslash}p{4cm}|>{\raggedright\arraybackslash}p{3cm}|}
\hline
\textbf{Application Domain} & \textbf{Number of Papers} \\
\hline
Multi-agent Systems and Autonomous Behaviors (Shepherding, Virtual Agents) & \cite{go2016reinforcement}  \\
\hline
Games and Simulations (Real-Time Strategy Games) & \cite{wender2012applying} \\
\hline
Energy and Power Management (IoT Networks, Smart Energy Systems) & \cite{jaiswal2020green}, \cite{wang2020peer} \\
\hline
Cloud-based Control and Encryption Systems & \cite{suh2021sarsa}  \\
\hline
Swarm Intelligence and Optimization Problems &  \cite{iima2008swarm} \\
\hline
Transportation and Routing Optimization (EVs) & \cite{aljohani2022real} \\
\hline
MPC Tuning & \cite{moradimaryamnegari2022model} \\
\hline
Network Resilience and Optimization & \cite{gonzalez2023comparison}, \cite{oh2020reinforcement} \\
\hline
Power Systems and Energy Management & \cite{muduli2023application}\\
\hline
Network Optimization (Optical Transport Networks, Fog
RAN) & \cite{nassar2019reinforcement} \\
\hline
Intelligent Traffic Signal Control &\cite{kekuda2021reinforcement} \\
\hline
Hybrid RL Algorithms & \cite{kuchibhotla2020n} \\
\hline
\end{tabular}
\label{tab:SARSA_Papers}
\end{table}

\subsubsection{Summary of Tabular Model-free Algorithms}
In this section, we delved into Tabular Model-free algorithms as one of the categories in RL. After providing a brief overview of each algorithm and studies that used those methods, it is time to give a complete summary in Table \ref{table:TMF}.

\begin{table*}[t]
\renewcommand{\arraystretch}{1.2} 
\centering
\caption{Comparison of Tabular Model-free Algorithms}
\begin{adjustbox}{max width=\textwidth} 
\begin{tabular}{|>{\raggedright\arraybackslash}p{4cm}|>{\centering\arraybackslash}p{3cm}|>{\centering\arraybackslash}p{3cm}|>{\centering\arraybackslash}p{3cm}|>{\raggedright\arraybackslash}p{5cm}|}
\hline
\textbf{Algorithm} & \textbf{On-Policy/ Off-Policy} & \textbf{Scalability} & \textbf{Sample-Efficiency} & \textbf{Additional Information} \\
\hline
TD Learning & On-Policy & High & Moderate & Online learning, bootstrapping \\
\hline
TD(0)-Replay Algorithm & Off-Policy & High & Moderate & Uses experience replay \\
\hline
TD($\lambda$) & On-Policy & High & Moderate & Trace decay parameter $\lambda$ \\
\hline
N-step Bootstrapping & On-Policy & High & Moderate & Generalization of TD \\
\hline
N-step TD Prediction & On-Policy & High & Moderate & Predictive, bootstrapping \\
\hline
N-step Off-Policy Learning & Off-Policy & High & Moderate & Uses off-policy returns \\
\hline
Q-learning & Off-Policy & High & Moderate & Finds optimal policy, bootstrapping \\
\hline
Double Q-learning & Off-Policy & High & Moderate & Reduces overestimation bias \\
\hline
SARSA & On-Policy & High & Moderate & Learns the action-value function \\
\hline
Expected SARSA & On-Policy & High & Moderate & Uses expected value of next state \\
\hline
N-step SARSA & On-Policy & High & Moderate & Generalization of SARSA \\
\hline
MC Methods & On-Policy & Moderate & Low & Does not bootstrap, needs full episodes \\
\hline
On-Policy MC & On-Policy & Moderate & Low & Requires exploration starts \\
\hline
MC Importance Sampling & Off-Policy & Moderate & Low & Corrects for different policies \\
\hline
\end{tabular}
\end{adjustbox}
\label{table:TMF}
\end{table*}

It is necessary to explain what scalability and sample efficiency mean. \textbf{Scalability} refers to the algorithm's ability to handle varying sizes of environments or state spaces. \textbf{High} scalability indicates that the algorithm remains efficient even in larger environments or state spaces. \textbf{Moderate} scalability means that the algorithm can handle medium-sized problems effectively. \textbf{Low} scalability suggests that the algorithm struggles with large state spaces, often due to computational complexity or memory requirements.
\textbf{Sample-efficiency} reflects how effectively the algorithm learns from the available data. \textbf{High} sample-efficiency denotes that the algorithm can learn effectively from fewer samples. \textbf{Moderate} sample-efficiency indicates a requirement for a moderate number of samples. \textbf{Low} sample efficiency suggests that the algorithm needs a large number of samples to learn effectively.
\textbf{Bootstrapping} involves using estimates to update values, accelerating the learning process.
\textbf{Eligibility Traces} track state-action pairs to improve learning efficiency by integrating information over multiple steps.
\textbf{Experience Replay} stores past experiences, which can be reused during learning to improve performance.
\textbf{Exploration Starts} ensure that all states are visited by starting from random states, promoting thorough exploration of the environment.

In the next section, we start analyzing another paradigm of Model-free algorithms, approximation-based algorithms, before analyzing part II of Tabular Model-based ones.

\subsection{Approximation Model-free Algorithms}

In this section, we analyze Approximation Model-free algorithm variations and their applications of in various domains. It must be noted we assume that readers have knowledge in DL before reading this section, and readers are referred to \cite{lecun2015deep, goodfellow2016deep, kelleher2019deep, rusk2016deep} to understand and learn DL. Approximation Model-free algorithms in RL comprise methods oriented to learning policies and value functions solely from interaction with an environment, though without an explicitly stated model of the environment's dynamics. These algorithms typically use function approximators, such as neural networks, and generalize from observed state–action pairs to unknown ones. Consequently, they are quite effective at handling large or continuous states and actions \cite{ramirez2022model, degris2012model, liu2021policy}.

Key features of the Approximation Model-free algorithms are:

\begin{itemize}
    \item These algorithms learn the policy directly by optimizing expected reward without an explicit construction of the model of the environment.
    \item In value-function estimation, the value function is estimated using function approximators predicting expected rewards for states or state–action pairs.
    \item The solutions are scalable, thus fitting for problems with large or continuous state and action spaces since they can generalize from limited data.
\end{itemize}

Popular examples include Q-learning algorithms with function approximation, called DQN. Approximation Model-free algorithms are thus essential enablers of practical applications in RL where an explicit environment model is infeasible or too computationally expensive to be created.

As the first Approximation Model-free algorithm, over the next subsection, we will analyze DQN, one of the most widely used algorithms in RL.

\subsubsection{Deep Q-Networks (DQN)}
\label{Sec_DQN}

\textbf{DQN} algorithm merges Q-learning \cite{watkins1992q} with Neural Networks to learn control policies directly from raw pixel inputs. It uses Convolutional Neural Networks (CNN) to process these inputs and an experience replay mechanism to stabilize learning by breaking correlations between consecutive experiences. The target network, updated less frequently, aids in stabilizing training. DQN achieved state-of-the-art performance on various Atari 2600 games, surpassing previous methods and, in some cases, human experts, using a consistent network architecture and hyperparameters across different games \cite{mnih2013playing}.
DQN combines the introduced Bellman Equation with DL approaches like Loss Function and Gradient Descent to find the optimal policy as below:

\textbf{Loss Function}
\begin{equation}
L_i(\theta_i) = \mathbb{E}_{(s,a,r,s') \sim D} \left[ \left( y_i - Q(s, a; \theta_i) \right)^2 \right]
\end{equation}
where
\begin{equation}
y_i = r + \gamma \max_{a'} Q(s', a'; \theta^-)
\end{equation}

\textbf{Gradient Descent Step}
\begin{equation}
\begin{aligned}
\nabla_{\theta_i} L_i(\theta_i) = \mathbb{E}_{(s,a,r,s') \sim D} \Big[ & \left( r + \gamma \max_{a'} Q(s', a'; \theta^- ) \right. \\
& \left. - Q(s, a; \theta_i) \right) \nabla_{\theta_i} Q(s, a; \theta_i) \Big]
\end{aligned}
\end{equation}

Alg. \ref{alg:Deep_Q_learning} details the complete Deep Q-learning algorithm, regardless of using CNNs, as per the first paper. This algorithm is a generalization to have a general overview of the algorithm.


\begin{algorithm}[t]
\caption{Deep Q-learning}
\begin{algorithmic}[1]
\State Initialize replay memory \(\mathcal{D}\) to capacity \(N\) and action-value function \(Q\) with random weights
\For{episode \(= 1\) to \(M\)}
    \State \parbox[t]{\dimexpr\linewidth-\algorithmicindent}{ Initialize sequence \(s_1 = \{x_1\}\) and preprocessed sequence \(\phi_1 = \phi(s_1)\)\strut}
    \For{\(t = 1\) to \(T\)}
        \State \parbox[t]{\dimexpr\linewidth-\algorithmicindent}{%
            With probability \(\epsilon\), select a random action \\
            \(a_t\)\strut}
        \State \parbox[t]{\dimexpr\linewidth-\algorithmicindent}{%
            otherwise select \(a_t = \max_a Q^*(\phi(s_t), a; \theta)\)\strut}
        \State \parbox[t]{\dimexpr\linewidth-\algorithmicindent}{%
            Execute action \(a_t\) in emulator and observe \\
            reward \(r_t\) and image \(x_{t+1}\)\strut}
        \State \parbox[t]{\dimexpr\linewidth-\algorithmicindent}{%
            Set \(s_{t+1} = s_t, a_t, x_{t+1}\) and preprocess \\
            \(\phi_{t+1} = \phi(s_{t+1})\)\strut}
        \State \parbox[t]{\dimexpr\linewidth-\algorithmicindent}{%
            Store transition \((\phi_t, a_t, r_t, \phi_{t+1})\) in \(\mathcal{D}\)\strut}
        \State \parbox[t]{\dimexpr\linewidth-\algorithmicindent}{%
        Sample random minibatch of \\
        transitions \((\phi_j, a_j, r_j, \phi_{j+1})\) from \(\mathcal{D}\)\strut}
        \State \resizebox{0.8\columnwidth}{!}{$
            y_j = \begin{cases}
                r_j & \text{for terminal } \phi_{j+1} \\
                r_j + \gamma \max_{a'} Q(\phi_{j+1}, a'; \theta) & \text{for non-terminal } \phi_{j+1}
            \end{cases}
        $}
        \State \parbox[t]{\dimexpr\linewidth-\algorithmicindent}{%
        Perform a gradient descent step on \\
        \((y_j - Q(\phi_j, a_j; \theta))^2\)\strut}
    \EndFor
\EndFor
\end{algorithmic}
\label{alg:Deep_Q_learning}
\end{algorithm}

After building a solid foundation of the algorithm, it is time to analyze the papers in the literature.
In \cite{ahn2020application}, authors explored the use of DQN for controlling Heating, Ventilation, and Air Conditioning (HVAC) systems in an energy-efficient manner without relying on simulation models. The authors applied DQN to a reference office building, simulating its performance with an EnergyPlus model, aiming to minimize energy use while maintaining indoor CO2 concentrations below 1,000 ppm. One of the key strengths of this study was its demonstration of how DQN could improve HVAC control by learning from previous actions, states, and rewards, thus offering a Model-free optimization approach. The paper reported a significant reduction in total energy usage (15.7\%) compared to baseline operations, highlighting the potential of DQN to enhance energy efficiency in buildings. The research also addressed the complexity and interconnectivity of HVAC systems, providing a practical solution to a traditionally challenging control problem. However, the paper also had some limitations. The reliance on simulation data for training and testing the DQN might not have fully captured the intricacies and variations of real-world scenarios. Additionally, the study focused on a specific type of building and HVAC setup, which might have limited the generalizability of the results to other building types or climates. Furthermore, the study did not delve deeply into the potential challenges of implementing such a system in practice, such as the need for extensive data collection and processing capabilities.

An innovative application of DQN to localize brain tumors in Magnetic Resonance Imaging (MRI) images was introduced in \cite{stember2022reinforcement}. The key strength of this study lay in its ability to generalize tumor localization with limited training data, addressing significant limitations of supervised DL which often required large annotated datasets and struggled with generalization. The authors demonstrated that the DQN approach achieved 70\% accuracy on a testing set with just 30 training images, significantly outperforming the supervised DL method that showed only 11\% accuracy due to over-fitting. This showcased the robustness of RL in handling small datasets and its potential for broader application in medical imaging. The use of a grid-world environment to define state-action spaces and a well-designed reward system further strengthened the methodology. A notable weakness, however, was the limitation to two-dimensional image slices, which might not have fully captured the complexities of three-dimensional medical imaging. Future work should have addressed this by extending the approach to 3D volumes and exploring more sophisticated techniques to improve stability and accuracy.

Authors in \cite{guo2022deep} presented a routing algorithm for enhancing the sustainability of Rechargeable Wireless Sensor Networks. The authors proposed an adaptive dual-mode routing approach that integrated multi-hop routing with direct upload routing, optimized through DQN. The strength of this paper lies in its innovative use of RL to dynamically adjust the routing mode based on the life expectancy of nodes, significantly improving the network's energy efficiency and lifespan. The simulation results demonstrated that the proposed algorithm achieved a correct routing mode selection rate of 95\% with limited network state information, showcasing its practical applicability and robustness. However, the paper did not fully address potential real-world challenges such as the computational overhead of implementing DQN in resource-constrained sensor nodes and the impact of network topology changes over time. Additionally, while the simulations showed promising results, practical deployment and testing in diverse environments would have been necessary to validate the generalizability of the approach.

Researchers in \cite{talaat2022effective} explored the application of DQN optimized using Particle Swarm Optimization (PSO) for resource allocation in fog computing environments, specifically tailored for healthcare applications. A notable strength of this study was its innovative combination of DQN and PSO, which effectively balanced resource allocation by reducing makespan and improving both average resource utilization and load balancing levels. The methodology leveraged real-time data to dynamically adjust resources, showcasing a practical application of DQN in a critical domain. However, the paper could have benefited from a more detailed discussion on the scalability of the proposed system and its performance under varying network conditions. Additionally, while the results were promising, they were primarily based on simulations, which might not have fully captured the complexities and unpredictability of real-world fog environments. Further validation in practical deployments would have been necessary to fully ascertain the efficacy of the proposed approach.

An adaptive power management strategy for Parallel Plug-in Hybrid EVs (PHEVs) using DQN was investigated in \cite{song2018power}. The strength of this paper lies in its practical application of DQN for real-time power distribution in PHEVs. The approach considered continuous state variables such as battery State of Charge (SOC), required power, vehicle speed, and remaining distance ratio, which enhanced the model's adaptability and precision in dynamic driving conditions. The DQN model successfully minimized fuel consumption while maintaining battery SOC within desired limits, showing a 6\% increase in fuel consumption compared to DP which was globally optimal but computationally impractical for real-time applications. However, the research was primarily based on simulations using the File Transfer Protocol (FTP)-72 driving cycle, which might not have fully captured the variability of real-world driving conditions. Additionally, the study focused on a specific type of PHEV and driving scenario, which might have limited the generalizability of the results. Further real-world testing and validation across different vehicle models and driving conditions were necessary to establish the robustness and practical applicability of the proposed strategy.

Authors in \cite{yoon2017deep} explored the application of DQN to create an AI for a visual fighting game. A significant strength of this study was its innovative reduction of the action space from 41 to 11 actions, which simplified the training process and enhanced the model's performance. The DQN architecture included convolutional and fully connected layers optimized for handling sequential frame inputs, effectively learning to perform complex combinations in a dynamic, competitive environment. However, a notable limitation was the reliance on a static opponent (None agent) during training, which might not have fully captured the complexities and adaptive behaviors of actual gameplay against diverse opponents. Additionally, the experiments were conducted in a controlled environment with specific hardware, potentially limiting the generalizability of the results to other setups or real-world gaming scenarios. Further work should have focused on testing against more dynamic and varied opponents to evaluate the robustness and adaptability of the AI.

Authors in \cite{lv2019path} presented a novel path-planning approach using an enhanced DQN combined with dense network structures. The strength of this work was its innovative policy of leveraging both depth and breadth of experience during different learning stages, which significantly accelerated the learning process. The introduction of a value evaluation network helped the model quickly grasp environmental rules, while the parallel exploration structure improved the accuracy by expanding the experience pool. The use of dense connections further enhanced feature propagation and reuse, contributing to improved learning efficiency and path planning success. However, the primary limitation was that the experiments were conducted in a controlled grid environment with specific sizes (5x5 and 8x8), which might not have fully represented the complexities of real-world scenarios. Additionally, the reliance on a fixed maximum number of steps could potentially have led to suboptimal policy evaluations in dynamic and larger environments. Future work should have focused on validating this approach in more diverse and scalable settings to assess its generalizability and robustness.

The use of DQN for the real-time control of ESS co-located with renewable energy generators was discussed in \cite{zamzam2019energy}. A key strength of this work was its Model-free approach, which did not rely on distributional assumptions for renewable energy generation or real-time prices. This flexibility allowed the DQN to learn optimal policies directly from interaction with the environment. The simulation results demonstrated that the DQN-based control policy achieved near-optimal performance, effectively balancing energy storage management tasks like charging and discharging without violating operational constraints. However, the primary limitation was the reliance on simulated data for both training and evaluation, which might not have captured all the complexities of real-world energy systems. Additionally, the approach assumed the availability of significant computational resources, which might not have been feasible for all consumers. Future work should have focused on real-world implementations and considered the computational constraints of practical deployment environments.

Study \cite{gao2019anti} presented a novel approach to countering intelligent Unmanned Aerial Vehicle (UAV) jamming attacks using a Stackelberg dynamic game framework. The UAV jammer, acting as the leader, used Deep Recurrent Q-Networks (DRQN) to optimize its jamming trajectory, while ground users, as followers, employed DQN to find optimal communication trajectories to evade the jamming. The strength of this paper was its comprehensive modeling of the UAV jamming problem using DRQN and DQN, which effectively handled the dynamic and partially observable nature of the environment. The approach proved effective in simulations, showing that both the UAV jammer and ground users could achieve optimal trajectories that maximized their respective long-term cumulative rewards. The Stackelberg equilibrium ensured that the proposed strategies were stable and effective in a competitive environment. However, the primary limitation was the complexity and computational demands of implementing DRQN and DQN in real-time scenarios, which might have been challenging in practical deployments. Additionally, the simulations were based on specific scenarios and parameters, which might not have fully captured the variability of real-world environments. Further validation through real-world experiments and a more extensive range of scenarios would have been necessary to confirm the robustness and scalability of the proposed approach.

Table \ref{tab:DQN_Papers} provides an overview of the examined papers in DQN and their applications across different domains.
Over the next paragraphs, we will cover the Double Deep Q-Networks (DDQN), an extension of the DQN designed to address the overestimation bias observed in Q-learning.

\begin{table}[t]
\centering
\renewcommand{\arraystretch}{1.2} 
\caption{DQN Papers Review}
\begin{tabular}{|>{\raggedright\arraybackslash}p{4cm}|>{\raggedright\arraybackslash}p{3cm}|}
\hline
\textbf{Application Domain} & \textbf{References} \\
\hline
General RL (Policy learning, raw experience) & \cite{stember2022reinforcement}
  \\
\hline
Network Optimization & \cite{guo2022deep}
 \\
\hline
Swarm Intelligence and Optimization Problems & \cite{talaat2022effective}, 
 \\
 \hline
Network Optimization (Optical Transport Networks, Fog) &  \cite{song2018power}
 \\
\hline
Games and Simulations & \cite{yoon2017deep}, \cite{lv2019path} 
 \\
\hline
Security Games and Strategy Optimization & \cite{gao2019anti} \\
\hline
\end{tabular}
\label{tab:DQN_Papers}
\end{table}

\subsubsection{Double Deep Q-Networks (DDQN)}

The DDQN algorithm is an extension of the DQN designed to address the overestimation bias observed in Q-learning. It achieves this by decoupling the selection of the action from the evaluation of the Q-value, thus providing a more accurate estimation of action values \cite{van2016deep, peng2021end}.
Key contributions of DDQN can be summarized as follows:

\begin{itemize}
    \item \textbf{Overestimation Reduction}: By using two separate networks to select and evaluate actions, DDQN mitigates the overestimation bias that is prevalent in standard DQN.
    \item \textbf{Improved Stability and Performance}: The decoupling mechanism improves the stability and performance of the learning process, particularly in complex environments.
\end{itemize}

The Double Q-learning update rule is:
\begin{equation}
y = r + \gamma Q(s', \arg\max_{a'} Q(s', a'; \theta); \theta^-)
\end{equation}
where \( \theta \) are the parameters of the online network and \( \theta^- \) are the parameters of the target network. Alg. \ref{alg:DDQN}, provides a general overview of the DDQN algorithm.


%

    
    
    
    

\begin{algorithm}[t]
\caption{DDQN}
\begin{algorithmic}[1]
\State \textbf{Initialize:} Online network parameters \(\theta\), Target network parameters \(\theta^-\), Replay buffer \(\mathcal{D}\), Exploration rate \(\epsilon\), and Discount factor \(\gamma\)
\State \textbf{for each episode:}
\While{(not done)}
    \State Observe state \(s\)
    \State Select action \(a\) based on \(\epsilon\)-greedy policy:
    \State \parbox[t]{\dimexpr\linewidth-\algorithmicindent}{ \hspace{10mm} 
    \[
    a \gets 
    \begin{cases} 
    \text{random action} & \text{probability } \epsilon \\
    \arg\max_a Q(s, a; \theta) & \text{probability } 1 - \epsilon 
    \end{cases}
    \]\strut}
    \State \parbox[t]{\dimexpr\linewidth-\algorithmicindent}{ Execute action \(a\), observe reward \(r\) and new state \(s'\)\strut}
    \State Store transition \((s, a, r, s')\) in replay buffer \(\mathcal{D}\)
    
    \State \parbox[t]{\dimexpr\linewidth-\algorithmicindent}{ Sample a mini-batch of transitions \((s, a, r, s')\) from \(\mathcal{D}\)\strut}
    
    \State Compute target:
    \[
    y = r + \gamma Q(s', \arg\max_{a'} Q(s', a'; \theta); \theta^-)
    \]
    
    \State \parbox[t]{\dimexpr\linewidth-\algorithmicindent}{ Update the online network by minimizing \\
    the loss:
    \[
    L(\theta) = \mathbb{E}_{(s, a, r, s') \sim \mathcal{D}} \left[ \left( y - Q(s, a; \theta) \right)^2 \right]
    \]\strut}
    
    \State Every \(C\) steps, update the target network:
    \[
    \theta^- \gets \theta
    \]
\EndWhile
\end{algorithmic}
\label{alg:DDQN}
\end{algorithm}

Let us analyze the papers that used DDQN in the literature.
Authors in \cite{iqbal2021double} explored the application of DDQN for optimizing energy efficiency in Cloud Radio Access Networks (CRAN). A key strength of this work was its innovative use of DDQN to address the Q-value overestimation issue inherent in traditional DQN methods. By separating the action selection and evaluation processes, the DDQN framework achieved more accurate and stable policy learning. The simulation results showed that DDQN significantly enhanced energy efficiency compared to baseline solutions and standard DQN approaches, demonstrating up to 22\% power savings and a 20\% improvement in EE. However, the study's reliance on simulated environments and predefined parameters might not have fully reflected the complexities of real-world CRAN scenarios. The scalability and adaptability of the proposed method to diverse network conditions and dynamic user demands needed further validation through practical implementations. Additionally, while the study provided a robust theoretical framework, the practical deployment challenges, and computational overheads associated with DDQN in live networks required more in-depth exploration.

The application of DDQN to object detection tasks was investigated in \cite{zuo2017double}. One of the key strengths of this work was its innovative use of DDQN to enhance the accuracy and efficiency of object detection. By decoupling action selection and evaluation using two separate Q-networks, the DDQN approach addressed the overestimation problem associated with traditional DQN methods. This led to higher precision and recall rates, as demonstrated in their experiments. The method proved efficient, requiring fewer steps to detect objects and showing strong adaptability to different environments, including person detection scenarios. However, the study relied heavily on specific datasets and controlled environments, which might have limited the generalizability of the results. Real-world applications could have presented more complex scenarios that might have challenged the robustness of the proposed method. Further validation in diverse and uncontrolled settings would have helped ascertain the practical applicability and scalability of the DDQN-based object detection framework.

Authors in \cite{zhang2019double} presented an enhanced path-tracking approach for robotic vehicles using DDQN. A key strength of this work was its innovative application of DDQN for both path smoothing and tracking, which significantly reduced overshoot and settling time compared to traditional methods like Pure Pursuit Control. The proposed method demonstrated superior performance in navigating paths with sharp turns, making it particularly suitable for agricultural applications where precise path following was critical. The use of a simulation environment for training and subsequent testing on a real rover enhanced the robustness of the developed algorithm. However, the reliance on a specific rover model and controlled environments might have limited the generalizability of the results. The study could have benefited from further validation in diverse real-world conditions and with different types of robotic vehicles to fully establish the robustness and adaptability of the proposed method. Additionally, the computational demands of implementing DDQN in real-time applications might have posed practical challenges that needed addressing.

A novel method for improving tactical decision-making in autonomous driving using DDQN enhanced with spatial and channel attention mechanisms was introduced in \cite{zhang2021tactical}. The strength of this paper lies in its innovative use of a hierarchical control structure that integrated spatial and channel attention modules to better encode the relative importance of different surrounding vehicles and their features. This approach allowed for more accurate and efficient decision-making, as evidenced by the significant improvement in safety rates (54\%) and average exploration distance (30\%) in simulated environments. The combination of the algorithm with double attention enhanced the agent's ability to make intelligent and safe tactical decisions, outperforming baseline models in terms of both reward and capability metrics. However, the reliance on simulation-based testing limited the assessment of the model's performance in real-world driving scenarios, which might have presented additional complexities and unpredictabilities. The computational overhead associated with the attention modules and the utilized algorithm could also have posed challenges for real-time implementation in autonomous vehicles. Further validation in diverse and realistic environments would have been essential to confirm the robustness and scalability of this approach.

Study \cite{li2023double} explored a novel approach to solving the distributed heterogeneous hybrid flow-shop scheduling problem with multiple priorities of jobs using a DDQN-based co-evolutionary algorithm. A key strength of this work was its comprehensive framework that integrated global and local searches to balance computational resources effectively. The proposed DDQN-based co-evolutionary algorithm showed significant improvements in minimizing total weighted tardiness and total energy consumption by efficiently selecting operators and accelerating convergence. The numerical experiments and comparisons with state-of-the-art algorithms demonstrated the superior performance of the proposed method, particularly in handling real-world scenarios and large-scale instances. However, the study primarily relied on simulations and controlled experiments, which might not have fully captured the complexities and variability of actual manufacturing environments. Additionally, the computational overhead associated with training the DDQN model could have posed challenges for real-time applications. Further validation through real-world implementations and exploring dynamic events such as new job inserts and due date changes would have been necessary to fully establish the robustness and practical applicability of the proposed method.

A method for autonomous mobile robot navigation and collision avoidance using DDQN was developed in \cite{xue2019deep}. A significant strength of this work was its innovative use of DDQN to reduce reaction delay and improve training efficiency. The proposed method demonstrated superior performance in navigating robots to target positions without collisions, even in multi-obstacle scenarios. By employing a Kinect2 depth camera for obstacle detection and leveraging a well-designed reward function, the method achieved quick convergence and effective path planning, as evidenced by both simulation and real-world experiments on the Qbot2 robot platform. However, the reliance on a controlled laboratory environment and specific hardware (e.g., Optitrack system for positioning) might have limited the generalizability of the results. Real-world applications could have introduced additional complexities and variabilities that the study did not address. Further validation in diverse and uncontrolled environments, along with a discussion on the computational requirements and scalability of the approach, would have been necessary to fully establish its practical applicability.

Authors in \cite{mo2019decision} implemented an RL-based method for autonomous vehicle decision-making in overtaking scenarios with oncoming traffic. The study leveraged DDQN to manage both longitudinal speed and lane-changing decisions. A notable strength of this research was the use of DDQN with Prioritized Experience Replay, which accelerated policy convergence and enhanced decision-making precision. The simulation results in SUMO showed that the proposed method improved average speed, reduced time spent in the opposite lane, and lowered the overall overtaking duration compared to traditional methods, such as SUMO's default lane-change model. The RL-based approach demonstrated a high collision-free rate (98.5\%) and effectively mimicked human decision-making behavior, showcasing significant improvements in safety and efficiency. However, the reliance on simulated environments and specific scenarios might have limited the generalizability of the results to real-world applications. The study assumed complete observability of the state space, which might not have been realistic in actual driving conditions where sensor imperfections and uncertainties were common. Future work should have focused on addressing these limitations by exploring POMDPs and validating the approach in diverse, real-world scenarios.

A DDQN-based control method for obstacle avoidance in agricultural robots was introduced in\cite{yu2023obstacle}. A key strength of this work was its innovative use of DDQN to handle the complexities of dynamic obstacle avoidance in a structured farmland environment. The proposed method effectively integrated real-time data from sensors and used a neural network to decide the optimal actions, leading to significant improvements in space utilization and time efficiency compared to traditional risk index-based methods. The study reported high success rates in obstacle avoidance (98-99\%) and demonstrated the model's robustness through extensive simulations and field experiments. However, the reliance on predefined paths and the assumption of only dynamic obstacles appearing on these paths might have limited the approach's flexibility in more complex or unstructured environments. Additionally, the computational demands of the DDQN framework might have posed challenges for deployment in resource-constrained settings typical of agricultural machinery. Further validation in diverse real-world scenarios and exploration of more scalable solutions could have enhanced the practical applicability of this method.

A DDQN-based algorithm for global path planning of amphibious Unmanned Surface Vehicles (USVs) was studied in \cite{xiaofei2022global}. A major strength of this paper was its innovative use of DDQN for handling the complex path-planning requirements of amphibious USVs, which must navigate both water and air environments. The integration of electronic nautical charts and elevation maps to build a detailed 3D simulation environment enhanced the realism and accuracy of the path planning. The proposed method effectively balanced multiple objectives such as minimizing travel time and energy consumption, making it suitable for diverse scenarios including emergency rescue and long-distance cruising. However, the study primarily relied on simulated environments and predefined scenarios, which might not have fully captured the complexities of real-world applications. The computational demands of the DDQN framework could also have posed challenges for real-time implementation, especially in dynamic environments where quick decision-making was crucial. Further validation through practical deployments and testing in various real-world conditions would have been necessary to fully assess the robustness and scalability of the proposed approach.

Authors in \cite{lee2022intelligent} presented a Hand-over (HO) strategy for 5G networks. The proposed Intelligent Dual Active Protocol Stack (I-DAPS) HO used DDQN to enhance the reliability and throughput of handovers by predicting and avoiding radio link failures. A key strength of this paper was its innovative approach to leveraging DDQN for dynamic and proactive handover decisions in highly variable mmWave environments. The use of a learning-based framework allowed the system to make informed decisions based on historical signal data, thereby significantly Reducing Hand-over Failure (HOF) rates and achieving zero millisecond Mobility Interruption Time (MIT). The simulation results demonstrated substantial improvements in throughput and reliability over conventional handover schemes like Conditional Handover (CHO) and baseline Dual Active Protocol Stack (DAPS) HO. However, the study's primary reliance on simulations and specific urban scenarios might have limited the generalizability of the findings. Real-world implementations might have faced additional challenges such as varying environmental conditions and hardware limitations that were not fully addressed in the simulations. Further validation in diverse real-world environments would have been necessary to confirm the practical applicability and robustness of the proposed I-DAPS HO scheme.

Authors in \cite{zhang2018human} implemented an advanced vehicle speed control system using DDQN. The approach integrated high-dimensional video data and low-dimensional sensor data to construct a comprehensive driving environment, allowing the system to mimic human-like driving behaviors. A key strength of this work was its effective use of DDQN to address the instability issues found in Q-learning, resulting in more accurate value estimates and higher policy quality. The use of naturalistic driving data from the Shanghai Naturalistic Driving Study enhanced the model's realism and applicability. The system demonstrated substantial improvements in both value accuracy and policy performance, achieving a score that was 271.73\% higher than that of DQN. However, the reliance on pre-recorded driving data and controlled environments might have limited the generalizability of the results. Real-world driving conditions could have been significantly more variable, and the system's performance in such dynamic environments needed further validation. Additionally, the computational demands of DDQN might have posed challenges for real-time implementation in autonomous vehicles, necessitating further optimization for practical deployment.

\begin{table}[t]
\centering
\renewcommand{\arraystretch}{1.2} 
\caption{DDQN Papers Review }
\begin{tabular}{|>{\raggedright\arraybackslash}p{4cm}|>{\raggedright\arraybackslash}p{3cm}|}
\hline
\textbf{Application Domain} & \textbf{References} \\
\hline
Energy and Power Management (IoT Networks, Smart Energy Systems) & \cite{iqbal2021double},\cite{li2019partially}  \\
\hline
Multi-agent Systems and Autonomous Behaviors & \cite{zhang2021tactical}, \cite{mo2019decision}, \cite{xiaofei2022global} \\
\hline
Optimization & \cite{li2023double} \\
\hline
Real-time Systems and Hardware Implementations & \cite{lee2022intelligent} \\
\hline
Vehicle Speed Control System & \cite{zhang2018human} \\
\hline
Robotics & \cite{zhang2019double}, \cite{yu2023obstacle}, \cite{xue2019deep} \\
\hline
\end{tabular}
\label{tab:DDQN_Papers}
\end{table}

Authors in \cite{li2019partially} addressed the dynamic multiple-channel access problem in IoT networks with energy harvesting devices using the DDQN approach. The key strength of this study lies in its innovative application of DDQN to manage scheduling policies in POMDP. By converting the partial observations of scheduled nodes into belief states for all nodes, the proposed method effectively reduced energy costs and extended the network lifetime. The simulation results demonstrated that DDQN outperformed other RL algorithms, including Q-learning and DQN, in terms of average reward per time slot. However, the paper's reliance on simulations and specific parameters might have limited its applicability to real-world IoT environments. The assumptions regarding the Poisson process for energy arrival and the fixed transmission energy threshold might not have fully captured the complexities of practical deployments. Further validation through real-world experiments and consideration of diverse environmental conditions would have been necessary to fully establish the robustness and scalability of the proposed scheduling approach.

Table \ref{tab:DDQN_Papers} offers a comprehensive summary of the papers reviewed in this section, categorized by their respective domains within the DDQN. 
In the next subsection, another variant of DQN that uses the Dueling architecture is introduced.

\subsubsection{Dueling Deep Q-Networks}

The Dueling Deep Q-Networks (Dueling DQN) algorithm introduces a new neural network architecture that separates the representation of state values and advantages to improve learning efficiency and performance. This architecture allows the network to estimate the value of each state more robustly, leading to better policy evaluation and improved stability. The dueling architecture splits the Q-network into two streams, one for estimating the state value function and the other for the advantage function, which is then combined to produce the Q-values. By separately estimating the state value and advantage, the dueling network provides more informative gradients, leading to more efficient learning. This architecture demonstrates improved performance in various RL tasks, particularly in environments with many similar-valued actions \cite{wang2016dueling}.
The dueling network update rule is given as:
\begin{equation}
\begin{aligned}
Q(s, a; \theta, \alpha, \beta) &= V(s; \theta, \beta) + \\
& \left( A(s, a; \theta, \alpha) - \frac{1}{|A|} \sum_{a'} A(s, a'; \theta, \alpha) \right)
\end{aligned}
\end{equation}
where \( \theta \) are the parameters of the shared network, \( \alpha \) are the parameters of the advantage stream, and \( \beta \) are the parameters of the value stream.

By separately estimating the state value and advantage, the dueling network provides more robust estimates of state values. The decoupling mechanism enhances stability and overall performance in various environments, especially where actions have similar values. Alg. \ref{alg:dueling_dqn} gives a comprehensive overview of the Deling DQN algorithm.

Over the next few paragraphs, several research studies are examined in detail to understand the Dueling DQN algorithm better.
A DRL-based framework utilizing Dueling Double Deep Q-Networks with Prioritized Replay (DDDQNPR) to solve adaptive job shop scheduling problems was implemented in \cite{han2020research}. The main strength of this study was its innovative combination of DDDQNPR with a disjunctive graph model, transforming scheduling into a sequential decision-making process. This approach allowed the model to adapt to dynamic environments and achieve optimal scheduling policies through offline training. The proposed method outperformed traditional heuristic rules and genetic algorithms in both static and dynamic scheduling scenarios, showcasing a significant improvement in scheduling performance and generalization ability. However, the reliance on predefined benchmarks and controlled experiments might have limited the applicability of the results to real-world manufacturing environments. The complexity and computational demands of the DDDQNPR framework might have posed challenges for real-time implementation in large-scale production settings. Future work should have focused on validating the approach in diverse real-world scenarios and exploring more efficient reward functions and neural network structures to enhance the method's scalability and robustness.

Managing Device-to-Device (D2D) communication using Dueling DQN architecture was proposed by authors in \cite{ban2020autonomous}. A key strength of this paper was its effective utilization of the Dueling DQN architecture to autonomously determine transmission decisions in D2D networks, without relying on centralized infrastructure. The approach leveraged easily obtainable Channel State Information (CSI), allowing each D2D transmitter to train its neural network independently. The proposed method successfully mitigated co-channel interference, demonstrating near-optimal sum rates in low Signal-to-Noise Ratio (SNR) environments and outperforming traditional schemes like No Control, Opportunistic, and Suboptimal in terms of efficiency and complexity reduction. However, the study's reliance on simulation data and specific channel models might have limited the generalizability of the results to real-world D2D networks. The approach assumed ideal conditions such as perfect CSI and zero delay in TDD-based full duplex communications, which might not have held in practical scenarios. Future work should have focused on validating the method in diverse real-world environments and addressing potential implementation challenges such as real-time computational requirements and varying network conditions.

Authors explored an advanced power allocation algorithm in edge computing environments using Dueling DQN, in \cite{xuan2021power}. A significant strength of this study was its innovative use of Dueling DQN, which separated the value and advantage functions within the Q-network. This architecture enhanced the accuracy and efficiency of power allocation by focusing on state-value estimation and advantage learning. The proposed algorithm outperformed several baseline methods, including traditional DQN, Fractional Programming (FP), and Weighted Minimum Mean Square Error, in both simulated scenarios and different user densities. The results demonstrated substantial improvements in average downlink rates and reduced computational overhead, highlighting the algorithm's robustness and efficiency in dynamic network conditions. However, the paper relied on simulations with specific parameters and controlled environments, which might have limited its applicability to real-world scenarios. The assumptions regarding perfect synchronization and the static nature of edge servers might not have fully captured the complexities of practical deployments. Further real-world testing and validation were needed to confirm the scalability and practical implementation of the proposed Dueling DQN approach in diverse and dynamic edge computing environments.

A hybrid approach combining Dueling DQN and Double Elite Co-Evolutionary Genetic Algorithm (DECGA) for parameter estimation in variogram models used in geo-statistics and environmental engineering was implemented in \cite{liu2020application}. A notable strength of this study was its innovative integration of RL with evolutionary algorithms. The Dueling DQN enhanced the parameter setting of the genetic algorithm, improving the convergence speed and avoiding premature convergence. The DECGA was effective in finding global optimal solutions by maintaining a diverse population. This combined approach significantly improved the accuracy of parameter estimation in both single and nested variogram models, as demonstrated by the experimental results on heavy metal concentration datasets, showing lower Residual Sum of Squares values compared to traditional methods. However, the reliance on controlled datasets and predefined parameters might have limited the method's generalizability to more complex, real-world scenarios. The computational complexity introduced by combining Dueling DQN with DECGA could also have posed challenges for practical implementation, especially in scenarios requiring real-time parameter estimation. Future work should have focused on validating the approach in diverse real-world environments and optimizing the computational efficiency for broader applicability.

\begin{figure}[tb]
\centering
\begin{algorithm}[H]
\caption{Dueling DQN}
\begin{algorithmic}[1]
\State \textbf{Input:} Learning rate $lr$, batch size $batch$, discount factor $\gamma$, exploration rate $\epsilon$, decay rate $decay$, target network update frequency $freq$, episodes
\State \textbf{Output:} Trained Dueling Q-Network
\State Initialize online network weights $\theta$
\State Initialize target network $\theta^- = \theta$
\State Initialize replay memory $\mathcal{D}$
\For{episode $e = 1$ to $episodes$}
    \State Initialize state $s$
    \While{state $s$ is not terminal}
        \State Select action $a_t$ using $\epsilon$-greedy policy
        \State Execute $a_t$, observe $r_t, s_{t+1}$
        \State Store $(s_t, a_t, r_t, s_{t+1})$ in $\mathcal{D}$
        \State Sample minibatch from $\mathcal{D}$
        \State Compute targets $y_j$
        \State Update $\theta$ by minimizing loss $L(\theta)$
        \If{update step mod $freq = 0$}
            \State Update target network $\theta^- \gets \theta$
        \EndIf
        \State $s \gets s_{t+1}$
        \State Decay $\epsilon$
    \EndWhile
\EndFor
\end{algorithmic}
\label{alg:dueling_dqn}
\end{algorithm}
\vspace{-2em}
\end{figure}

Authors in \cite{liu2019green} presented a method for optimizing UAV flight paths in IoT sensor networks using Dueling DQN. The goal was to balance energy consumption and data delay. A significant strength of the study was its effective use of Dueling DQN to manage UAV paths, enhancing both energy efficiency and network performance. The approach utilized a grid-based method to model network states, reducing computational complexity. Simulation results showed notable improvements in power consumption and data transmission delays compared to existing methods. However, the reliance on simulated environments might have limited the applicability of the results to real-world scenarios, where sensor data and network conditions could have been more variable. Further validation through practical implementations and testing in diverse environments would have been essential to confirm the robustness and scalability of the proposed method. Additionally, the computational demands of Dueling DQN might have posed challenges for real-time applications in resource-constrained IoT networks.

An innovative method for optimizing the Age of Information (AoI) in vehicular fog systems using Dueling DQN was presented in \cite{tadele2024optimization}. A major strength of this work was its holistic approach to AoI optimization, considering both the transmission time from the information source to the monitor and from the monitor to the destination. The Dueling DQN algorithm effectively minimized the end-to-end AoI by leveraging edge computing and vehicular fog nodes to handle real-time data more efficiently. The proposed method demonstrated significant improvements in system performance and information freshness compared to DQN and analytical methods. However, the reliance on simulations to validate the approach might have limited the generalizability of the findings to real-world applications. The assumptions made regarding the processing capabilities and network conditions might not have fully captured the complexities of practical implementations. Further research involving real-world testing and validation in diverse environments was necessary to fully establish the robustness and scalability of the proposed method.

The authors in \cite{jiang2021research} presented an enhanced Dueling DQN method for improving the autonomous capabilities of UAVs in terms of obstacle avoidance and target tracking. One of the key strengths of this study was the effective integration of improved Dueling DQN with a target tracking mechanism, which allowed the UAV to make more precise and timely decisions. The improved algorithm showed superior performance in simulation tests, with significant improvements in obstacle avoidance accuracy and target tracking efficiency. The authors provided a comprehensive analysis of the algorithm's ability to maintain stability and adaptability in dynamic environments, showcasing its practical potential for real-world UAV applications. However, the study's dependence on simulation environments and controlled scenarios might not have fully captured the complexities and unpredictability of real-world applications. The effectiveness of the improved Dueling DQN algorithm in diverse and unstructured environments needed further validation through real-world testing. Additionally, the computational demands of the improved algorithm could have posed challenges for real-time processing and decision-making on resource-constrained UAV platforms. Future work should have addressed these aspects to establish the practicality and scalability of the proposed approach fully.

An advanced path planning approach for Unmanned Surface Vessels (USVs) using Dueling DQN enhanced with a tree sampling mechanism was implemented in \cite{huang2022usv}. A key strength of this work was the innovative combination of Dueling DQN with a tree-based priority sampling mechanism, which significantly improved the convergence speed and efficiency of the learning process. The approach leveraged the decomposition of the value function into state-value and advantage functions, enhancing the policy evaluation and decision-making accuracy. The simulation results demonstrated that the proposed method achieved faster convergence and more effective obstacle avoidance and path planning compared to DQN algorithms. Specifically, the Dueling DQN algorithm showed improved performance in terms of the number of steps and time required to reach the target across various test environments. However, the study's reliance on simulated static environments might have limited the generalizability of the results to real-world scenarios where the USVs would have encountered dynamic and unpredictable conditions. The paper also did not address the computational complexity introduced by the tree sampling mechanism, which could have posed challenges for real-time implementation on resource-constrained USVs. Future work should have focused on validating the approach in dynamic real-world environments and exploring ways to optimize the computational efficiency for practical applications. 
A summary of the analyzed papers is given in Table \ref{tab:Dueling_DQN_Papers}.
In the next subsection, another part of Tabular Model-based algorithms, Model-based Planning, is introduced.

\begin{table}[t]
\centering
\renewcommand{\arraystretch}{1.2} 
\caption{Dueling DQN Papers Review}
\begin{tabular}{|>{\raggedright\arraybackslash}p{4cm}|>{\raggedright\arraybackslash}p{3cm}|}
\hline
\textbf{Application Domain} & \textbf{References} \\
\hline
Energy and Power Management (IoT Networks, Smart Energy Systems) & \cite{xuan2021power}, \cite{liu2019green}  \\
\hline
Transportation and Routing Optimization & \cite{huang2022usv} \\
\hline
Swarm Intelligence and Optimization Problems & \cite{han2020research} \\
\hline
Network Optimization & \cite{ban2020autonomous}, \cite{tadele2024optimization}
 \\
\hline
Geo-statistics and Environmental Engineering & \cite{liu2020application} \\
\hline
Autonomous UAVs & \cite{xue2019deep}, \cite{jiang2021research} \\
\hline
Path planning approach for USVs & \cite{huang2022usv}
 \\
\hline
\end{tabular}
\label{tab:Dueling_DQN_Papers}
\end{table}

\subsection{Advanced Tabular Model-based Methods}
\label{Sec_TMB2}

In this subsection, we dive into the second part of Tabular Model-based methods. Key characteristics of Model-based algorithms include Model Representation, which uses transition probabilities \( P(s'|s,a) \) and reward functions \( R(s, a) \) to define state transitions and rewards; Planning and Policy Evaluation, which iteratively approximates the value function \( V(s) \) with the Bellman equation until convergence; and Value Iteration, which also iteratively refines \( V(s) \) using the Bellman equation to determine the optimal policy by evaluating future states and actions \cite{sutton2018reinforcement}.

After analyzing the first part of Tabular Model-based algorithms, DP approaches, we now shed lights on the second part, Model-based Planning methods.

\subsubsection{Model-based Planning}
\label{Sec_MBP}

Model-based Planning in RL refers to the approach where the agent builds and utilizes a model of the environment to plan its actions and make decisions. This model can be either learned from interactions with the environment or predefined if the environment's dynamics are known.
Model-based methods can be more sample-efficient as they leverage learned models of the environment to simulate many scenarios, allowing the agent to gain more insights without needing to interact directly with the environment each time. Model-based Planning enables the agent to balance exploration—trying out new actions—and exploitation—using known good actions more effectively. By simulating different actions and their outcomes, the agent can make more informed decisions, optimizing its behavior. The learned model enables the agent to adapt its behavior easily to environmental changes. By updating the model and re-planning, the agent can respond dynamically to new situations, maintaining its performance even in changing environments \cite{sutton2018reinforcement,  moerland2023model, silver2010monte}.

There are three main streams in Model-based Planning, Monte Carlo Tree Search (MCTS), Prioritized Sweeping, and Dyna-Q. We start by analyzing MCTS over the next paragraphs.

\paragraph{Monte Carlo Tree Search}

MCTS is a powerful algorithm used to solve sequential decision-making problems under uncertainty. Its application spans various domains, most notably in game AI, where it has achieved significant success. MCTS combines the precision of tree search with the generality of random sampling (MC simulation). This technique has been instrumental in the development of AI systems like AlphaGo and AlphaZero, which have demonstrated superhuman performance in games like Go, chess, and shogi. The term "MCTS" was first coined by \cite{coulom2006efficient} when it was applied it to the Go-playing program, Crazy Stone. Prior to this, game-playing algorithms primarily relied on deterministic search techniques, such as those used in chess. The significant challenge in games like Go, due to their vast state spaces, necessitated a more intelligent and efficient approach to search, leading to the adoption of MCTS.

MCTS operates by building a search tree incrementally and asymmetrically. Each iteration of MCTS involves four main steps \cite{sutton2018reinforcement}:
\begin{enumerate}
    \item Selection: Starting from the root node, the algorithm recursively selects child nodes until it reaches a leaf node (lines 14-17). This selection process balances exploration and exploitation, typically using a UCB formula to choose moves with high estimated value and uncertainty.
    \item Expansion: If the leaf node is not a terminal node, it is expanded by adding one or more child nodes (line 18). This represents exploring new moves or actions that have not yet been considered.
    \item Simulation: From the expanded node, a simulation (or playout) is run to the end of the game using a default policy, often random (19-21). This simulation provides an outcome that approximates the value of the moves from the expanded node.
    \item Backpropagation: The result of the simulation is propagated back up the tree, updating the values of all nodes along the path (line 22). This helps to refine the estimated value of moves based on new information.
\end{enumerate}
These steps are repeated until a computational budget (time or iterations) is exhausted. The move with the highest value at the root node is then chosen as the optimal move. Based on the overview presented in \cite{rossi2022monte}, Alg. \ref{alg:MCTS} provides a comprehensive overview of MCTS.

Key Advantages of MCTS are Domain Independence (MCTS does not require domain-specific knowledge, only the rules of the game or problem.), Scalability (It can handle very large state spaces by focusing computational effort on the most promising parts of the search tree.), and Flexibility (It can be adapted to various types of problems, including those with stochastic elements.). \cite{coulom2006efficient, sutton2018reinforcement, fu2020tutorial}

\begin{algorithm}[t]
\caption{MCTS}
\begin{algorithmic}[1]
\State \textbf{Input:} 
\State \hskip1.5em PD: problem’s dimension 
\State \hskip1.5em FG: fitness goal 
\State \hskip1.5em RL: roll-out length 
\State \textbf{Output:} 
\State \hskip1.5em n*: the best solution found
\State \hskip1.5em f: the maximum performance level reached 
\State \hskip1.5em t: the total number of roll-outs performed 

\State Create root node \(n_r\) and add it to the tree
\State Make \(n_r\) the actual node \(n\)
\State \(n^* \gets \text{null}\)
\State \(f \gets 0\)
\State \(t \gets 0\)

\While{move\_available(\(n\))}
    \While{not is\_leaf(\(n\))}
        \State \(n \gets \) best\_child(\(n\))
    \EndWhile
    \State \(n \gets \) add\_random\_child\_node(\(n\))
    \State RL \( \gets \) rand(PD, \(n\))
    \State \(f, n' \gets \) roll\_out(\(n\), RL)
    \State \(t \gets t + 1\)
    \State back\_propagate(\(n\), \(f\))
    \If{\(f > \text{FG}\)}
        \State \(n^* \gets n'\)
        \State \textbf{break}
    \EndIf
    \State \(n \gets n_r\)
\EndWhile

\State \textbf{return} \(n^*\), \(f\), \(t\)
\end{algorithmic}
\label{alg:MCTS}
\end{algorithm}

MCTS is a fundamental and important algorithm in the realm of RL. Thus, let us examine some studies that utilized this approach to solve different problems.
Study \cite{wang2012multi} presented the Multi-objective MCTS (MO-MCTS) algorithm, an extension of MCTS designed for multi-objective sequential decision-making. The key innovation lies in using the hyper-volume indicator to replace the UCB criterion, enabling the algorithm to handle multi-dimensional rewards and discover several Pareto-optimal policies within a single tree. The MO-MCTS algorithm stood out for its ability to effectively manage multi-objective optimization by integrating the hyper-volume indicator, which provided a comprehensive measure of the solution set's quality. This approach allowed the algorithm to balance exploration and exploitation in multi-dimensional spaces, making it more efficient in discovering Pareto-optimal solutions compared to traditional scalarized RL methods. The use of the hyper-volume indicator as an action selection criterion ensured that the algorithm could capture a diverse set of optimal policies, addressing the limitations of linear-scalarization methods that failed in non-convex regions of the Pareto front. The experimental validation on the Deep Sea Treasure (DST) problem and grid scheduling tasks demonstrated that MO-MCTS achieved superior performance and scalability, matching or surpassing state-of-the-art non-RL-based methods despite higher computational costs. However, the reliance on accurate computation of the hyper-volume indicator introduced significant computational overhead, particularly in high-dimensional objective spaces. The complexity of maintaining and updating the Pareto archive could also pose challenges, especially as the number of objectives increased. While the algorithm showed robust performance in deterministic settings, its scalability and efficiency in highly stochastic or dynamic environments remained to be fully explored. The need for domain-specific knowledge to define the reference point and other hyper-volume-related parameters could also limit the algorithm's generalizability. Additionally, the method's computational intensity might hinder its real-time applicability, requiring further optimization or approximation techniques to reduce processing time.

A novel approach to Neural Architecture Search (NAS) using MCTS was introduced in \cite{su2021prioritized}. The authors proposed a method that captured the dependencies among layers by modeling the search space as a Monte-Carlo Tree, enhancing the exploration-exploitation balance and efficiently storing intermediate results for better future decisions. The method was validated through experiments on the NAS-Bench-Macro benchmark and ImageNet dataset. The primary strength of this approach lies in its ability to incorporate dependencies between different layers during architecture search, which many existing NAS methods overlooked. By utilizing MCTS, the proposed method effectively balanced exploration and exploitation, leading to more efficient architecture sampling. The use of a Monte-Carlo Tree allowed for the storage of intermediate results, significantly improving the search efficiency and reducing the need for redundant computations. The introduction of node communication and hierarchical node selection techniques further refined the evaluation of numerous nodes, ensuring more accurate and efficient search processes. Experimental results demonstrated that the proposed method achieved higher performance and search efficiency compared to state-of-the-art NAS methods, particularly on the NAS-Bench-Macro and ImageNet datasets. On the other hand, the reliance on the accurate modeling of the search space as a Monte-Carlo Tree introduced additional complexity, particularly in terms of computational overhead and memory requirements. The method's performance heavily depended on the proper tuning of hyperparameters, such as the temperature term and reduction ratio, which could be challenging to optimize for different datasets and tasks. While the experiments showed promising results, further validation in more diverse and complex real-world scenarios was necessary to fully assess the scalability and robustness of the approach. The additional computational cost associated with maintaining and updating the Monte-Carlo Tree and performing hierarchical node selection could impact the method's applicability in real-time applications or resource-constrained environments.

Study \cite{zerbel2019multiagent} introduced Multi-agent MCTS (MAMCTS), a novel extension of MCTS tailored for cooperative multi-agent systems. The primary innovation was the integration of difference evaluations, which significantly enhanced coordination strategies among agents. The performance of MAMCTS was demonstrated in a multi-agent path-planning domain called Multi-agent Grid-world (MAG), showcasing substantial improvements over traditional reward evaluation methods. The MAMCTS approach effectively leveraged difference evaluations to prioritize actions that contributed positively to the overall system, leading to significant improvements in learning efficiency and coordination among agents. By combining MCTS with different rewards, the algorithm balanced exploration and exploitation, ensuring that agents could efficiently navigate the search space to find optimal policies. The experimental results in the 100x100 MAG environment demonstrated that MAMCTS outperformed both local and global reward methods, achieving up to 31.4\% and 88.9\% better performance, respectively. This superior performance was consistent across various agent and goal configurations, highlighting the scalability and robustness of the approach. The use of a structured search process and prioritized updates ensured that the algorithm could handle large-scale multi-agent environments effectively. The reliance on accurate computation and maintenance of difference evaluations introduced additional computational overhead, particularly in environments with a large number of agents and goals. The method's performance was sensitive to the accuracy of these evaluations, which might be challenging to maintain in highly dynamic or unpredictable environments. While the simulations provided strong evidence of the method's efficacy, further validation in more diverse real-world scenarios was necessary to fully assess its scalability and practical utility. The complexity of managing multiple agents and ensuring synchronized updates could also pose challenges, particularly in real-time applications where computational resources are limited. Additionally, the paper focused primarily on cooperative settings, and the applicability of MAMCTS to competitive or adversarial multi-agent environments remained an area for future exploration.

A MO-MCTS algorithm tailored for real-time games was developed in \cite{perez2014multiobjective}. It focused on balancing multiple objectives simultaneously, leveraging the hyper-volume indicator to replace the traditional UCB criterion. The algorithm was tested against single-objective MCTS and Non-dominated Sorting Genetic Algorithm (NSGA)-II, showcasing superior performance in benchmarks like DST and the Multi-objective Physical Traveling Salesman Problem (MO-PTSP). The MO-MCTS algorithm excelled in handling multi-dimensional reward structures, efficiently balancing exploration and exploitation. By incorporating the hyper-volume indicator, the algorithm could discover a diverse set of Pareto-optimal solutions, effectively addressing the limitations of linear-scalarization methods in non-convex regions of the Pareto front. The empirical results on DST and MO-PTSP benchmarks demonstrated the algorithm's ability to converge to optimal solutions quickly, outperforming both single-objective MCTS and NSGA-II in terms of exploration efficiency and solution quality. The use of a weighted sum and Euclidean distance mechanisms for action selection further enhanced the adaptability of MO-MCTS to various game scenarios, providing a robust framework for real-time decision-making. Nonetheless, the reliance on the hyper-volume indicator introduced significant computational overhead, which could limit the algorithm's scalability in high-dimensional objective spaces. The need for maintaining a Pareto archive and computing the hyper-volume for action selection added to the complexity, potentially impacting real-time performance. While the algorithm showed strong results in the tested benchmarks, its applicability to more complex and dynamic real-time games required further validation. The computational intensity of the approach might hinder its practicality in resource-constrained environments, necessitating the exploration of optimization techniques to reduce processing time without compromising performance. Additionally, the paper primarily focused on deterministic settings, leaving the performance in stochastic or highly variable environments less explored.

Researchers in \cite{baier2013monte} explored hybrid algorithms that integrated MCTS with mini-max search to leverage the strategic strengths of MCTS and the tactical precision of minimax. The authors proposed three hybrid approaches: employing minimax during the selection/expansion phase, the rollout phase, and the backpropagation phase of MCTS. These hybrids aimed to address the weaknesses of MCTS in tactical situations by incorporating shallow minimax searches within the MCTS framework. The hybrid algorithms presented in the paper offered a promising combination of MCTS's ability to handle large search spaces with mini-max's tactical accuracy. By integrating shallow mini-max searches, the hybrids could better navigate shallow traps that MCTS might overlook, leading to more robust decision-making in games with high tactical demands. The experimental results in games like Connect-4 and Breakthrough demonstrated that these hybrid approaches could outperform standard MCTS, particularly in environments where tactical precision was crucial. The use of mini-max in the selection/expansion phase and the backpropagation phase significantly improved the ability to avoid blunders and recognize winning strategies early, enhancing the overall efficiency and effectiveness of the search process. However, the inclusion of mini-max searches introduced additional computational overhead, which could slow down the overall search process, especially for deeper mini-max searches. The performance improvements were heavily dependent on the correct tuning of parameters, such as the depth of the mini-max search and the criteria for triggering these searches. While the hybrids showed improved performance in the tested games, their scalability and effectiveness in more complex and dynamic real-world scenarios remained to be fully validated. The reliance on game-specific characteristics, such as the presence of shallow traps, might limit the generalizability of the results. Further exploration was needed to assess the impact of these hybrids in a broader range of domains and under different conditions.

Authors in \cite{de2016monte} introduced the Option-MCTS (O-MCTS) algorithm, which extended the MCTS by incorporating high-level action sequences, or "options," aimed at achieving specific subgoals. The proposed algorithm aimed to enhance general video game playing by utilizing higher-level planning, enabling it to perform well across a diverse set of games from the General Video Game AI competition. Additionally, the paper introduced Option Learning MCTS (OL-MCTS), which applied a progressive widening technique to focus exploration on the most promising options. The integration of options into MCTS was a significant advancement, allowing the algorithm to plan more efficiently by considering higher-level strategies. This higher abstraction level helped the algorithm deal with complex games that required achieving multiple subgoals. The use of options reduced the branching factor in the search tree, enabling deeper exploration within the same computational budget. The empirical results demonstrated that O-MCTS outperformed traditional MCTS in games requiring sequential subgoal achievements, such as collecting keys to open doors, showcasing its strength in strategic planning. The introduction of OL-MCTS further improved performance by learning which options were most effective, thus focusing the search on more promising parts of the game tree and improving efficiency. On the other hand, the reliance on predefined options and their proper tuning could be a limitation, as the performance of O-MCTS heavily depended on the quality and relevance of these options to the specific games being played. The initial computational overhead associated with constructing and managing a large set of options might impact the algorithm's performance, particularly in games with numerous sprites and complex dynamics. The progressive widening technique in OL-MCTS, while beneficial for focusing exploration, introduced additional complexity and overhead, potentially reducing real-time applicability. Further validation was needed to assess the scalability and robustness of these algorithms in a wider range of real-world game scenarios, where the diversity and unpredictability of game mechanics might present new challenges.

An extension of MCTS, incorporating heuristic evaluations through implicit mini-max backups, was investigated in \cite{lanctot2014monte}. The approach aimed to combine the strengths of MCTS and mini-max search to improve decision-making in strategic games by maintaining separate estimations of win rates and heuristic evaluations and using these to guide simulations. The integration of implicit minimax backups within MCTS significantly enhanced the quality of simulations by leveraging heuristic evaluations to inform decision-making. This hybrid approach addressed the limitations of pure MCTS in domains where tactical precision was crucial, effectively balancing strategic exploration with tactical accuracy. By maintaining separate values for win rates and heuristic evaluations, the algorithm could better navigate complex game states, leading to stronger play performance. The empirical results in games like Kalah, Breakthrough, and Lines of Action demonstrated substantial improvements over standard MCTS, validating the effectiveness of implicit mini-max backups in diverse strategic environments. The method also showed robust performance across different parameter settings, highlighting its adaptability and potential for broader application in game-playing AI. However, the reliance on accurate heuristic evaluations introduced complexity in environments where such evaluations were not readily available or were difficult to compute. The additional computational overhead associated with maintaining and updating separate value estimates might impact the algorithm's efficiency, particularly in real-time applications. While the approach showed significant improvements in specific games, further validation was necessary to assess its scalability and robustness in more complex and dynamic scenarios. The dependence on domain-specific knowledge for heuristic evaluations might limit the generalizability of the method to a wider range of applications. Additionally, the complexity of tuning the parameter that weighted the influence of heuristic evaluations could pose challenges in optimizing the algorithm for different environments.

Authors in \cite{winands2010monte} explored the application of MCTS to the game of Lines of Action (LoA), a two-person zero-sum game known for its tactical depth and moderate branching factor. The authors proposed several enhancements to standard MCTS to handle the tactical complexities of LoA, including game-theoretical value proving, domain-specific simulation strategies, and effective use of progressive bias. The key strength of this paper lies in its innovative enhancements to MCTS, enabling it to handle the tactical and progression properties of LoA effectively. By incorporating game-theoretical value proving, the algorithm could identify and propagate winning and losing positions more efficiently, reducing the computational burden of extensive simulations. The use of domain-specific simulation strategies significantly improved the quality of the simulations, leading to more accurate evaluations and better overall performance. The empirical results demonstrated that the enhanced MCTS variant outperformed the world's best $\alpha\beta$-based LoA program, marking a significant milestone for MCTS in handling highly tactical games. The detailed analysis and systematic approach to integrating domain knowledge into MCTS provided a robust framework for applying MCTS to other complex games. Despite its advantages, the approach introduced additional computational overhead due to the need for maintaining and updating multiple enhancements, such as the progressive bias and game-theoretical value proving. The reliance on domain-specific knowledge for simulation strategies and the progressive bias limited the generalizability of the method to other games without similar properties. While the algorithm performed well in the controlled environment of LoA, its scalability, and robustness in more dynamic and less structured environments remained to be fully explored. The complexity of tuning various parameters, such as the progressive bias coefficient and the simulation cutoff threshold, could also pose challenges, particularly for practitioners without deep domain knowledge. Further validation in diverse real-world scenarios was necessary to assess the practical applicability and long-term benefits of the proposed enhancements.

Authors in \cite{santos2017monte} explored the application of MCTS to the popular collectible card game "Hearthstone: Heroes of Warcraft." Given the game's complexity, uncertainty, and hidden information, the authors proposed enriching MCTS with heuristic guidance and a database of decks to enhance performance. The approach was empirically validated against vanilla MCTS and state-of-the-art AI, showing significant performance gains. The integration of expert knowledge through heuristics and a deck database addressed two major challenges in Hearthstone: hidden information and large search space. By incorporating a heuristic function into the selection and simulation steps, the algorithm could more effectively navigate the game's complex state space, leading to improved decision-making and performance. The use of a deck database allowed the MCTS algorithm to predict the opponent's cards more accurately, enhancing the quality of simulations and overall strategy. The empirical results demonstrated that the enhanced MCTS approach significantly outperformed vanilla MCTS and was competitive with existing AI players, showcasing its potential for complex, strategic games like Hearthstone. However, the reliance on pre-constructed heuristics and deck databases introduced additional complexity and potential limitations. The effectiveness of the heuristic function was highly dependent on its design and tuning, which might vary across different game scenarios and decks. Similarly, the deck database's accuracy and comprehensiveness were crucial for predicting opponent strategies, which might be challenging to maintain as new cards and strategies emerged. The additional computational overhead associated with managing and updating these enhancements could impact real-time performance, particularly in time-constrained gameplay environments. While the approach showed strong performance in controlled experiments, further validation in diverse and dynamic real-world scenarios was necessary to fully assess its robustness and adaptability.

Information Set MCTS (ISMCTS), an extension of MCTS designed for games with hidden information and uncertainty was introduced in \cite{cowling2012information}. ISMCTS algorithms searched trees of information sets rather than game states, providing a more accurate representation of games with imperfect information. The approach was tested across three domains with different characteristics.
The ISMCTS algorithms excelled in efficiently managing hidden information by constructing trees where nodes represented information sets instead of individual states. This method mitigated the strategy fusion problem seen in determinization approaches, where different states in an information set were treated independently. By unifying statistics about moves within a single tree, ISMCTS made better use of the computational budget, leading to more informed decision-making. The empirical results demonstrated that ISMCTS outperformed traditional determinization-based methods in games with significant hidden information and partially observable moves. For instance, in Lord of the Rings: The Confrontation and the Phantom, ISMCTS achieved superior performance by effectively leveraging the structure of the game and reducing the impact of strategy fusion. However, the ISMCTS approach introduced additional complexity in maintaining and updating information sets, which could lead to increased computational overhead. The scalability of the method in environments with extensive hidden information and large state spaces remained a challenge, as the branching factor in information set trees could become substantial. While ISMCTS showed promising results in the tested domains, further validation in more diverse and dynamic scenarios was necessary to fully assess its robustness and general applicability. The reliance on accurate modeling of information sets and the necessity for domain-specific adaptations could limit the ease of implementation and the algorithm's flexibility across different types of games.

Authors in \cite{ciancarini2010monte} investigated the application of MCTS to Kriegspiel, a variant of chess characterized by hidden information and dynamic uncertainty. The authors explored three MCTS-based methods, incrementally refining the approach to handle the complexities of Kriegspiel. They compared these methods to a strong minimax-based Kriegspiel program, demonstrating the effectiveness of MCTS in this challenging environment. The authors' incremental refinement of MCTS methods for Kriegspiel effectively addressed the game's dynamic uncertainty and large state space. By leveraging a probabilistic model of the opponent's pieces and incorporating domain-specific heuristics, the refined MCTS algorithms significantly improved performance compared to the initial naive implementation. The experimental results showed that the final MCTS approach outperformed the minimax-based program, achieving better strategic planning and decision-making. The innovative use of a three-tiered game tree representation and opponent modeling techniques demonstrated the adaptability and robustness of MCTS in handling partial information games. This study provided valuable insights into the application of MCTS in environments with incomplete information, highlighting its potential for broader applications in similar domains. Having said that, the reliance on extensive probabilistic modeling and heuristic adjustments introduced additional complexity, which could be computationally intensive and challenging to maintain. The performance improvements were heavily dependent on the accuracy and relevance of the probabilistic models, which might vary across different game scenarios and opponents. While the approach showed strong performance in the controlled environment of Kriegspiel, its scalability, and robustness in more diverse and dynamic real-world scenarios remained to be fully validated. The necessity for domain-specific knowledge to construct effective heuristics and models limited the generalizability of the method to other games with different characteristics. Further research was needed to explore the method's applicability and efficiency in a wider range of partial information games and real-world decision-making tasks.

Two extensions of MCTS tailored for Asymmetric Trees (MCTS-T) and environments with loops were introduced in \cite{moerland2018monte}. The first algorithm, MCTS-T, addressed the challenges posed by asymmetric tree structures by incorporating tree uncertainty into the UCB formula, thus improving exploration. The second algorithm, MCTS-T+, further extended this approach to handle loops by detecting and appropriately managing repeated states within a single trace. The effectiveness of these algorithms was demonstrated through benchmarks on OpenAI Gym and Atari 2600 games. The key strength of MCTS-T lies in its ability to efficiently manage asymmetric trees by backing up the uncertainty related to the tree structure. This approach ensured that the algorithm could prioritize exploration in less explored subtrees, significantly improving the learning efficiency in environments with uneven tree depths. MCTS-T+ further enhanced this capability by addressing loops, preventing redundant expansions of the same state, and thus saving computational resources. The empirical results on benchmark tasks demonstrated that both MCTS-T and MCTS-T+ consistently outperformed standard MCTS, particularly in environments with strong asymmetry and loops. These improvements highlighted the potential of the proposed methods to extend the applicability of MCTS to more complex domains, such as robotic control and single-player video games, where asymmetry and looping were common. However, the reliance on non-stochastic environments and full state observability limited the generalizability of MCTS-T and MCTS-T+. These assumptions might not hold in many real-world scenarios, such as partially observable or highly stochastic environments, reducing the algorithms' applicability. The computational overhead associated with maintaining and updating the tree uncertainty and detecting loops could be substantial, particularly in large and dynamic state spaces. While the proposed methods showed promising results in controlled experiments, further validation in diverse real-world applications was necessary to fully assess their scalability and robustness. The complexity of tuning additional parameters, such as the uncertainty thresholds and loop detection mechanisms, posed additional challenges for practical implementation.

In the next subsection, we focus on Prioritized Sweeping, another algorithm that falls under the umbrella of Model-based Planning.

\paragraph{Prioritized Sweeping}

Prioritized Sweeping was introduced by \cite{moore1993prioritized}, which enhances Model-based RL by focusing updates on the most critical state-action pairs. This prioritization accelerates learning and convergence. As shown in Alg. \ref{alg:Prioritized_Sweeping}, Prioritized Sweeping's fundamental mechanisms include model learning, priority queue management, and backward sweeping. In model learning, the algorithm constructs a model of the environment, capturing state transition probabilities and rewards (lines 3-6). A priority queue is maintained, where state-action pairs are prioritized based on the magnitude of their potential update (error) (lines 7-8). Backward sweeping ensures that significant changes are promptly addressed by propagating their impact backward to predecessors (9-17). The process begins with experience collection (lines 3-5), where the agent interacts with the environment to gather transitions (state, action, reward, next state). The model is then updated with these new transitions (line 6). For each affected state-action pair, the algorithm calculates its priority (lines 7-8), reflecting the expected magnitude of the value update. Priorities are managed in a queue, and the algorithm performs value updates in order of priority until the queue is exhausted or a convergence criterion is met (lines 9-17) \cite{moore1993prioritized, peng1993efficient, sutton2018reinforcement}.

Prioritized Sweeping refines the Bellman Equation by updating states in a priority-driven manner as follows:
\begin{equation}
P(s,a) = \left| R(s,a) + \gamma \left( \max_{a'} Q(s',a') \right) - Q(s,a) \right|
\end{equation}
where \( P(s,a) \) is the priority of the state-action pair \((s,a)\). 
By analyzing several papers over the following paragraphs, the primary advantage of Prioritized Sweeping can be seen in its efficiency. By focusing updates on high-priority areas, it reduces unnecessary computations and accelerates learning. This prioritization also leads to faster convergence compared to traditional DP methods. However, the algorithm introduces complexity by maintaining a priority queue and performing backward sweeps, adding computational overhead. Additionally, the efficiency gains depend on the accuracy of the model, which might not be perfect in all scenarios \cite{moore1993prioritized, sutton2018reinforcement}.

\begin{algorithm}[t]
\caption{Prioritized Sweeping}
\begin{algorithmic}[1]
\State Initialize \( Q(s, a) \), \( \text{Model}(s, a) \), for all \( s, a \), and \( PQueue \) to empty
\While{true} \Comment{Do forever}
    \State \( S \gets \) current (non-terminal) state
    \State \( A \gets \text{policy}(S, Q) \)
    \State \parbox[t]{\dimexpr\linewidth-\algorithmicindent}{ Execute action \( A \); observe resultant reward, \( R \), and state, \( S' \)\strut}
    \State \( \text{Model}(S, A) \gets (R, S') \)
    \State \( P \gets | R + \gamma \max_a Q(S', a) - Q(S, A) | \)
    \If{\( P > \theta \)}
        \State Insert \( (S, A) \) into \( PQueue \) with priority \( P \)
    \EndIf
    \For{\( i = 1 \) to \( n \)}
        \If{\( PQueue \) is not empty}
            \State \( S, A \gets \text{first}(PQueue) \)
            \State \( R, S' \gets \text{Model}(S, A) \)
            \State \parbox[t]{\dimexpr\linewidth-\algorithmicindent}{ \( Q(S, A) \gets Q(S, A) \\
            + \alpha [ R + \gamma \max_a Q(S', a) - Q(S, A) ] \)\strut}
            \For{each \( (\tilde{S}, \tilde{A}) \) predicted to lead to \( S \)}
                \State \( \tilde{R} \gets \text{predicted reward for} \, (\tilde{S}, \tilde{A}, S) \)
                \State \( P \gets | \tilde{R} + \gamma \max_a Q(S, a) - Q(\tilde{S}, \tilde{A}) | \)
                \If{\( P > \theta \)}
                    \State \parbox[t]{\dimexpr\linewidth-\algorithmicindent}{ Insert \( (\tilde{S}, \tilde{A}) \) into \( PQueue \) with \\
                    priority \( P \)\strut}
                \EndIf
            \EndFor
        \EndIf
    \EndFor
\EndWhile
\end{algorithmic}
\label{alg:Prioritized_Sweeping}
\end{algorithm}

Prioritized Sweeping algorithm was introduced in \cite{peng1993efficient} to improve learning efficiency in stochastic Markov systems. The method combined the rapid performance of incremental learning methods like TD and Q-learning with the accuracy of DP by prioritizing updates. By maintaining a priority queue, the algorithm processed the most impactful updates first, leading to faster convergence in large state-space environments. Although the algorithm showed superior performance in simulations, it involved significant computational overhead due to the priority queue maintenance and state tracking. The complexity of setting up and tuning the priority mechanism posed challenges, particularly in dynamic environments with unpredictable state transitions.

An enhancement to the Deep Deterministic Policy Gradient (DDPG) (discussed in subsection \ref{Sec_DDPG}) algorithm by incorporating Prioritized Sweeping for morphing UAVs is introduced in \cite{li2020morphing}. The objective is to improve the efficiency and effectiveness of morphing strategy decisions by avoiding the random selection of state-action pairs and instead focusing on those with a significant impact on learning outcomes. The integration of Prioritized Sweeping with the DDPG framework notably enhances learning efficiency by prioritizing state-action pairs that are most influential in updating the policy. This targeted approach accelerates convergence and improves decision-making accuracy, which is crucial for the dynamic and complex task of UAV morphing. The method effectively combines the strengths of Value-based and Policy Gradient-based RL, leveraging deep neural networks for state evaluation and policy improvement. The simulation results demonstrate that the improved algorithm achieves faster learning and higher total rewards compared to DDPG, validating its superiority in handling complex morphing scenarios. Despite its advantages, the reliance on accurate prioritization mechanisms introduces additional complexity in maintaining and updating the priority queue, which could impact performance in highly dynamic environments where state changes are rapid and unpredictable. The assumption that changes in sweep angle do not significantly affect flight status within short time frames may oversimplify real-world conditions, potentially limiting the model's accuracy. While the simulation results are promising, further validation in real-world applications is necessary to fully assess the method's robustness and scalability. The initial phase of training and data generation still requires significant computational resources, which could be a bottleneck in practical deployments.

Authors in \cite{zajdel2018epoch} introduced modifications to Dyna-learning and Prioritized Sweeping algorithms, incorporating an epoch-incremental approach using Breadth-first Search (BFS). The combination of incremental and epoch-based updates improved learning efficiency, leading to faster convergence in dynamic environments like grid worlds. The use of BFS after episodes provided a more comprehensive understanding of the state space. However, managing dual modes of policy updates and accurate BFS calculations introduced complexity, potentially increasing computational overhead. The method showed strong simulation results, but real-world validation was necessary to understand its scalability and practical implications.

Authors in \cite{van2013planning} introduced an innovative extension to the Prioritized Sweeping algorithm by employing small backups instead of full backups. The primary aim is to enhance the computational efficiency of Model-based RL by reducing the complexity of value updates, making it more suitable for environments with a large number of successor states. The most notable advantage of this approach is its ability to perform value updates using only the current value of a single successor state, significantly reducing the computation time. By utilizing small backups, the algorithm allows for finer control over the planning process, leading to more efficient update strategies. The empirical results demonstrate that the small backup implementation of Prioritized Sweeping achieves substantial performance improvements over traditional methods, particularly in environments where full backups are computationally prohibitive. The theoretical foundation provided in the paper supports the robustness of the small backup approach, ensuring that it maintains the accuracy of value updates while enhancing computational efficiency. Additionally, the parameter-free nature of small backups eliminates the need for tuning step-size parameters, which is a common challenge in traditional sample-based methods. On the other hand, the approach's reliance on maintaining and updating a priority queue introduces additional memory and computational overhead, particularly in environments with many state-action pairs. While the method improves computational efficiency, it requires careful management of memory resources to store component values associated with small backups. The simulations primarily focus on deterministic environments, leaving the performance of the small backup approach in highly stochastic or dynamic settings less explored. Further validation in real-world applications is necessary to fully assess the scalability and practicality of the method. Additionally, the initial phase of model construction and priority queue management could introduce latency, impacting the algorithm's real-time performance in complex environments.

Authors in \cite{bargiacchi2020model} introduced Cooperative Prioritized Sweeping to efficiently handle Multi-Agent Markov Decision Processes (MMDPs) using factored Q-functions and Dynamic Decision Networks (DDNs). CPS managed large state-action spaces effectively, leading to faster convergence in multi-agent environments. However, the reliance on accurately specified DDN structures and the batch update mechanism introduced complexity, increasing computational overhead in dynamic environments. While simulations showed Cooperative Prioritized Sweeping's potential, further real-world validation was needed to assess its scalability and robustness.

A prioritized sweeping approach combined with Confidence-based Dual RL (CB-DuRL) for routing in Mobile Ad-Hoc Networks (MANETs) is investigated by researchers in \cite{desai2017prioritized}. The proposed method dynamically selects routes based on real-time traffic conditions, aiming to minimize delivery time and congestion. The key strength of this approach is its dynamic adaptability to real-time traffic conditions, addressing the limitations of traditional shortest path routing methods. By leveraging prioritized sweeping, the algorithm prioritizes updates to the most critical state-action pairs, enhancing learning efficiency and ensuring optimal path selection under varying network loads. The inclusion of CB-DuRL refines routing decisions by considering the reliability of Q-values, thus improving the robustness and reliability of the routing protocol. Empirical results from simulations on a 50-node MANET demonstrate that the proposed method significantly outperforms traditional routing protocols in terms of packet delivery ratio, dropping ratio, and delay, showcasing its effectiveness in handling high traffic conditions and reducing network congestion. However, the approach's dependence on accurate estimation of traffic conditions and Q-values may limit its adaptability in highly dynamic or unpredictable environments where traffic patterns change rapidly. The initial phase of gathering sufficient data to populate the Q-tables and confidence values can introduce latency and computational overhead, potentially impacting the algorithm's performance in real-time applications. While the simulations provide strong evidence of the method's efficacy, further validation in larger and more diverse network topologies is necessary to fully assess its scalability and robustness. Managing priority queues and updating Q-values in a distributed manner across multiple nodes can also pose challenges in maintaining synchronization and consistency in real-world deployments.

A structured version of Prioritized Sweeping to enhance RL efficiency in large state spaces by leveraging Dynamic Bayesian Networks (DBNs) was introduced in \cite{dearden2001structured}. The method accelerated learning by grouping states with similar values, reducing updates needed for convergence. DBNs provided a compact environment representation, further improving computational efficiency. However, the reliance on predefined DBN structures limited applicability in dynamic environments, and maintaining these structures added computational overhead. While promising in simulations, the method required further validation in diverse real-world scenarios to fully assess its scalability and robustness.

\paragraph{Dyna-Q}

There are two types of Dyna-Q we are twiddling with in RL; Dyna-Q when using with tabular representation and when using function approximation. Acknowledging this, we analyze both Dyna-Q's variations in this section.

The Dyna-Q algorithm, introduced in \cite{sutton1990integrated}, combines traditional RL with planning in a novel way to improve the efficiency and adaptability of learning agents. This section provides an overview of the essential concepts and mechanisms underlying the Dyna-Q algorithm. This algorithm integrates learning from real experiences with planning using a learned model of the environment. This integration allows the agent to improve its policy more rapidly and efficiently by leveraging both actual and simulated experiences. Dyna-Q utilizes Q-learning \cite{watkins1992q}. This estimate combines the immediate reward and the discounted value of the next state, guiding the agent towards actions that maximize long-term rewards. On top of that, its architecture includes a model of the environment that predicts the next state and reward given a current state and action. This model allows the agent to simulate hypothetical experiences, effectively planning future actions without needing to interact with the real environment continuously. Also, the agent uses the learned model to generate simulated experiences. These hypothetical experiences are treated similarly to real experiences, updating the $Q(s,a)$ values and refining the agent's policy. This process accelerates learning by allowing the agent to practice and plan multiple scenarios internally. Dyna-Q's planning process is incremental, meaning it can be interrupted and resumed at any time. This feature makes it highly adaptable to dynamic environments where the agent must continuously learn and update its policy based on new information.
This algorithm operates through a series of steps illustrated in Alg. \ref{alg:Dyna-Q}. First, the agent interacts with the environment, observes the current state, selects an action, and receives a reward and the subsequent state (lines 3-5). Based on this experience, the $Q(s,a)$ value is updated (line 6). Next, the world model is revised using the observed transition $(s, a, r, s')$ (line 7), which enhances its predictions for future planning. The agent then utilizes the world model to generate hypothetical experiences by selecting states and actions and predicting their outcomes (lines 8-14). These hypothetical experiences are used to further update the $Q(s,a)$ values, enabling the agent to refine its policy without additional real-world interactions. 

\begin{algorithm}[t]
\caption{Dyna-Q}
\begin{algorithmic}[1]
\State Initialize \( Q(s, a) \) and \( \text{Model}(s, a) \) for all \( s \in S \) and \( a \in A(s) \)
\Repeat
    \State \( S \gets \) current (non-terminal) state
    \State \( A \gets \epsilon\)-greedy\((S, Q)\)
    \State \parbox[t]{\dimexpr\linewidth-\algorithmicindent}{ Execute action \( A \); observe resultant reward, \( R \), and state, \( S' \)\strut}
    \State \parbox[t]{\dimexpr\linewidth-\algorithmicindent}{ \( Q(S, A) \gets Q(S, A) + \alpha [ R + \gamma \max_a Q(S', a) - Q(S, A) ] \)\strut}
    \State \parbox[t]{\dimexpr\linewidth-\algorithmicindent}{ \( \text{Model}(S, A) \gets R, S' \) \text{(assuming deterministic environment)}\strut}
    \For{\( i = 1 \) to \( n \)}
        \State \( S \gets \) random previously observed state
        \State \( A \gets \) random action previously taken in \( S \)
        \State \( R, S' \gets \text{Model}(S, A) \)
        \State \parbox[t]{\dimexpr\linewidth-\algorithmicindent}{ \( Q(S, A) \gets Q(S, A) + \\
        \alpha [ R + \gamma \max_a Q(S', a) - Q(S, A) ] \)\strut}
    \EndFor
\Until{convergence or a stopping criterion is met}
\end{algorithmic}
\label{alg:Dyna-Q}
\end{algorithm}

Over the next few paragraphs, several studies are analyzed regarding Dyna-Q algorithm.

This paper \cite{santos2012dyna} explored the development and application of the Dyna-H algorithm, a strategy that incorporated heuristic planning for optimizing decision-making in Role-playing Games (RPGs). The Dyna-H algorithm aimed to enhance path-finding efficiency by blending heuristic search with the Model-free characteristics of RL, specifically within the context of the Dyna architecture. The primary advantage of the Dyna-H algorithm lay in its innovative combination of heuristic search and, effectively addressing the challenges posed by large and complex search spaces typical in RPG scenarios. This synergy allowed the algorithm to focus on promising solution branches, significantly improving learning speed and policy quality compared to traditional methods like Q-learning and Dyna-Q. Additionally, the use of heuristic planning enabled the Dyna-H algorithm to perform efficiently without requiring a complete model of the environment beforehand, which was particularly beneficial for real-time applications. However, one drawback of the approach was the reliance on predefined heuristic functions, which might have limited its adaptability to various game environments where such heuristics were not readily available or accurate. The paper's evaluation, while thorough, focused primarily on deterministic grid-like environments, potentially overlooking the complexities and stochastic nature of more dynamic and unpredictable game settings. This narrow evaluation scope might have raised questions about the generalizability of the algorithm's performance in broader contexts.

In \cite{chai2021multi}, a Multi-objective Dyna-Q based Routing (MODQR) algorithm designed to optimize both delay and energy efficiency in wireless mesh networks was introduced. By leveraging the Dyna-Q technique, the algorithm aimed to address dynamic network conditions, ensuring that selected paths exhibited minimal delay and optimal energy usage. The key strength of the MODQR approach was its innovative integration of Dyna-Q with multi-objective optimization, effectively balancing delay and energy efficiency in real time. The use of Dyna-Q enhanced the convergence speed, making the algorithm well-suited for dynamic and complex network environments where conditions frequently change. Additionally, the consideration of both forward and reverse link conditions in determining path quality provided a more accurate assessment of route reliability. The inclusion of dynamic learning and exploration rates further refined the algorithm's adaptability, ensuring continuous improvement and optimization as the network evolved. However, the approach's reliance on predefined model parameters and assumptions, such as the gray physical interference model, might have limited its flexibility across different network topologies and conditions. While the Dyna-Q algorithm's fast convergence was advantageous, its performance in highly volatile environments with extreme fluctuations might not have been as robust. Moreover, the evaluation primarily focused on simulation results, which, while promising, might not have fully captured the complexities and unpredictable behaviors encountered in real-world deployments. The lack of comprehensive real-world testing left some uncertainty regarding the algorithm's practical effectiveness and scalability.

Authors in \cite{del2023new} presented a novel approach for managing Type-1 Diabetes Mellitus through a Glycemic control system using Dyna-Q. The system aimed to automate insulin infusion based on Continuous Glucose Monitoring (CGM) data, eliminating the need for manual carbohydrate input from patients. The primary strength of this approach lies in its innovative use of the Dyna-Q algorithm, which combines Model-based and Model-free RL to enhance learning efficiency and control accuracy. By leveraging past CGM and insulin data, the system could predict future glucose levels and optimize insulin dosage in real time, resulting in improved Glycemic control. The algorithm's ability to operate effectively without explicit carbohydrate information significantly reduced the cognitive load on patients, aligning well with the goals of developing a fully automated artificial pancreas. Furthermore, the incorporation of a precision medicine approach tailored the model to individual patients, enhancing the adaptability and accuracy of the Glycemic predictions. On the other hand, the reliance on precise CGM data and insulin records for model training might have posed challenges in real-world scenarios where data quality could vary. While the algorithm showed promise in simulations and preliminary tests with real patients, its generalizability across diverse patient populations and long-term robustness remained areas of concern. 
The study's limited real-world testing meant that further extensive clinical trials were necessary to fully validate the system's effectiveness and safety in everyday use.

Dyna-T which integrated Dyna-Q with Upper Confidence Bounds applied to Trees (UCT) for enhanced planning efficiency, was developed in \cite{faycal2022dyna}. The method aimed to improve action selection by using UCT to explore simulated experiences generated by the Dyna-Q model, demonstrating its effectiveness in several OpenAI Gym environments. The primary strength of Dyna-T was its ability to combine the Model-based learning of Dyna-Q with the robust exploration capabilities of UCT. This combination allowed for a more directed exploration strategy, which enhanced the agent's ability to find optimal actions in complex and stochastic environments. The algorithm's use of UCT ensured that the most promising action paths were prioritized, significantly improving learning efficiency and convergence speed. The empirical results indicated that Dyna-T outperformed traditional RL methods, especially in environments with high variability and sparse rewards, showcasing its potential for broader applications in RL tasks. Despite its advantages, Dyna-T exhibited some limitations, particularly in deterministic environments where the added computational complexity of UCT might not have yielded significant benefits. The initial overhead of constructing and maintaining the UCT structure could be costly, especially in simpler tasks where traditional Dyna-Q or Q-learning might have sufficed. Moreover, while Dyna-T showed promise in simulated environments, its performance in real-world scenarios with continuous state and action spaces remained to be fully validated. The reliance on the exploration parameter 'c' and its impact on performance also warranted further investigation to ensure robust adaptability across different problem domains.

A novel multi-path load-balancing routing algorithm for Wireless Sensor Networks (WSNs) leveraging the Dyna-Q algorithm was proposed in \cite{xu2015dyna}. The proposed method termed ELMRRL (Energy-efficient Load-balancing Multi-path Routing with RL), aimed to minimize energy consumption and enhance network lifetime by selecting optimal routing paths based on residual energy, hop count, and energy consumption of nodes. The main advantage of the ELMRRL algorithm was its effective integration of Dyna-Q RL, which combined real-time learning with planning to adaptively select optimal routing paths. This dynamic adjustment ensured that the algorithm could respond to changes in network conditions, thus extending the network lifetime. The use of RL enabled each sensor node to act as an agent, making independent decisions based on local information, which was crucial for distributed environments like WSNs. Furthermore, the algorithm’s focus on both immediate and long-term cumulative rewards led to more balanced energy consumption across the network, preventing premature node failures and improving overall network resilience. On the other hand, the reliance on local information for decision-making might have limited the algorithm's effectiveness in scenarios where global network knowledge could provide more optimal solutions. The initial learning phase, where nodes gathered and processed data to update their Q-values, could have introduced latency and computational overhead, potentially affecting the algorithm’s performance in highly dynamic environments. Additionally, the approach’s dependence on the correct parameterization of the reward function and learning rates could have been challenging, as these parameters significantly impacted the algorithm’s efficiency and convergence speed. The paper primarily validated the algorithm through simulations, so its real-world applicability in diverse WSN deployments remained to be fully explored.

Researchers in \cite{li2023motion} introduced a new motion control method for path planning in unfamiliar environments using the Dyna-Q algorithm. The goal of the proposed method was to improve the efficiency of motion control by combining direct RL with model learning, enabling agents to effectively navigate both dynamic and static obstacles. The main benefit of this method was its use of the Dyna-Q algorithm, which merged Model-based and Model-free RL techniques, facilitating concurrent learning and planning. This integration notably enhanced convergence speed and adaptability to dynamic environments, as demonstrated by faster path optimization compared to traditional Q-learning. The algorithm’s capacity to simulate experiences allowed the agent to plan more efficiently, resulting in improved decision-making in complex situations. Furthermore, employing an $\epsilon$-greedy policy for action selection ensured a balanced exploration-exploitation trade-off, which was essential for finding optimal paths in unknown environments. Nonetheless, the method's dependence on accurate environment models for effective planning might have restricted its performance in highly unpredictable settings where models might not accurately reflect real-world dynamics. The computational burden of maintaining and updating the Q-table and environment model could also have been significant, especially in large state-action spaces. Additionally, while the method demonstrated promising results in simulation environments, its scalability and robustness in real-world applications with continuous state spaces and real-time constraints needed further validation. The reliance on predefined parameters, such as learning rates and discount factors, could also have impacted the algorithm's efficiency and effectiveness, requiring careful tuning for different scenarios.

Researchers in \cite{pei2021improved} introduced an improved Dyna-Q algorithm for mobile robot path planning in unknown, dynamic environments. The proposed method integrated heuristic search strategies, Simulated Annealing (SA), and a new action-selection strategy to enhance learning efficiency and path optimization. The enhanced Dyna-Q algorithm effectively merged heuristic search and SA, significantly improving the exploration-exploitation balance. By incorporating heuristic rewards and actions, the algorithm ensured efficient navigation through complex environments, avoiding local minima and achieving faster convergence. The novel SA-$\epsilon$-greedy policy dynamically adjusted exploration rates, optimizing the learning process. Empirical results from simulations and practical experiments showed that the improved algorithm outperformed Q-learning and Dyna-Q methods, demonstrating superior global search capabilities, enhanced learning efficiency, and robust convergence properties in both static and dynamic obstacle scenarios. Despite the performance improvements from integrating heuristic search and SA, the approach's reliance on predefined heuristic functions might have limited its adaptability to diverse environments. The initial training phase required extensive exploration, potentially leading to higher computational overhead and longer training times. Additionally, the dependence on grid-based environment representation might have restricted the algorithm's scalability to continuous state spaces. The focus on simulated and controlled real-world environments raised concerns about the algorithm's robustness and generalizability in more complex and unpredictable real-world applications. Further studies were necessary to validate the method's effectiveness in larger, unstructured, and more dynamic environments.

An innovative approach for scheduling the charging of Plug-in Electric Vehicles (PEVs) using the Dyna-Q algorithm was developed in \cite{wang2020autonomous}. The primary goal was to minimize long-term charging costs while considering the stochastic nature of driving behavior, traffic conditions, energy usage, and fluctuating energy prices. The method formulated the problem as a MDP and employed DRL techniques for solution optimization. The key strength of this approach lies in its effective combination of Model-based and Model-free RL, which enhances learning speed and efficiency. By continuously updating the model with real experiences and generating synthetic experiences, the Dyna-Q algorithm ensured rapid convergence to an optimal charging policy. This dual approach allowed for robust decision-making even in the face of uncertain and dynamic real-world conditions, such as varying energy prices and unpredictable driving patterns. Additionally, the integration of a deep-Q network for value approximation facilitated handling the vast state space inherent in PEV charging scenarios, ensuring the method's scalability and applicability to real-world settings. However, the reliance on accurate initial parameter values and predefined models for the user’s driving behavior might have limited the algorithm's adaptability during the initial learning phase. The system's performance heavily depended on the quality and accuracy of these models, which might have varied across different users and environments. Moreover, while the approach demonstrated significant improvements in simulations, its effectiveness in diverse and more complex real-world scenarios required further validation. The potential computational overhead associated with maintaining and updating the deep-Q network and model could have posed challenges for real-time applications, especially in large-scale deployments with numerous PEVs.

Authors in \cite{liu2022improved} presented an enhanced Dyna-Q algorithm tailored for Automated Guided Vehicles (AGVs) navigating complex, dynamic environments. The key improvements included a global path guidance mechanism based on heuristic graphs and a dynamic reward function designed to address issues of sparse rewards and slow convergence in large state spaces. The improved Dyna-Q algorithm stood out by integrating heuristic graphs that provided a global perspective on path planning, significantly reducing the search space and enhancing efficiency. This method enabled AGVs to quickly orient towards their goals by leveraging precomputed shortest path information, thus mitigating the problem of sparse rewards commonly encountered in extensive environments. Additionally, the dynamic reward function intensified feedback, guiding AGVs more effectively through complex terrains and around obstacles. The experimental results in various scenarios with static and dynamic obstacles demonstrated superior convergence speed and learning efficiency compared to traditional Q-learning and standard Dyna-Q algorithms, highlighting its robustness and effectiveness in dynamic settings. However, the dependency on heuristic graphs, which required prior computation, might have limited the algorithm's adaptability in environments where real-time updates were necessary or in scenarios with unpredictable changes. The initial setup phase, involving the creation of the heuristic graph, could have introduced overheads that might not have been feasible for all applications. Furthermore, while the dynamic reward function enhanced learning efficiency, its design relied heavily on accurate modeling of the environment, which could have been challenging in highly variable or noisy conditions. The paper's focus on simulated environments left room for further validation in real-world applications, where additional factors such as sensor noise and real-time constraints could have impacted performance.

A Dyna-Q based anti-jamming algorithm designed to enhance the efficiency of path selection in wireless communication networks subject to malicious jamming was introduced in \cite{zhang2022anti}. By leveraging both Model-based and Model-free RL techniques, the algorithm aimed to optimize multi-hop path selection, reducing packet loss and improving transmission reliability in hostile environments. The application of the Dyna-Q algorithm in this context was innovative, combining direct learning with simulated experiences to accelerate the convergence of the Q-values. This dual approach allowed the system to adapt quickly to dynamic jamming conditions, ensuring more reliable path selection and communication efficiency. The inclusion of a reward function that considered various modulation modes based on the Signal-to-Jamming Noise Ratio (SJNR) enhanced the robustness of the algorithm. Simulation results demonstrated that the Dyna-Q algorithm significantly outperformed traditional Q-learning and multi-armed bandit models, achieving faster convergence to optimal paths and better handling of interference, thus showcasing its potential for real-time applications in complex electromagnetic environments. Nevertheless, the method's reliance on pre-established environmental models might have limited its effectiveness in highly unpredictable or rapidly changing conditions, where the initial models might not accurately capture real-time dynamics. The need for accurate initial state representations and model updates introduced additional computational overhead, which could have impacted performance in larger or more complex networks. Furthermore, while the algorithm showed promise in simulated environments, its scalability and adaptability in real-world applications with varying node densities and jamming strategies required further validation. The focus on hop count minimization might also have overlooked other critical factors such as energy consumption and latency, which were essential for comprehensive network performance assessment.

Researchers in \cite{hwang2014model} introduced a Model-based RL method using Dyna-Q tailored for multi-agent systems. It emphasized efficient environmental modeling through a tree structure and proposed methods for model sharing among agents to enhance learning speed and reduce computational costs. The approach leveraged the concept of knowledge sharing, where agents with more experience assisted others by disseminating valuable information. The integration of a tree-based model for environmental learning within the Dyna-Q framework significantly enhanced the efficiency of model construction and memory usage. This method allowed agents to generate virtual experiences, thus accelerating the learning process. The innovative model sharing techniques proposed in the paper, such as grafting partial branches and resampling, enabled agents to build more accurate models collaboratively. By reducing redundant data transfer and focusing on useful experiences, these sharing methods improved sample efficiency and learning speed in complex environments. The simulation results demonstrated the effectiveness of these techniques in multi-agent cooperation scenarios, highlighting their potential to optimize learning in large, continuous state spaces. Despite its advantages, the reliance on accurate decision tree models for sharing experiences might have limited the approach's flexibility in highly dynamic or heterogeneous environments. The effectiveness of the sharing methods depended on the quality and relevance of the shared models, which might have varied across different agents and scenarios. Additionally, the initial phase of building accurate tree models could have been computationally intensive, particularly in environments with high variability. While the proposed methods showed promising results in simulations, further validation in diverse real-world applications was needed to fully assess their scalability and robustness. The paper also assumed a certain level of homogeneity among agents, which might not always have been applicable in more varied multi-agent systems.

The application of the Dyna-Q algorithm for path planning and obstacle avoidance in Unmanned Ground Vehicles (UGVs) and UAVs within complex urban environments was explored in \cite{huang2022path}. The study focused on utilizing a vector field-based approach for effective navigation and air-ground collaboration tasks. The integration of the Dyna-Q algorithm with a vector field method significantly enhanced the efficiency and accuracy of path planning in dynamic urban settings. The approach leveraged both real and simulated experiences to adaptively update the agent's policy, ensuring rapid convergence to optimal paths. By simplifying the urban environment into a grid world, the method allowed for precise waypoint calculation, facilitating smooth navigation and effective obstacle avoidance. The use of PID controllers for UAV and UGV coordination further improved the stability and responsiveness of the system, enabling robust air-ground collaboration. Simulation results demonstrated that the proposed method effectively handled dynamic obstacles and complex path scenarios, showcasing its potential for real-world applications in urban environments. On the other hand, the reliance on grid-based environment representation might have limited the algorithm's scalability and adaptability to continuous state spaces found in more diverse and unstructured urban areas. The initial phase of creating the vector field and the grid map could have introduced computational overheads, which might have impacted real-time performance. While the paper focused on simulated environments, further validation in real-world scenarios was necessary to assess the approach's robustness and effectiveness under varying conditions. Additionally, the method's dependency on accurate dynamic models for both UGV and UAV could have posed challenges, as any discrepancies between the model and the real environment might have affected the overall performance.

This paper \cite{vitolo2018performance} evaluated the performance of the Dyna-Q algorithm in robot navigation within partially known environments containing static obstacles. The study extended the Dyna-Q algorithm to multi-robot systems and conducted extensive simulations to assess its efficiency and effectiveness. The primary strength of this study lies in its thorough analysis of the Dyna-Q algorithm in both single and multi-agent contexts. By integrating planning and Model-based learning, the Dyna-Q algorithm sped up the learning process, enabling robots to navigate efficiently even with limited prior knowledge of the environment. The use of simulations with the Robot Motion Toolbox allowed for a comprehensive evaluation of the algorithm’s performance across various parameters, providing valuable insights into the optimal settings for different scenarios. Extending the Dyna-Q algorithm to multi-robot systems showcased its adaptability and potential for complex task coordination, where agents could share knowledge to enhance overall system performance. However, the paper's focus on static obstacles and deterministic environments might have limited the applicability of the findings to more dynamic and stochastic settings. The initial need for environment discretization and model construction introduced additional computational overhead, which could have been a bottleneck in real-time applications. While the simulations offered a robust analysis, the lack of real-world validation left some uncertainty about the algorithm's practical effectiveness in unpredictable and continuously changing environments. Furthermore, the performance degradation observed in multi-agent scenarios indicated that further refinement was needed to improve coordination and reduce inter-agent interference. 

The primary advantage of Dyna-Q is its ability to accelerate learning by combining real and simulated experiences. This dual approach reduces the dependency on extensive real-world interactions, making it suitable for applications where such interactions are costly or limited. Moreover, Dyna-Q is inherently adaptable to changing environments. The integration of continuous learning and planning allows the agent to update its policy dynamically in response to new information or changes in the environment \cite{sutton1990integrated, sutton2018reinforcement}. Table \ref{tab:Model_Based_Papers} provides a summary of the papers that utilized Model-based Planning algorithms.

\begin{table}[t]
\centering
\renewcommand{\arraystretch}{1.2} 
\caption{Model-based Planning Papers Review }
\begin{tabular}{|>{\raggedright\arraybackslash}p{4cm}|>{\raggedright\arraybackslash}p{3cm}|}
\hline
\textbf{Application Domain} & \textbf{References} \\
\hline
Theoretical Research (Convergence, stability) & \cite{peng1993efficient}, \cite{van2013planning}  \\
\hline
Multi-agent Systems and Autonomous Behaviors & \cite{zerbel2019multiagent}, \cite{bargiacchi2020model} \cite{hwang2014model}, \cite{wang2020autonomous}, \cite{liu2022improved}, \cite{huang2022path} \\
\hline
Games and Simulations & \cite{coulom2006efficient}, \cite{wang2012multi}, \cite{perez2014multiobjective}, \cite{de2016monte}, \cite{winands2010monte}, \cite{santos2017monte}, \cite{cowling2012information}, \cite{ciancarini2010monte}, \cite{moerland2018monte} \cite{santos2012dyna}, \cite{pei2021improved}  \\
\hline
Energy and Power Management (IoT Networks, Smart Energy Systems) & \cite{xu2015dyna}, \cite{del2023new} \\
\hline
Transportation and Routing Optimization & \cite{desai2017prioritized}\\
\hline
Network Resilience and Optimization & \cite{zhang2022anti}, \cite{xu2015dyna} \\
\hline
Hybrid RL Algorithms & \cite{su2021prioritized}, \cite{baier2013monte}, \cite{lanctot2014monte}, \cite{li2020morphing}
\cite{faycal2022dyna} \\
\hline
Dynamic Bayesian networks & \cite{dearden2001structured}
 \\
\hline
Dynamic Environments & \cite{chai2021multi}, \cite{li2023motion}, \cite{pei2021improved}, \cite{liu2022improved}  
 \\
\hline
Robotics & \cite{vitolo2018performance} \\
\hline
\end{tabular}
\label{tab:Model_Based_Papers}
\end{table}

Over the next section, a complete introduction to another paradigm of RL, Policy-based Methods, is given, along with analyzing various algorithms that fall under its umbrella.

\section{Policy-based Methods}\label{sec:PBM}

Policy-based methods are another fundamental RL method that more strongly emphasizes direct policy optimization in the process of choosing actions for an agent. In contrast to Value-based methods, which search for the value function implicit in the task, and then derive an optimal policy, Policy-based methods directly parameterize and optimize the policy. This approach offers several advantages, particularly better dealing with very challenging environments that have high-dimensional action spaces or where policies are inherently stochastic. Perhaps at the core, Policy-based methods conduct their operation based on the parameterization of policies, usually denoted as $\pi(a|s;\theta)$. Here, $\theta$ is used to denote the parameters of the policy, while $s$ denotes the state and $a$ denotes the action. In other words, it finds the optimal parameters $\theta^*$ that maximize the expected cumulative reward. Needless to say, this is generally done by gradient ascent techniques and more specifically by Policy Gradient methods that explicitly compute the gradient of expected reward with respect to the policy parameters, modifying parameters in the direction of reward increase \cite{sutton2018reinforcement, kaelbling1996reinforcement, kober2013reinforcement}.

The gradient of the policy with respect to Q values is estimated in the following way:

\begin{equation}
\nabla_{\theta} J(\theta) = \mathbb{E}_{\pi_{\theta}} \left[ \nabla_{\theta} \log \pi_{\theta}(a|s) Q_{\pi_{\theta}}(s, a) \right]
\end{equation}

Here, $Q_{\pi_{\theta}}(s, a)$ represents the action-value function under the current policy. This gradient can be estimated using samples from the environment, allowing the application of gradient ascent to update the policy parameters iteratively. The advantages of Policy-based Methods can be simplified to:

\begin{itemize}
    \item \textbf{Direct Optimization of the Policy:} These methods thus optimize the policy directly, hence, in contrast to the Value-based methods, working more successfully with continuous and high-dimensional action spaces. They usually work very well for problems within robotics and control where actions are naturally continuous.
    
    \item \textbf{Stochastic Policies:} Policy-based methods naturally accommodate stochastic policies, which may be important in settings where exploration is necessary or the optimal policy is inherently stochastic. Stochastic policies aid in shooting up the exploration-exploitation trade-off more efficiently as well.
    
    \item \textbf{Improved Convergence Properties:} Sometimes, Policy-based methods may have much smoother convergence than Value-based methods, especially when the latter are prone to a number of instabilities and divergence related to the Church condition for value function approximations.
\end{itemize}

Now, it is time to talk about the first Policy Gradient method, REINFORCE, in the next subsection.

\subsection{REINFORCE}
The REINFORCE algorithm is a seminal contribution to RL, particularly within the context of policy gradient methods. The algorithm is designed to optimize the expected cumulative reward by adjusting the policy parameters in the direction of the gradient of the expected reward. As demonstrated in Alg. \ref{alg:REINFORCE}, the REINFORCE algorithm is rooted in the stochastic policy framework, where the policy, parameterized by \(\theta\), defines a probability distribution over actions given the current state. The key insight of the REINFORCE algorithm is to use the log-likelihood gradient estimator to update the policy parameters \cite{williams1992simple}. The gradient of the expected reward with respect to the policy parameters \(\theta\) is given by:

\begin{equation}
\nabla_\theta J(\theta) = \mathbb{E}_\pi \left[ \nabla_\theta \log \pi_\theta(a|s) G_t \right],
\end{equation}
where \(\pi_\theta(a|s)\) is the probability of taking action \(a\) in the state \(s\) under policy \(\pi\) parameterized by \(\theta\), and \(G_t\) is the return (cumulative future reward) following time step \(t\). This gradient estimation forms the basis for the parameter update rule in REINFORCE (line 7):

\begin{equation}
\theta \leftarrow \theta + \alpha \nabla_\theta \log \pi_\theta(a|s) G_t,
\end{equation}
where \(\alpha\) is the learning rate. 

\begin{algorithm}[t]
\caption{REINFORCE}
\begin{algorithmic}[1]
\State \textbf{Input:} a differentiable policy parameterization \(\pi(a|s, \theta)\)
\State Initialize policy parameter \(\theta \in \mathbb{R}^{d'}\)
\Repeat \Comment{forever:}
    \State \parbox[t]{\dimexpr\linewidth-\algorithmicindent}{ Generate an episode  \(S_0, A_0, R_1, \dots, S_{T-1}, A_{T-1}, R_T\), following \(\pi(\cdot|\cdot; \theta)\)\strut}
    \For{each step of the episode \(t = 0, \dots, T-1\)}
        \State \(G \gets\) return from step \(t\)
        \State \(\theta \gets \theta + \alpha \gamma^t G \nabla_{\theta} \ln \pi(A_t | S_t, \theta)\)
    \EndFor
\Until{convergence or a stopping criterion is met}
\end{algorithmic}
\label{alg:REINFORCE}
\end{algorithm}

One of the strengths of the REINFORCE algorithm is its simplicity and generality, allowing it to be applied across a wide range of problems. However, it also faces challenges such as high variance in the gradient estimates and slow convergence, particularly in environments with sparse or delayed rewards. By analyzing the applications of REINFORCE, we will have a deeper look at these advantages and disadvantages.

Authors in \cite{zhang2021sample} investigated the global convergence rates of the REINFORCE algorithm in episodic RL settings. The authors aimed to close the gap between theoretical and practical implementations of policy gradient methods by providing new convergence results for the REINFORCE algorithm. The paper's strengths lay in its comprehensive theoretical analysis and practical relevance. The authors derived performance bounds for the REINFORCE algorithm using a fixed mini-batch size, aligning more closely with practical implementations. They provided the first set of global convergence results for the REINFORCE algorithm, including sub-linear high probability regret bounds and almost sure global convergence of average regret. The focus on the widely-used REINFORCE gradient estimation procedure rather than state-action visitation measure-based estimators addressed a significant gap between theory and practice. The authors established that the REINFORCE algorithm was sample efficient, with polynomial complexity, which was crucial for practical applications due to the high cost of obtaining samples. However, the complexity of the analysis might have challenged practitioners less familiar with the mathematical concepts. The reliance on assumptions such as the log-barrier regularization term and soft-max policy parametrization might have limited the generality of the results. The paper focused on the stationary infinite-horizon discounted setting, and a more detailed discussion on applying the results to other settings would have enhanced its relevance. The absence of empirical validation of the proposed convergence bounds and sample efficiency was another limitation. Including experimental results would have provided additional evidence of the practical utility of the findings.

A novel approach to Energy Management Strategies (EMS) in Fuel Cell Hybrid EVs (FCHEV) using the fuzzy REINFORCE algorithm was introduced in \cite{guo2023function}. This method integrated a fuzzy inference system (FIS) with Policy Gradient RL (PGRL) to optimize energy management, achieve hydrogen savings, and maintain battery operation. One of the key strengths of the paper was the innovative combination of fuzzy logic with the REINFORCE algorithm. By employing a fuzzy inference system to approximate the policy function, the authors effectively leveraged the generalization capabilities of fuzzy logic to handle the complexity and uncertainty inherent in energy management tasks. This integration helped to address the limitations of traditional EMS methods that relied heavily on expert knowledge and static rules, thus providing a more adaptive and robust solution. The use of a fuzzy baseline function to stabilize the training process and reduce the variance in policy gradient updates was another notable advantage. This approach enhanced the convergence rate and stability of the learning process, which was particularly beneficial in real-time applications where computational efficiency and robustness were critical. The paper's demonstration of the algorithm's adaptability to changing driving conditions and system states further underscored its practical relevance and effectiveness. However, the complexity of the proposed method might have posed implementation challenges, particularly for practitioners who were less familiar with fuzzy logic. The integration of FIS and PGRL required careful tuning of parameters and membership functions, which could have been time-consuming and computationally intensive. Additionally, while the fuzzy REINFORCE algorithm showed promise in reducing the computational burden and improving convergence, the reliance on fuzzy logic introduced an additional layer of complexity that might not have been necessary for all applications. The paper also provided a comprehensive analysis of the simulation and hardware-in-loop experiments, validating the effectiveness of the proposed method in real-world scenarios. The results indicated that the fuzzy REINFORCE algorithm could achieve near-optimal performance without requiring accurate system models or extensive prior knowledge, making it a versatile and practical solution for EMS in FCHEVs.

Authors in \cite{lauffenburger2021reinforcement} presented a study protocol for a trial aimed at improving medication adherence among patients with type 2 diabetes using an RL-based text messaging program. One of the key strengths of this study was its innovative use of RL to personalize text message interventions. By tailoring messages based on individual responses to previous messages, the approach had the potential to optimize engagement and improve adherence more effectively than generic messaging strategies. This personalized communication could have led to more significant behavior changes and better health outcomes for patients with diabetes. The study's design also enhanced its practical relevance. Conducted in a real-world setting at Brigham and Women’s Hospital, it involved patients with suboptimal diabetes control, which reflected a common clinical scenario. The use of electronic pill bottles to monitor adherence provided accurate and objective data, supporting the reliability of the study outcomes. Additionally, the trial's primary outcome of average medication adherence over six months was a meaningful measure that directly related to the study's objective. However, there were some weaknesses and challenges associated with the study. The requirement for patients to use electronic pill bottles and smartphones with a data plan or WiFi might have limited the generalizability of the findings to populations without access to such technology. Furthermore, the study's reliance on self-reported adherence as a secondary outcome introduced the potential for reporting bias. The study also faced potential limitations related to the length of the follow-up period and the evaluation of the long-term sustainability of the intervention. While a six-month follow-up period was sufficient to assess initial adherence improvements, longer-term studies would have been necessary to determine whether the benefits of the intervention were sustained over time.

In \cite{tao2024reinforcement}, the authors presented a novel method for rate adaptation in 802.11 wireless networks leveraging the REINFORCE algorithm. The proposed approach, named ReinRate, integrated a comprehensive set of observations, including received signal strength, contention window size, current modulation and coding scheme, and throughput, to adapt dynamically to varying network conditions and optimize network throughput. One of the key strengths of this paper was its innovative application of the REINFORCE algorithm to WiFi rate adaptation. Traditional rate adaptation algorithms like Minstrel and Ideal relied on limited observations such as packet loss rate or signal-to-noise ratio, which could have been insufficient in dynamic wireless environments. In contrast, ReinRate's broader set of observations allowed for a more nuanced response to varying conditions, leading to significant improvements in network performance. The authors demonstrated that ReinRate outperformed Minstrel and Ideal algorithms by up to 102.5\% and 30.6\% in network scenarios without interference, and by up to 35.1\% and 66.6\% in scenarios with interference. Another strength was the comprehensive evaluation of ReinRate using the ns-3 network simulator and ns3-ai OpenAI Gym. The authors conducted extensive simulations under various network scenarios, both static and dynamic, with and without interference. This thorough evaluation provided strong evidence of the algorithm's effectiveness and adaptability in real-world conditions. The results indicated that ReinRate consistently achieved higher throughput compared to traditional algorithms, showcasing its ability to handle the challenges of dynamically changing wireless environments. However, the complexity of the proposed method might have posed challenges for practical implementation. The integration of multiple observations and the application of the REINFORCE algorithm required careful tuning of parameters and computational resources.

A new approach to enhance DRL for outdoor robot navigation was investigated in \cite{weerakoon2022htron}. The key innovation was the use of a heavy-tailed policy parameterization, which induced exploration in sparse reward settings, a common challenge in outdoor navigation tasks. A significant strength of the paper lies in addressing the sparse reward issue, which was prevalent in many real-world navigation scenarios. Traditional DRL methods often relied on carefully designed dense reward functions, which could have been impractical to implement. The authors proposed HTRON, an algorithm that leveraged heavy-tailed policy parameterizations, such as the Cauchy distribution, to enhance exploration without needing complex reward shaping. This approach allowed the algorithm to learn efficient behaviors even with sparse rewards, making it more applicable to real-world scenarios. The paper's thorough experimental evaluation was another strong point. The authors tested HTRON against established algorithms like REINFORCE, Proximal Policy Optimization (PPO), and Trust Region Policy Optimization (TRPO) (explained later in the upcoming subsections) across three different outdoor scenarios: goal-reaching, obstacle avoidance, and uneven terrain navigation. HTRON outperformed these algorithms in terms of success rate, average time steps to reach the goal and elevation cost, demonstrating its effectiveness and efficiency. The use of a realistic unity-based simulator and the deployment of the algorithm on a Clearpath Husky robot further validated the practical applicability of the proposed method. However, the complexity of the proposed algorithm and the specific choice of heavy-tailed distributions might have posed challenges. The implementation of heavy-tailed policy gradients could have introduced instability, especially in the initial learning phases. While the authors mitigated this with adaptive moment estimation and gradient clipping, these techniques required careful tuning and expertise, potentially limiting accessibility for practitioners.

Table \ref{tab:REINFORCE_Papers}, gives an overview of the papers reviewed in this section.
The next Policy-based algorithm, which we need to cover is TRPO. Over the next subsection, we cover this algorithm.

\begin{table}[t]
\centering
\renewcommand{\arraystretch}{1.2} 
\caption{REINFORCE Papers Review}
\begin{tabular}{|>{\raggedright\arraybackslash}p{4cm}|>{\raggedright\arraybackslash}p{3cm}|}
\hline
\textbf{Application Domain} & \textbf{References} \\
\hline
Energy and Power Management & \cite{guo2023function}  \\
\hline
Theoretical Research (Convergence, stability) & \cite{zhang2021sample} \\
\hline
Network Optimization & \cite{tao2024reinforcement}
 \\
\hline
Robotics & \cite{weerakoon2022htron} \\
\hline
\end{tabular}
\label{tab:REINFORCE_Papers}
\end{table}

\subsection{Trust Region Policy Optimization (TRPO)}
TRPO, introduced by \cite{schulman2015trust}, is an advancement in RL, specifically within policy optimization methods. The primary objective of TRPO is to optimize control policies with guaranteed monotonic improvement, addressing the shortcomings of previous methods \cite{bertsekas2012dynamic, peters2008reinforcement, szita2006learning} that often resulted in unstable policy updates and poor performance on complex tasks.

TRPO is designed to handle large, nonlinear policies such as those represented by neural networks. The algorithm ensures that each policy update results in a performance improvement by maintaining the updated policy within a "trust region" around the current policy. This trust region is defined using a constraint on the KL divergence between the new and old policies, effectively preventing large, destabilizing updates \cite{liu2019neural, schulman2015trust}.
TRPO operates within the stochastic policy framework, where the policy \(\pi_\theta\) is parameterized by \(\theta\) and defines a probability distribution over actions given the states. The expected discounted reward for a policy \(\pi\) is given by:

\begin{equation}
    J(\pi) = \mathbb{E}\left[\sum_{t=0}^{\infty} \gamma^t r(s_t, a_t) \right],
\end{equation}
where \(\gamma\) is the discount factor, \(r(s_t, a_t)\) is the reward at time step \(t\), and the expectation is taken over the state and action trajectories induced by the policy.
To ensure that the policy update remains within a safe boundary, TRPO constrains the KL divergence between the new policy \(\pi_{\theta'}\) and the old policy \(\pi_{\theta}\):

\begin{equation}
    D_{KL}(\pi_{\theta} \| \pi_{\theta'}) \leq \zeta,
\end{equation}
where \(\zeta\) is a small positive constant. This constraint ensures that the new policy does not deviate too much from the old policy, thereby providing stability to the learning process.
TRPO optimizes a surrogate objective function that approximates the true objective while respecting the trust region constraint. The surrogate objective \(L(\theta)\) is defined as:

\begin{equation}
    L(\theta) = \hat{\mathbb{E}} \left[ \frac{\pi_{\theta}(a|s)}{\pi_{\theta_{\text{old}}}(a|s)} \hat{A}_{\theta_{\text{old}}}(s, a) \right],
\end{equation}
where \(\hat{A}_{\theta_{\text{old}}}(s, a)\) is an estimate of the advantage function, which measures the relative value of taking action \(a\) in state \(s\) under the old policy.
%
%
As demonstrated in Alg. \ref {alg:TRPO}, the practical implementation of TRPO involves the following steps:

\begin{enumerate}
    \item \textbf{Sample Trajectories:} Collect a set of trajectories using the current policy \(\pi_{\theta_{\text{old}}}\) (line 3).
    \item \textbf{Estimate Advantages:} Compute the advantage function \(\hat{A}_{\theta_{\text{old}}}(s, a)\) using the collected trajectories (line 4).
    \item \textbf{Optimize Surrogate Objective:} Solve the constrained optimization problem to find the new policy parameters \(\theta'\) (lines 5-7):
    \begin{equation}
    \theta' = \arg\max_{\theta} \hat{\mathbb{E}} \left[ \frac{\pi_{\theta}(a|s)}{\pi_{\theta_{\text{old}}}(a|s)} \hat{A}_{\theta_{\text{old}}}(s, a) \right],
    \end{equation}
    subject to
    \begin{equation}
    D_{KL}(\pi_{\theta_{\text{old}}} \| \pi_{\theta}) \leq \zeta.
    \end{equation}
    \item \textbf{Update Policy:} Update the policy parameters to \(\theta'\) (lines 9-11).
\end{enumerate}

\begin{algorithm}[t]
\caption{TRPO}
\begin{algorithmic}[1]
\State \textbf{Input:} initial policy parameters \(\theta_0\)
\For{\(k = 0, 1, 2, \dots\)}
    \State \parbox[t]{\dimexpr\linewidth-\algorithmicindent}{ Collect set of trajectories \(\mathcal{D}_k\) on policy \(\pi_k = \pi(\theta_k)\)\strut}
    \State \parbox[t]{\dimexpr\linewidth-\algorithmicindent}{ Estimate advantages \(\hat{A}^{\pi_k}_t\)\ using any advantage estimation algorithm\strut}
    \State Form sample estimates for:
    \State \parbox[t]{\dimexpr\linewidth-\algorithmicindent}{ policy gradient \(\hat{g}_k\) (using advantage estimates)\strut}
    \State \parbox[t]{\dimexpr\linewidth-\algorithmicindent}{and KL-divergence Hessian-vector product function \(f(v) = \hat{H}_k v\)\strut}
    \State \parbox[t]{\dimexpr\linewidth-\algorithmicindent}{ Use CG with \(n_{cg}\) iterations to obtain \(x_k \approx \hat{H}_k^{-1} \hat{g}_k\)\strut}
    \State Estimate proposed step \(\Delta_k \approx \sqrt{\frac{2\delta}{x_k^T \hat{H}_k x_k}} x_k\)
    \State \parbox[t]{\dimexpr\linewidth-\algorithmicindent}{ Perform backtracking line search with exponential decay to obtain final update\strut}
    \State \(\theta_{k+1} = \theta_k + \alpha^j \Delta_k\)
\EndFor
\end{algorithmic}
\label{alg:TRPO}
\end{algorithm}

Let us analyze a handful of research studies that used TRPO to grasp a better understanding.
A Monotonic Policy Optimization (MPO) algorithm was designed to address the challenges associated with high-dimensional continuous control tasks in \cite{yuan2019monotonic}. The primary focus was on ensuring monotonic improvement in policy performance, which was crucial for stability and efficiency in RL. One of the significant strengths of this paper was its innovative approach to policy optimization. The authors derived a new lower bound on policy improvement that penalized average policy divergence on the state space, rather than the maximum divergence. This approach addressed a critical limitation of previous algorithms, such as TRPO, which could suffer from worst-case degradation in policy performance. By focusing on average divergence, the MPO algorithm ensured more consistent and reliable improvements, making it particularly suitable for high-dimensional continuous control tasks. The empirical evaluation of the MPO algorithm was another strong point. The authors conducted extensive simulations using the MuJoCo physics engine, testing the algorithm on various challenging robot locomotion tasks, including swimming, quadruped locomotion, bipedal locomotion, and more. The results demonstrated that the MPO algorithm consistently outperformed state-of-the-art methods like TRPO and Vanilla Policy Gradient in terms of average discounted rewards. The algorithm's performance in these high-dimensional tasks highlighted its robustness and practical applicability. However, the complexity of the proposed algorithm might have posed challenges for practical implementation. The need for computing the natural policy gradient and performing line searches to determine optimal step sizes required significant computational resources. This complexity could have limited the accessibility of the MPO algorithm for practitioners without advanced computational capabilities. Additionally, while the algorithm guaranteed monotonic improvement, this came at the cost of slower training speed due to conservative step sizes. The authors suggested combining MPO with other faster methods during the initial training phase to mitigate this issue, but this added another layer of complexity to the implementation.

\cite{roostaie2021entrpo} introduced an enhancement to the TRPO algorithm by incorporating entropy regularization. This modification, termed EnTRPO, aimed to improve exploration and generalization by encouraging more stochastic policy choices. The paper demonstrated the effectiveness of EnTRPO through experiments on the CartPole system, showcasing better performance compared to the original TRPO. One of the main strengths of the paper was its innovative use of entropy regularization to enhance the TRPO algorithm. By adding an entropy term to the advantage function, the authors effectively encouraged exploration, which was crucial to avoid premature convergence to suboptimal policies. This approach addressed a common limitation of TRPO, which could sometimes restrict exploration due to its strict KL divergence constraints between consecutive policies. The entropy regularization helped maintain a balance between exploration and exploitation, leading to more robust learning outcomes. The empirical evaluation provided in the paper was another significant strength. The authors conducted thorough experiments using the Cart-Pole system, a well-known benchmark in the field. The results showed that EnTRPO converged faster and more reliably than TRPO, particularly when the discount factor was set to 0.85. This indicated that the proposed method not only improved exploration but also enhanced the overall convergence speed and stability of the learning process. The use of a well-defined experimental setup, including details on neural network architectures and hyperparameters, added credibility to the findings. A potential limitation was the reliance on a single benchmark task for evaluation. While the Cart-Pole system was a standard benchmark, it was relatively simple compared to many real-world applications. The paper would have benefited from additional experiments on more complex tasks and environments to demonstrate the generalizability and robustness of EnTRPO. This would have provided stronger evidence of the method's effectiveness across a wider range of scenarios.

The challenge of applying trust region methods to Multi-agent RL (MARL) was investigated in \cite{kuba2021trust}. The authors introduced Heterogeneous-agent TRPO (HATRPO) and Heterogeneous-Agent Proximal Policy Optimization (HAPPO) algorithms. These methods were designed to guarantee monotonic policy improvement without requiring agents to share parameters or relying on restrictive assumptions about the decomposability of the joint value function. A strength of the paper was its theoretical foundation. The authors extended the theory of trust region learning to cooperative MARL by developing a multi-agent advantage decomposition lemma and a sequential policy update scheme. This theoretical advancement allowed HATRPO and HAPPO to ensure monotonic improvement in joint policy performance, a key advantage over existing MARL algorithms that did not guarantee such improvement. This theoretical guarantee was essential for stable and reliable learning in multi-agent settings, where individual policy updates could often lead to non-stationary environments and suboptimal outcomes. The empirical validation of HATRPO and HAPPO on benchmarks such as Multi-Agent MuJoCo and StarCraft II demonstrated the effectiveness of these algorithms. The results showed that HATRPO and HAPPO significantly outperformed strong baselines, including Independent Proximal Policy Optimization (IPPO), MAPPO, and MADDPG, in various tasks. This performance improvement highlighted the practical applicability of the proposed methods in complex, high-dimensional environments. The thorough experimental evaluation across multiple scenarios provided strong evidence of the robustness and generalizability of HATRPO and HAPPO. However, the complexity of implementing HATRPO and HAPPO could have been a potential limitation. The algorithms required the computation of multi-agent advantage functions and sequential updates, which could have been computationally intensive and challenging to implement efficiently. This complexity might have limited the accessibility of these methods to practitioners who might not have had advanced computational resources or expertise in implementing sophisticated algorithms.

Authors in \cite{liu2018extreme} presented a method to actively recognize objects by choosing a sequence of actions for an active camera. This method utilized TRPO combined with Extreme Learning Machines (ELMs) to enhance the efficiency of the optimization algorithm. One of the significant strengths of this paper was its innovative application of TRPO in conjunction with ELMs. ELMs provided a simple yet effective way to approximate policies, reducing the computational complexity compared to traditional deep neural networks. This resulted in an efficient optimization process, crucial for real-time applications like active object recognition. The use of ELMs allowed for faster convergence and more straightforward implementation, making the proposed method accessible for practical applications. However, the complexity of integrating TRPO with ELMs could have posed challenges for some practitioners. Although ELMs simplified the optimization process, they still required careful tuning of parameters, such as the number of hidden nodes and the distribution of random weights. This additional layer of complexity might have limited the method's accessibility for users without extensive experience in RL and neural networks.

In \cite{santoso2020multiagent}, authors explored the application of the TRPO algorithm in MARL environments, specifically focusing on hide-and-seek games. The authors compared the performance of TRPO with the Vanilla Policy Gradient (VPG) algorithm to determine the most effective method for this type of game. One of the primary strengths of this paper was its focus on a well-defined, complex multi-agent environment. Hide and seek games inherently involved dynamic interactions between agents, making them an excellent testbed for evaluating algorithms. By using TRPO, which was designed to ensure monotonic policy improvement, the authors addressed a significant challenge in MARL: maintaining stable and consistent learning despite the presence of multiple interacting agents. The empirical results presented in the paper highlighted the strengths of TRPO, especially in scenarios where the testing environment differed from the training environment. TRPO's ability to adapt to new environments and maintain high performance was a notable advantage over the VPG algorithm, which performed better in environments identical to the training conditions but struggled when faced with variability. This adaptability was crucial for practical applications of MARL, where agents often encountered unpredictable changes in their environment. Another strength was the comprehensive experimental setup, which included various configurations and scenarios. The authors meticulously compared the performance of TRPO and VPG across different numbers of agents and types of environments (quadrant and random walls scenarios). This thorough approach provided robust evidence supporting the efficacy of TRPO in MARL settings. However, the paper also had some limitations. The complexity of implementing TRPO in a multi-agent context could have been a barrier for practitioners. TRPO required careful tuning and substantial computational resources, which might not have been readily available in all settings. Additionally, the reliance on simulation results raised questions about the real-world applicability of the findings. While the hide-and-seek game was a useful simulation environment, real-world deployments could have presented additional challenges not captured in the simulations.

The application of TRPO to improve Cross-Site Scripting (XSS) detection systems was analyzed by authors in \cite{mondal2023xss}. The authors aimed to enhance the resilience of XSS filters against adversarial attacks by using RL techniques to identify and counter malicious inputs. One of the main strengths of this paper was its innovative approach to applying TRPO in Cybersecurity, specifically for XSS detection. Traditional XSS detection methods often relied on static rules and signatures, which could be easily bypassed by sophisticated attackers. By leveraging TRPO, the authors introduced a dynamic and adaptive mechanism that could learn to detect and counteract adversarial attempts to exploit XSS vulnerabilities. This use of TRPO enhanced the robustness of the detection system, making it more resilient to evolving threats. A limitation of this study was the reliance on specific hyperparameters, such as the learning rate and discount factor, which could have significantly impacted the model's performance. The paper would have benefited from a more detailed discussion on how these parameters were selected and their influence on the detection model. Providing guidelines or heuristics for parameter tuning would have helped practitioners replicate and extend the study's findings.

Authors in \cite{erens2024universal} aimed to create a universal policy for a locomotion task that could adapt to various robot morphologies, using TRPO. The study investigated the use of surrogate models, specifically Polynomial Chaos Expansion (PCE) and model ensembles, to model the dynamics of the robots. One of the primary strengths of this thesis was its innovative approach to developing a universal policy. The use of TRPO ensured stability and reliable policy updates even in complex environments. The focus on creating a policy that could generalize across different robot configurations was particularly noteworthy, as it addressed the challenge of designing controllers that were not limited to a single robot morphology. The integration of surrogate models, especially the PCE, was another strong point. PCE allowed for efficient sampling and modeling of the stochastic environment, potentially reducing the number of interactions required with the real environment. This was crucial for practical applications where real-world interactions could have been costly or risky. The theoretical foundation laid for using PCE in this context was robust and showed promise for future research. However, the thesis also highlighted several challenges and limitations. The complexity of accurately modeling the dynamics with PCE was a significant hurdle. The results indicated that while PCE showed potential, it currently could not model the dynamics accurately enough to be used in combination with TRPO effectively. The computational time required for PCE was also a practical concern, limiting its immediate applicability. The model ensemble surrogate showed some promise but ultimately failed to train a successful policy. This pointed to the difficulty of creating surrogate models that could capture the complexities of robot dynamics sufficiently. The thesis suggested that using the original environment from the RoboGrammar library yielded better results, emphasizing the need for more advanced or alternative surrogate modeling techniques.

A novel approach for optimizing Home Energy Management Systems (HEMS) using Multi-agent TRPO (MA-TRPO) was investigated in \cite{thattai2023consumer}. This approach aimed to improve energy efficiency, cost savings, and consumer satisfaction by leveraging TRPO techniques in a multi-agent setup. One of the primary strengths of this paper was its consumer-centric approach. Traditional HEMS solutions often prioritized energy efficiency and cost savings without adequately considering consumer preferences and comfort. By incorporating a preference factor for Interruptible-Deferrable Appliances (IDAs), the proposed MA-TRPO algorithm ensured that consumer satisfaction was taken into account, leading to a more holistic and practical solution. This consumer-centric focus was crucial for the widespread adoption of HEMS in real-world settings. Another strength was the comprehensive use of real-world data for training and validation. The authors utilized five-minute retail electricity prices derived from wholesale market prices and real-world Photovoltaic (PV) generation profiles. This approach enhanced the practical relevance and robustness of the proposed method, as it demonstrated the algorithm's effectiveness under realistic conditions. Additionally, the paper provided a detailed explanation of the various components of the smart home environment, including the non-controllable base load, IDA, Battery Energy Storage System (BESS), and PV system, which added clarity and depth to the study. The use of MA-TRPO in a multi-agent setup was also a significant contribution. The proposed method modeled and trained separate agents for different components of the HEMS, such as the IDA and BESS, allowing for more specialized and effective control strategies. This multi-agent approach addressed the complexities and inter-dependencies within the home energy environment, leading to more efficient and coordinated energy management. The paper's reliance on simulation results, while comprehensive, still left questions about the real-world applicability of the proposed method. Although the use of real-world data enhanced relevance, further validation in actual home environments would have strengthened the case for practical deployment. Real-world testing was essential to ensure the robustness and effectiveness of the MA-TRPO algorithm in diverse and dynamic home energy scenarios. Additionally, the reliance on discrete action spaces simplified the problem but might not have fully captured the nuances of continuous control in real-world applications. Future work could explore extending the algorithm to handle continuous action spaces for more precise control.

Authors in \cite{peng2022aoi} investigated the application of TRPO to address the joint spectrum and power allocation problem in the Internet of Vehicles (IoV). The objective was to minimize AoI and power consumption, which were crucial for maintaining real-time communication and energy efficiency in vehicular networks. One of the key strengths of this paper was its focus on AoI, a vital metric for ensuring timely and accurate information exchange in vehicular communications. By incorporating AoI into the optimization framework, the authors addressed a significant challenge in IoV networks, where the freshness of information directly impacted road safety and traffic efficiency. The proposed TRPO-based approach effectively balanced the trade-off between minimizing AoI and reducing power consumption, showcasing its practical relevance. The paper's reliance on certain assumptions, such as the availability of CSI and the periodic reporting of CSI to the base station, might have limited its generalizability. In real-world scenarios, obtaining accurate and timely CSI could have been challenging due to various factors like signal interference and mobility. Future work could explore more practical approaches to CSI estimation and reporting to enhance the applicability of the proposed solution.

Table \ref{tab:TRPO_Papers} categorizes the papers reviewed in this section by their domain, offering a summary of the research landscape in TPRO.
An advancement over TRPO was proposed as a new algorithm, PPO.

\begin{table}[t]
\centering
\renewcommand{\arraystretch}{1.2} 
\caption{TRPO Papers Review}
\begin{tabular}{|>{\raggedright\arraybackslash}p{4cm}|>{\raggedright\arraybackslash}p{3cm}|}
\hline
\textbf{Application Domain} & \textbf{References} \\
\hline
Object Recognition & \cite{liu2018extreme}
 \\
\hline
Theoretical Research (Convergence, stability) & \cite{yuan2019monotonic}\\
\hline
Hybrid RL Algorithms & \cite{roostaie2021entrpo}
 \\
\hline
Multi-agent Systems and Autonomous Behaviors & \cite{kuba2021trust}, \cite{santoso2020multiagent} \\
\hline
Cybersecurity & \cite{mondal2023xss} \\
\hline
Robotics & \cite{erens2024universal}
 \\
\hline
Energy and Power Management & \cite{thattai2023consumer}, \cite{iqbal2021double} ,\cite{li2019partially}
 \\
\hline
\end{tabular}
\label{tab:TRPO_Papers}
\end{table}

\subsection{Proximal Policy Optimization (PPO)}

PPO, proposed by \cite{schulman2017proximal}, represents a significant advancement within policy gradient methods. PPO aims to achieve reliable performance and sample efficiency, addressing the limitations of previous policy optimization algorithms such as VPG methods and TRPO.


Using policy gradient methods, the policy parameters are optimized through stochastic gradient ascent by estimating the gradient of the policy. One of the most commonly used policy gradient estimators is:

\begin{equation}
    \hat{g} = \hat{\mathbb{E}}_t \left[ \nabla_\theta \log \pi_\theta(a_t | s_t) \hat{A}_t \right],
\end{equation}

where \(\pi_\theta\) represents the policy parameterized by \(\theta\), and \(\hat{A}_t\) is an estimator of the advantage function at time step \(t\). This estimator helps construct an objective function whose gradient corresponds to the policy gradient estimator:

\begin{equation}
    L_{PG}(\theta) = \hat{\mathbb{E}}_t \left[ \log \pi_\theta(a_t | s_t) \hat{A}_t \right].
\end{equation}

PPO simplifies TRPO by using a surrogate objective with a clipped probability ratio, allowing for multiple epochs of mini-batch updates. In order to preserve learning, large policy updates should be avoided. As a result, the PPO objective is as follows:

\begin{equation}
\begin{aligned}
L_{CLIP}(\theta) = \hat{\mathbb{E}}_t \Big[ \min \big( & r_t(\theta) \hat{A}_t, \\
& \text{clip}(r_t(\theta), 1 - \epsilon, 1 + \epsilon) \hat{A}_t \big) \Big]
\end{aligned}
\end{equation}
where \( r_t(\theta) = \frac{\pi_\theta(a_t | s_t)}{\pi_{\theta_{\text{old}}}(a_t | s_t)} \) is the probability ratio, and \(\epsilon\) is a hyperparameter. This objective clips the probability ratio to ensure it stays within a reasonable range, preventing excessively large updates.
%
%
The practical implementation of PPO involves the following steps:

\begin{enumerate}
    \item \textbf{Sample Trajectories:} Collect trajectories using the current policy \(\pi_{\theta_{\text{old}}}\).
    \item \textbf{Estimate Advantages:} Compute the advantage function \(\hat{A}_t\) using the collected trajectories.
    \item \textbf{Optimize Surrogate Objective:} Perform several epochs of optimization on the surrogate objective \(L_{CLIP}(\theta)\) using minibatch stochastic gradient descent.
    \item \textbf{Update Policy:} Update the policy parameters to the new parameters \(\theta\).
\end{enumerate}

\begin{algorithm}[t]
\caption{PPO}
\label{alg:PPO}
\begin{algorithmic}[1]
    \State Initialize policy parameters \( \theta \), and value function parameters \( \phi \)
    \State Initialize old policy \( \pi_{\theta_{\text{old}}} \leftarrow \pi_{\theta} \)
    \For{iteration \( i = 1 \) to \( M \)}
        \State \parbox[t]{\dimexpr\linewidth-\algorithmicindent}{ Collect trajectories \( \{ (s_t, a_t, r_t, s_{t+1}) \} \) by running policy \( \pi_{\theta_{\text{old}}} \)\strut}
        \State \parbox[t]{\dimexpr\linewidth-\algorithmicindent}{ Compute advantage estimates \( \hat{A}_t \) for each trajectory using the value function\strut}
        \For{epoch \( j = 1 \) to \( K \)}
            \State Compute probability ratio \( r_t(\theta) = \frac{\pi_\theta(a_t | s_t)}{\pi_{\theta_{\text{old}}}(a_t | s_t)} \)
            \State \parbox[t]{\dimexpr\linewidth-\algorithmicindent}{ Define the surrogate loss:
            \begin{align}
            \nonumber L_{CLIP}(\theta) = \hat{\mathbb{E}}_t \Big[ & \min \left( r_t(\theta) \hat{A}_t, \right. \nonumber \\
            & \left. \nonumber \text{clip}(r_t(\theta), 1 - \epsilon, 1 + \epsilon) \hat{A}_t \right) \Big]
            \end{align}\strut}

            \State \parbox[t]{\dimexpr\linewidth-\algorithmicindent}{ Perform mini-batch gradient ascent on the \\
            surrogate objective \( L_{CLIP}(\theta) \)\strut}
            \State \parbox[t]{\dimexpr\linewidth-\algorithmicindent}{ Update the value function parameters \( \phi \) by \\
            minimizing:
            \[
            L_V(\phi) = \hat{\mathbb{E}}_t \left[ \left( V_\phi(s_t) - R_t \right)^2 \right]
            \]\strut}
        \EndFor
        \State Update old policy: \( \pi_{\theta_{\text{old}}} \leftarrow \pi_\theta \)
    \EndFor
\end{algorithmic}
\end{algorithm}

Before delving into related papers, to illustrate the algorithm's functionality, PPO's algorithm is provided in Alg. \ref{alg:PPO}.
A DRL approach for optimizing traffic flow in mixed-autonomy scenarios, where both Connected Autonomous Vehicles (CAVs) and human-driven vehicles coexisted, was introduced in \cite{wei2019mixed}. The authors proposed three distributed learning control policies for CAVs using PPO, a policy gradient DRL method, and conducted experiments with varying traffic settings and CAV penetration rates on the Flow framework, a new open-source microscopic traffic simulator. One of the primary strengths of the paper was its innovative approach to mixed-autonomy traffic optimization at a network level. The authors hypothesized that controlling distributed CAVs at a network level could outperform individually controlled CAVs, and their experimental results supported this hypothesis. The network-level RL policies for controlling CAVs significantly improved the total rewards and average velocity compared to individual-level RL policies. This finding was crucial for the development of ITS which aimed to optimize traffic flow and reduce congestion. The use of the Flow framework and SUMO environment for experiments was another strength of the paper. These tools provided a realistic and flexible simulation environment for testing the proposed control policies. The comprehensive evaluation, including different traffic settings and CAV penetration rates (10\%, 20\%, and 30\%), added robustness to the findings. The authors’ use of three different learning strategies—single-agent asynchronous learning, joint global cooperative learning, and joint local cooperative learning—allowed for a thorough comparison of the effectiveness of network-level control policies. The paper also highlighted the potential high communication overhead associated with the global joint cooperative policy, especially as the penetration rate of CAVs increased. This overhead could have limited the scalability of the solution in real-world scenarios where communication resources were constrained. The authors suggested that the local joint cooperative policy, which required less communication, could have been a more practical choice in such situations, though it might not have performed as well as the global policy.

In \cite{zhang2020real}, authors presented a novel method for managing power grid line flows in real-time using PPO. This approach addressed the challenges posed by the increasing penetration of renewable resources and the unpredictability of power system conditions, aiming to prevent line overloading and potential cascading outages. One of the key strengths of this paper was the application of PPO in a critical real-time grid operation context. PPO was well-suited for this task due to its ability to ensure stable updates within trust regions, thus maintaining the reliability of the control policies. This characteristic was crucial in power systems, where stability and safety were paramount. The authors’ decision to use PPO over other algorithms like DDPG and Policy Gradient was justified by PPO’s robustness in handling larger and more complex action spaces. A limitation was the reliance on simulation results for validation. While the use of high-fidelity power grid simulators was a strength, real-world deployments could present additional challenges not captured in simulations. Factors such as communication delays, unexpected equipment failures, and human operator interventions could have impacted the effectiveness of the control strategies. Further real-world testing and validation were necessary to fully assess the robustness and practicality of the proposed method in diverse and dynamic power grid environments.

A framework for optimizing metro service schedules and train compositions using PPO within a DRL framework was proposed in \cite{ying2021adaptive}. This method was applied to handle the dynamic and complex problem of metro train operations, focusing on minimizing operational costs and improving service regularity. A significant strength of the paper was its innovative application of PPO to metro service scheduling and train composition. PPO, known for its stability and efficiency in handling large-scale optimization problems, was effectively utilized to address the dynamic nature of metro operations. The integration of PPO with Artificial Neural Networks (ANNs) for approximating value functions and policies demonstrated a robust approach to tackling the high-dimensional decision space inherent in metro scheduling. This combination enhanced the framework's ability to adapt to varying passenger demands and operational constraints. The paper’s use of a real-world scenario, specifically the Victoria Line of the London Underground, for testing and validation was another strong point. The authors provided a comprehensive evaluation, comparing their PPO-based method with established meta-heuristic algorithms like Genetic Algorithm and Differential Evolution. The results indicated that the PPO-based approach outperformed these traditional methods in terms of both solution quality and computational efficiency. This practical validation underscored the method’s applicability and effectiveness in real-world settings. The reliance on a specific set of operational constraints, such as fixed headways and trailer limits, might have also limited the method’s generalizability. While these constraints were necessary for practical implementations, exploring more flexible constraint formulations could have further enhanced the method’s adaptability to different metro systems and operational conditions.

Authors in \cite{jin2021optimal} proposed an advanced scheduling algorithm for multitask environments using a combination of PPO and an optimal policy characterization. The main focus was to optimize the scheduling of multiple tasks across multiple servers, taking into account random task arrivals and renewable energy generation. One of the primary strengths of this paper was its innovative combination of PPO with a priority rule, the Earlier Deadline and Less Demand First (ED-LDF). This rule prioritized tasks with earlier deadlines and lower demands, which was shown to be optimal under heavy traffic conditions. The integration of ED-LDF with PPO effectively reduced the dimensionality of the action space, making the algorithm scalable and efficient even in large-scale settings. This reduction in complexity was crucial for practical applications where the number of tasks and servers could have been substantial. However, the complexity of implementing the proposed method might have posed challenges. The integration of the ED-LDF rule with PPO required careful tuning and a deep understanding of both RL and optimal scheduling principles. This complexity could have limited the accessibility of the method for practitioners who might not have had advanced expertise in these areas. Additionally, the reliance on heavy-traffic assumptions for the optimality of the ED-LDF rule might have limited the generalizability of the approach to all scheduling environments. Real-world scenarios could have presented varying traffic conditions that did not always align with these assumptions. The paper's focus on renewable energy as a factor in the scheduling decision was another strength, as it aligned with the growing importance of sustainability in computing operations. However, the paper could have benefited from further exploration of how variations in renewable energy availability impacted the scheduling performance. This aspect was crucial for real-world applications where renewable energy sources could have been highly variable.

An innovative approach to enhancing image captioning models using PPO was designed by authors in \cite{zhang2021image}. The authors aimed to improve the quality of generated captions by incorporating PPO into the phase of training, specifically targeting the optimization of scores. The study explored various modifications to the PPO algorithm to adapt it effectively for the image captioning task. A significant strength of the paper was its integration of PPO with image captioning models, which traditionally relied on VPG methods. The authors argued that PPO could provide better performance due to its ability to enforce trust-region constraints, thereby improving sample complexity and ensuring stable policy updates. This was particularly important for image captioning, where maintaining high-quality training trajectories was crucial. The authors’ experimentation with different regularization techniques and baselines was another strong point. They found that combining PPO with dropout decreased performance, which they attributed to increased KL-divergence of RL policies. This empirical observation was critical as it guided future implementations of PPO in similar contexts. Furthermore, the adoption of a word-level baseline via MC estimation, as opposed to the traditional sentence-level baseline, was a noteworthy innovation. This approach was expected to reduce the variance of policy gradient estimators more effectively, contributing to improved model performance. While the results were promising, they were primarily validated on the MSCOCO dataset. Further validation on other datasets and in real-world applications would have been beneficial to assess the generalizability and robustness of the approach. The paper could have also benefited from a more detailed discussion on the impact of different hyperparameter settings and the specific configurations used in the experiments. Providing this information would have enhanced the reproducibility of the study and allowed other researchers to build on the authors’ work more effectively.

In \cite{guan2020centralized}, a centralized coordination scheme for CAVs at intersections without traffic signals was developed. The authors introduced the Model Accelerated PPO (MA-PPO) algorithm, which incorporated a prior model into the PPO algorithm to enhance sample efficiency and reduce computational overhead. One of the significant strengths of this paper was its focus on improving computational efficiency, a major challenge in centralized coordination methods. Traditional methods, such as MPC, were computationally demanding, making them impractical for real-time applications with a large number of vehicles. By using MA-PPO, the authors significantly reduced the computation time required for coordination, achieving an impressive reduction to 1/400 of the time needed by MPC. This efficiency gain was crucial for real-time deployment in busy intersections. A limitation was the focus on a specific intersection scenario. While the four-way single-lane intersection was a common setup, real-world intersections could have varied significantly in complexity and traffic patterns. Future work could have explored the applicability of MA-PPO to more complex intersection scenarios and different traffic conditions to ensure broader generalizability.

The application of PPO in developing a DRL controller for the nonlinear attitude control of fixed-wing UAVs was explored in \cite{bohn2019deep}. The study presented a proof-of-concept controller capable of stabilizing a fixed-wing UAV from various initial conditions to desired roll, pitch, and airspeed values. One of the primary strengths of the paper was its innovative use of PPO for UAV attitude control. PPO was known for its stability and efficient policy updates, making it well-suited for complex control tasks like UAV attitude stabilization. The choice of PPO over other RL algorithms was justified by its robust performance and low computational complexity, which were crucial for real-time control applications. The authors also highlighted the practical advantages of using PPO, such as its hyperparameter robustness across different tasks. One of the limitations was the reliance on a single UAV model and specific aerodynamic coefficients for validation. While the Skywalker-X8 was a popular fixed-wing UAV, the generalizability of the proposed approach to other UAV models with different aerodynamic characteristics remained to be explored. Future work could have benefited from testing the PPO-based controller on a wider range of UAV models to ensure broader applicability.

The use of PPO to control the position of a quadrotor was investigated in \cite{lopes2018intelligent}. The primary goal was to achieve stable flight control without relying on a predefined mathematical model of the quadrotor’s dynamics. One of the major strengths of this paper was its application of PPO in the context of quadrotor control. By using PPO, the authors ensured that the control policy updates remained stable and efficient, which was crucial for real-time applications like quadrotor control. The choice of PPO over other RL algorithms was well justified due to its robustness and low computational complexity. The authors’ approach to utilizing a stochastic policy gradient method during training, which was then converted to a deterministic policy for control, was another notable strength. This strategy ensured efficient exploration during training, allowing the quadrotor to learn a robust control policy. The use of a simple reward function that focused on minimizing the position error between the quadrotor and the target further added to the efficiency of the training process. One limitation was the reliance on specific initial conditions and a fixed simulation environment. While the authors showed that the PPO-based controller could recover from harsh initial conditions, the generalizability of the method to different quadrotor models and varying environmental conditions remained to be explored. Future work could have benefited from testing the controller on a wider range of scenarios and incorporating additional environmental factors to ensure broader applicability.

Authors in \cite{ye2020automated} addressed the development of an intelligent lane change strategy for autonomous vehicles using PPO. This approach PPO to manage lane change maneuvers in dynamic and complex traffic environments, focusing on enhancing safety, efficiency, and comfort. The authors’ design of a comprehensive reward function that considered multiple aspects of lane change maneuvers was one of the strengths. The reward function incorporated components for safety (avoiding collisions and near-collisions), efficiency (minimizing travel time and aligning with desired speed and position), and comfort (reducing lateral and longitudinal jerks). This multi-faceted approach ensured that the learned policy optimized for a holistic driving experience, balancing the often competing demands of these different aspects. The inclusion of a safety intervention module to prevent catastrophic actions was a particularly noteworthy feature. This module labeled actions as "catastrophic" or "safe" and could replace potentially dangerous actions with safer alternatives, enhancing the robustness of the learning process. This safety-centric approach addressed a critical concern in applying DRL to real-world autonomous driving tasks, where safety was paramount. However, the complexity of implementing PPO for lane change maneuvers posed challenges. The need for continuous training and fine-tuning of parameters could have been resource-intensive and might not have been feasible for all developers or organizations, acknowledging that it is true for some other algorithms. Table \ref{tab:PPO_Papers} shows a summary of analyzed papers. Moreover, Table \ref{tab:Policy_Papers} represents all papers reviewed in this section as a package.

\begin{table}[t]
\centering
\renewcommand{\arraystretch}{1.2} 
\caption{PPO Papers Review}
\begin{tabular}{|>{\raggedright\arraybackslash}p{4cm}|>{\raggedright\arraybackslash}p{3cm}|}
\hline
\textbf{Application Domain} & \textbf{References} \\
\hline
Distributed DRL Control for Mixed-Autonomy Traffic Optimization & \cite{wei2019mixed} \\
\hline
Power Systems and Energy Management & \cite{zhang2020real}\\
\hline
Transportation and Routing Optimization (EVs) & \cite{ying2021adaptive} \\
\hline
Real-time Systems and Hardware & \cite{jin2021optimal} \\
\hline
Image Captioning Models & \cite{zhang2021image} \\
\hline
Hybrid RL Algorithms & \cite{zhang2021image}
 \\
\hline
Intelligent Traffic Signal Control & \cite{guan2020centralized}
 \\
\hline
Real-time Systems and Hardware & \cite{bohn2019deep}, \cite{lopes2018intelligent}
 \\
\hline
\end{tabular}
\label{tab:PPO_Papers}
\end{table}

\begin{table}[t]
\centering
\renewcommand{\arraystretch}{1.2} 
\caption{Policy-based Papers Review}
\begin{tabular}{|>{\raggedright\arraybackslash}p{4cm}|>{\raggedright\arraybackslash}p{3cm}|}
\hline
\textbf{Application Domain} & \textbf{References} \\
\hline
Distributed DRL Control for Mixed-Autonomy Traffic Optimization & \cite{wei2019mixed} \\
\hline
Power Systems and Energy Management & \cite{zhang2020real}\\
\hline
Transportation and Routing Optimization (EVs) & \cite{ying2021adaptive} \\
\hline
Real-time Systems and Hardware & \cite{jin2021optimal}, \cite{bohn2019deep}, \cite{lopes2018intelligent} \\
\hline
Image Captioning Models & \cite{zhang2021image} \\
\hline
Hybrid RL Algorithms & \cite{zhang2021image}, \cite{roostaie2021entrpo}
 \\
\hline
Intelligent Traffic Signal Control & \cite{guan2020centralized}
 \\
\hline
Energy and Power Management & \cite{guo2023function}, \cite{thattai2023consumer}, \cite{iqbal2021double} ,\cite{li2019partially}  \\
\hline
Theoretical Research (Convergence, stability) & \cite{zhang2021sample}, \cite{yuan2019monotonic} \\
\hline
Network Optimization & \cite{tao2024reinforcement}
 \\
\hline
Object Recognition & \cite{liu2018extreme}
 \\
\hline
Multi-agent Systems and Autonomous Behaviors & \cite{kuba2021trust}, \cite{santoso2020multiagent} \\
\hline
Cybersecurity & \cite{mondal2023xss} \\
\hline
Robotics & \cite{erens2024universal}, \cite{weerakoon2022htron}
 \\
\hline
\end{tabular}
\label{tab:Policy_Papers}
\end{table}

We now shall analyze the last group of methods, the Actor-Critic methods. We start by introducing the general Actor-Critics, which combine Value-based and Policy-based approaches.

\section{Actor-Critic Methods}\label{sec:ACM}

Actor-critic methods combine Value-based and Policy-based approaches. Essentially, these methods consist of two components: the Actor, who selects actions based on a policy, and the Critic, who evaluates the actions based on their value function. By providing feedback on the quality of the actions taken, the critic guides the actor in updating the policy directly. As a result of this synergy, learning can be more stable and efficient, addressing some limitations of pure policy or Value-based approaches \cite{grondman2012survey, arulkumaran2017deep}.

In the next subsection, we first introduce two main versions of Actor-Critic methods, Asynchronous Advantage Actor-Critic (A3C) \& Advantage Actor-Critic (A2C), and then, we will analyze various applications of each.

\subsection{A3C \& A2C}

The A2C algorithm is a synchronous variant of the A3C algorithm, which was introduced by \cite{mnih2016asynchronous}. A2C  maintains the key principles of A3C but simplifies the training process by synchronizing the updates of multiple agents, thereby leveraging the strengths of both Actor-Critic methods and advantage estimation.
The Actor-Critic architecture combines two primary components, in both algorithms: the actor, which is responsible for selecting actions, and the critic, which evaluates the actions by estimating the value function. The actor updates the policy parameters in a direction that is expected to increase the expected reward, while the critic provides feedback by computing the TD error. This integration allows for more stable and efficient learning compared to using Actor-only or critic-only methods \cite{konda1999actor}.
Advantage estimation is a technique used to reduce the variance of the policy gradient updates. The advantage function \(A(s, a)\) represents the difference between the action-value function \(Q(s, a)\) and the value function \(V(s)\):

\begin{equation}
    A(s, a) = Q(s, a) - V(s).
\end{equation}

By using the advantage function, A2C focuses on actions that yield higher returns than the average, which helps in making more informed updates to the policy \cite{sutton2018reinforcement}.
Unlike A3C, where multiple agents update the global model asynchronously, A2C synchronizes these updates. Multiple agents run in parallel environments, collecting experiences and calculating gradients, which are then aggregated and used to update the global model synchronously. This synchronization reduces the complexity of implementation and avoids issues related to asynchronous updates, such as non-deterministic behavior and potential overwriting of gradients. 

The A3C algorithm operates as follows:

Based on Alg. \ref{alg:A3C}, first, the parameters of the policy network (actor) \(\theta\) and the value network (critic) \(\phi\) are initialized (line 1). Then, multiple agents are run in parallel, each interacting with its own copy of the environment (lines 2-14). Each agent independently collects a batch of experiences \((s_t, a_t, r_t, s_{t+1})\) (lines 5-9). For each agent, the advantage is computed using the collected experiences. Subsequently, the gradients of the policy and value networks are calculated using the advantage estimates. \textbf{Finally, each agent independently updates the global model parameters \(\theta\) and \(\phi\) asynchronously using the computed gradients (line 13).}

\begin{algorithm}[t]
\caption{A3C}
\begin{algorithmic}[1]
\State Initialize actor and critic networks with random weights
\For{each episode \(\in [1, n]\)}
    \State \parbox[t]{\dimexpr\linewidth-\algorithmicindent}{ Download weights from the headquarters to each AC\strut}
    \For{each AC}
        \State Initialize the random state \(s_0\)
        \For{each \(t \in [1, k]\)}
            \State Select action \(a_t\) from the actor network
            \State \parbox[t]{\dimexpr\linewidth-\algorithmicindent}{ Execute action \(a_t\) and observe reward \\
            \(r_t\) from the critic network and next state \\
            \(s_t\)\strut}
            \State Update the actor network parameters
        \EndFor
        \State Update the critic network parameters
    \EndFor
    \State \parbox[t]{\dimexpr\linewidth-\algorithmicindent}{ Upload weights of each AC network to the headquarters\strut}
\EndFor
\end{algorithmic}
\label{alg:A3C}
\end{algorithm}

The A2C algorithm, on the other hand, operates as follows:
Based on Alg. \ref{alg:A2C}, first, the parameters of the policy network (actor) \(\theta\) and the value network (critic) \(\phi\) are initialized (line 1). Then, multiple agents are run in parallel, each interacting with its own copy of the environment (line 4). Each agent collects a batch of experiences \((s_t, a_t, r_t, s_{t+1})\) (line 10). For each agent, the advantage is computed using the collected experiences. Subsequently, the gradients of the policy and value networks are calculated using the advantage estimates (lines 13-15). \textbf{Finally, the gradients from all agents are aggregated and used to update the global model parameters \(\theta\) and \(\phi\) (lines 17-18).}

\begin{algorithm}[t]
\caption{A2C}
\begin{algorithmic}[1]
\State Initialize policy network (actor) parameters \(\theta\) and value network (critic) parameters \(\phi\)
\State Set number of parallel agents \(N\)
\Repeat
    \For{each agent \(i \gets 1\) to \(N\) \textbf{in parallel}}
        \State Get initial state \(s_i\)
        \State Initialize local episode storage for agent \(i\)
        \For{each step \(t\)}
            \State Sample action \(a_i\) from policy \(\pi_\theta(a_i | s_i)\)
            \State \parbox[t]{\dimexpr\linewidth-\algorithmicindent}{ Execute \(a_i\), observe reward \(r_i\) and next \\
            state \(s_i'\)\strut}
            \State \parbox[t]{\dimexpr\linewidth-\algorithmicindent}{ Store \((s_i, a_i, r_i, s_i')\) in local storage for \\
            agent \(i\)\strut}
            \State \(s_i \gets s_i'\)
        \EndFor
        \State Compute advantage estimates \(\hat{A}_i\) for agent \(i\)
        \State \parbox[t]{\dimexpr\linewidth-\algorithmicindent}{ and policy gradient \(\nabla_\theta L_{\text{PG}}(\theta)\) for agent $i$\strut}
        \State and value loss \(\nabla_\phi L_V(\phi)\) for agent \(i\)
    \EndFor
    \State Aggregate gradients from all agents
    \State \parbox[t]{\dimexpr\linewidth-\algorithmicindent}{ Update global actor parameters \(\theta\) and critic parameters \(\phi\) using aggregated gradients\strut}
\Until{convergence or maximum steps reached}
\end{algorithmic}
\label{alg:A2C}
\end{algorithm}

\paragraph{Overview of A3C applications in the literature}
\cite{mangalampalli2024multi} introduced a scheduler named MOPTSA3C, which prioritized tasks and virtual machines based on various factors such as task length, runtime processing capacities, and electricity unit costs. This approach aimed to optimize makespan, resource utilization, and resource cost using an enhanced A3C. One of the significant strengths of this paper was its comprehensive multi-objective approach. By addressing multiple objectives simultaneously, including minimizing makespan, optimizing resource utilization, and reducing resource costs, the proposed scheduler ensured a balanced and efficient task scheduling process in multi-cloud environments. The use of an improved A3C algorithm enhanced the robustness and efficiency of the scheduling process. The asynchronous nature of A3C allowed for parallel training and faster convergence, which was crucial for dynamic and large-scale cloud environments. The scheduler's ability to prioritize tasks and virtual machines based on multiple factors led to more informed and effective scheduling decisions. Cloud environments were inherently dynamic and could exhibit unpredictable changes in workload, resource availability, and cost structures. The proposed method might have needed further enhancements to adapt to these real-time variations effectively and ensure robust performance under fluctuating conditions. While the paper presented a significant advancement in task scheduling for cloud computing, addressing the computational complexity, scalability, and real-world implementation challenges would have been crucial for its practical adoption and effectiveness in diverse cloud environments.

A latency-oriented cooperative caching strategy using A3C was proposed in \cite{jiang2023asynchronous}. This approach aimed to minimize average download latency by predicting content popularity and optimizing caching decisions in real-time, thereby improving user experience in Fog Radio Access Networks (F-RANs). One strength was the comprehensive evaluation of the proposed method through extensive simulations. The authors compared their A3C-based approach with several baseline algorithms, including greedy caching, random caching, and a cooperative caching strategy that did not consider user equipment caching capacity. The results demonstrated significant reductions in average download latency, highlighting the effectiveness of the proposed strategy in optimizing caching performance. Another limitation was the reliance on accurate and timely content popularity predictions. The effectiveness of the caching strategy depended heavily on the accuracy of these predictions. In real-world applications, user preferences and content popularity could change unpredictably, and any deviations from the predicted values could negatively impact the performance of the caching algorithm. Additionally, the assumption that users did not request the same content repeatedly might not have held true in all scenarios, potentially affecting the reliability of the popularity prediction model.

Authors in \cite{du2021resource} presented a comprehensive approach to optimizing resource allocation and pricing in Mobile Edge Computing (MEC)-enabled blockchain systems using the A3C algorithm. The study's strengths lay in its innovative integration of blockchain with MEC to enhance resource management. The A3C algorithm's capability to handle both continuous and high-dimensional discrete action spaces made it well-suited for the dynamic nature of MEC environments. The use of prospect theory to balance risks and rewards based on miner preferences added a nuanced understanding of real-world scenarios. The results demonstrated that the proposed A3C-based algorithm outperformed baseline algorithms in terms of total reward and convergence speed, indicating its effectiveness in optimizing long-term performance. However, the paper had several limitations. The reliance on a specific MEC server configuration and a fixed number of mobile devices might have limited the generalizability of the findings to other settings with different configurations. The assumption that all validators were honest simplified the model but might not have reflected real-world blockchain environments where malicious actors could have existed. The additional complexity introduced by the collaborative local MEC task processing mode might have increased the computational overhead, potentially affecting the scalability of the proposed solution. Moreover, the paper did not address the potential impact of network latency and varying network conditions on the performance of the A3C algorithm, which could have been significant in practical deployments.

In \cite{joypriyanka2023chess}, authors explored the application of A3C to create a cognitive agent designed to help Alzheimer's patients play chess. The primary goal was to enhance cognitive skills and boost brain activity through chess, a game known for its cognitive benefits. One of the key strengths of the paper was its innovative approach to leveraging A3C to assist individuals with cognitive disabilities in playing chess. The cognitive agent provided real-time assistance by suggesting offensive and defensive moves, thereby helping players improve their strategies and cognitive abilities. This approach addressed the gap in traditional AI chess opponents, which did not educate players on strategies and tactics. The cognitive agent relied on accurate feedback for consistency. Agent effectiveness was heavily dependent on its ability to provide relevant and timely suggestions. The agent needed to accurately interpret the player's intentions and provide appropriate feedback in real-world applications. Users who had difficulty navigating digital interfaces might also have experienced accessibility issues due to the agent's reliance on digital interfaces. Another concern was scalability. A controlled environment showed promising results, but the system might not have been able to handle a broader range of cognitive impairments. Additional users and interactions could have introduced additional complexity, making it difficult to maintain performance.

Researchers in \cite{tiong2022autonomous} introduced a novel approach to autonomous valet parking using a combination of PPO and A3C. This method aimed to address the control errors due to the non-linear dynamics of vehicles to optimize parking maneuvers. An important strength of this study was the use of the A3C-PPO algorithm. Combining the advantages of both PPO and A3C, this hybrid approach resulted in a more stable and efficient learning process. A3C's asynchronous nature allowed parallel training, which sped convergence and improved state-action exploration. In addition, PPO prevented drastic changes from destabilizing the learning process, by limiting the magnitude of policy updates. Incorporating manual hyperparameter tuning further optimized the training process, resulting in better rewards. One of the limitations was the reliance on specific assumptions about the environment and sensor accuracy. The effectiveness of the proposed method depended on the accurate detection and interpretation of the surroundings by sensors such as cameras and LiDAR. Any inaccuracies or deviations in sensor data could have impacted the performance and robustness of the algorithm. In real-world applications, environmental conditions such as lighting, weather, and obstacles could have varied significantly, and ensuring reliable sensor performance under these conditions was crucial. Additionally, the study did not address the potential impact of network latency and communication issues between the vehicle and the central control system, which could have affected the real-time decision-making process. Ensuring robust communication and minimizing latency was critical for the practical implementation of autonomous valet parking systems.

An advanced approach for optimizing content caching in 5G networks using A3C was proposed in \cite{shi2019content}. This method aimed to minimize the total transmission cost by learning optimal caching and sharing policies among cooperative Base Stations (BSs) without prior knowledge of content popularity distribution. One of the main strengths of this paper was the use of the A3C algorithm, which leveraged the asynchronous nature of multiple agents to achieve faster convergence and reduce time correlation in learning samples. The algorithm's ability to operate with multiple environment instances in parallel enhanced computational efficiency and significantly improved the learning process. By considering cooperative BSs that could have fetched content from neighboring BSs or the backbone network, the proposed method effectively reduced data traffic and transmission costs in 5G networks. The empirical results demonstrated the superiority of the A3C-based algorithm over classical caching policies such as Least Recently Used, Least Frequently Used, and Adaptive Replacement Cache, showcasing lower transmission costs and faster convergence rates. One limitation of this study was the reliance on accurate and timely updates of content popularity distributions. While the paper assumed that content popularity followed a Zipf distribution and varied over time, the accuracy of these assumptions could have significantly impacted the performance of the caching algorithm. In real-world applications, user preferences and content popularity could have changed unpredictably, and any deviations from the assumed distribution could have affected the effectiveness of the caching policy. Ensuring robust performance under varying content popularity distributions was crucial for the practical implementation of the proposed method.

Table \ref{tab:A3C_Papers} organizes the papers discussed in this section, offering a domain-specific breakdown of the research conducted in the A3C area.

\begin{table}[t]
\centering
\renewcommand{\arraystretch}{1.2} 
\caption{A3C Papers Review }
\begin{tabular}{|>{\raggedright\arraybackslash}p{4cm}|>{\raggedright\arraybackslash}p{3cm}|}
\hline
\textbf{Application Domain} & \textbf{References} \\
\hline
Cloud-based Control and Encryption Systems & \cite{mangalampalli2024multi}, \cite{du2021resource}
 \\
\hline
Games & \cite{joypriyanka2023chess}\\
\hline
Multi-agent Systems and Autonomous Behaviors & \cite{tiong2022autonomous}
 \\
\hline
Network Optimization & \cite{shi2019content}
 \\
\hline
\end{tabular}
\label{tab:A3C_Papers}
\end{table}

\paragraph{Overview of A2C applications in the literature}

In \cite{yang2024ha}, authors introduced an innovative approach to low-latency task scheduling in edge computing environments, addressing several significant challenges inherent in such settings. The primary focus of the paper was on integrating a hard attention mechanism with the A2C algorithm to enhance task scheduling efficiency and reduce latency. The strengths of the HA-A2C method lay in its ability to significantly reduce task latency by approximately 40\% compared to the DQN method. The hard attention mechanism employed by HA-A2C was particularly effective in reducing computational complexity and increasing efficiency, allowing the model to process tasks more quickly. Additionally, the method showcased improved scalability, maintaining low task latency even as the number of tasks increased. The use of the A2C algorithm, which combined policy gradient and value function estimates, enhanced the stability and effectiveness of the policy network, further contributing to the overall performance of the model. However, there were some limitations to the HA-A2C approach. One notable weakness was the potential complexity of implementing the hard attention mechanism in real-world scenarios, where the dynamic and heterogeneous nature of edge environments might have posed additional challenges. Furthermore, while the HA-A2C method outperformed other DRL methods in terms of latency, it might have still faced difficulties in scenarios with extremely high-dimensional states and action spaces, where further optimization might have been necessary. Another consideration was the reliance on accurate and timely data for effective attention allocation, which might not have always been feasible in practical applications.

A task scheduling mechanism in cloud-fog environments that leveraged the A2C algorithm was presented in \cite{choppara2024reliability}. This approach aimed to optimize the scheduling process for scalability, reliability, trust, and makespan efficiency. One of the significant strengths of the paper was the holistic approach it took toward task scheduling in heterogeneous cloud-fog environments. The use of the A2C algorithm was particularly effective in handling the dynamic nature of task scheduling, as it allowed the scheduler to make real-time decisions based on the current state of the system. By dynamically adjusting the number of virtual machines according to workload demands, the proposed scheduler ensured efficient resource utilization, which was crucial for maintaining system performance under varying conditions. One limitation was the reliance on specific system parameters and assumptions about the environment. The proposed method assumed accurate estimation of factors such as task priorities and VM capacities, which might not have always been feasible in practical applications. Deviations from these assumptions could have impacted the performance and robustness of the scheduling algorithm. Further research could have explored the robustness of the proposed approach under more relaxed assumptions and in diverse real-world scenarios.

Authors in \cite{dantas2023beam} introduced an A2C-learning-based framework for optimizing beam selection and transmission power in mmWave networks. This approach aimed to improve energy efficiency while maintaining coverage in dynamic and complex network environments. A notable strength of the paper was its innovative application of the A2C algorithm for joint optimization of beam selection and transmission power. This dual optimization was particularly effective in addressing the significant challenge of energy consumption in mmWave networks. By leveraging A2C, the proposed method dynamically adjusted beam selection and power levels based on the current state of the network, which was represented by the Signal-to-Interference-plus-Noise Ratio (SINR) values. The use of A2C ensured stable and efficient learning through policy gradients and value function approximations, making it suitable for real-time applications. One of the limitations was the assumption of specific system parameters and environmental conditions. The method assumed accurate estimation of SINR values and predefined beam angles, which might not have always been feasible in practical applications. Deviations from these assumptions could have impacted the performance and robustness of the optimization algorithm. Future research could have explored the robustness of the proposed method under more relaxed assumptions and in diverse real-world scenarios.

Authors explored the use of the A2C algorithm to estimate the power delay profile (PDP) in 5G New Radio environments in \cite{kwon2022learning}. This approach aimed to enhance channel estimation performance by leveraging DRL techniques. A notable strength of this paper was its innovative application of the A2C algorithm to the problem of PDP estimation. By framing the estimation problem within an RL context, the proposed method directly targeted the minimization of Mean Square Error in channel estimation, rather than aiming to approximate an ideal PDP. This pragmatic approach allowed the algorithm to adapt to the inherent approximations and imperfections in practical channel estimation processes, leading to improved performance. However, the complexity of implementing the A2C algorithm in real-world scenarios posed challenges. The need for extensive training and parameter tuning required significant computational resources and expertise in RL, which might not have been readily available in all settings. Additionally, while the simulation results were promising, further validation in real-world deployments was necessary to fully assess the robustness and practicality of the proposed approach. Real-world environments could have introduced additional challenges, such as varying traffic patterns and hardware constraints, which were not fully captured in simulations.

The application of various Actor-Critic algorithms to develop a trading agent for the Indian stock market was investigated in \cite{vishal2021trading}. The study evaluated the performance of PPO, DDPG, A2C, and Twin Delayed DDPG (TD3) algorithms in making trading decisions. One of the primary strengths of this paper was its comprehensive approach to evaluating multiple Actor-Critic algorithms in a real-world financial trading context. By considering different algorithms, the authors provided a broad perspective on the effectiveness of it in stock trading. The use of historical stock data from the Yahoo Finance API, covering a substantial period (2006-2021), ensured that the models were tested against diverse market conditions, enhancing the robustness and reliability of the results. The study's focus on a single market (Indian stock market) might have limited the generalizability of the findings. Future research could have explored the application of these algorithms in different financial markets to ensure broader applicability.

The authors presented an innovative framework for optimizing task segmentation and parallel scheduling in edge computing networks using the A2C algorithm in \cite{sun2022a2c}. The approach focused on minimizing total task execution delay by splitting multiple computing-intensive tasks into sub-tasks and scheduling them efficiently across different edge servers. A key strength of this paper was its holistic approach to task segmentation and scheduling. By jointly optimizing both processes, the proposed method ensured that tasks were not only divided efficiently but also assigned to the most suitable edge servers for processing. This joint optimization was crucial in dynamic edge computing environments, where both computation capacity and task requirements could have varied significantly over time. The use of A2C, allowed the system to adapt to these changes in real-time, enhancing overall system performance. The authors' method of decoupling the complex mixed-integer non-convex problem into more manageable sub-problems was another strength. By first addressing the task segmentation problem and then tackling the sub-tasks parallel scheduling, the paper presented a structured and logical approach to solving the optimization challenge. The introduction of the optimal task split ratio function and its integration into the A2C algorithm further enhanced the efficiency and effectiveness of the proposed solution.

Researchers in \cite{han2019multi} presented an innovative approach to enhancing multi-UAV obstacle avoidance using A2C combined with an experience-sharing mechanism. This method aimed to optimize obstacle avoidance strategies in complex, dynamic environments by sharing positive experiences among UAVs to expedite the training process. One of the key strengths of this paper was the introduction of the experience-sharing mechanism to the A2C algorithm. This mechanism significantly enhanced the efficiency and robustness of the training process by allowing UAVs to share positive experiences. This collective learning approach accelerated the convergence of the algorithm, enabling UAVs to quickly learn effective obstacle avoidance strategies. The experience-sharing mechanism was particularly valuable in multi-agent systems, where individual agents could have benefited from the knowledge gained by others, leading to faster and more robust learning. However, the experience-sharing mechanism relied on consistent and reliable communication between UAVs. In practical applications, communication constraints and network reliability issues could have significantly impacted the effectiveness of this mechanism. Inter-UAV communication latency and packet loss could have led to outdated or incomplete information being shared, thereby reducing the overall efficiency of the learning process. Also, the method assumed a certain level of accuracy in modeling the environment and the dynamic obstacles within it. Any deviations from these assumptions, such as unexpected changes in obstacle behavior or environmental conditions, could have affected the performance and robustness of the algorithm. Real-world environments were inherently unpredictable, and the algorithm must have been tested extensively in diverse scenarios to ensure its reliability.

A robust approach to visual navigation using an Actor-Critic method enhanced with Generalized Advantage Estimation (GAE) was developed by authors in \cite{shao2018visual}. This method demonstrated significant strengths in terms of learning efficiency and stability, as well as effective navigation in complex environments like ViZDoom. One major strength of this approach was its ability to rapidly converge and achieve high performance in both basic and complex visual navigation tasks. By employing the A2C method with GAE, the algorithm reduced variance in policy gradient estimates, leading to more stable learning. This was particularly evident in the ViZDoom health gathering scenarios, where the A2C with GAE agent achieved the highest scores with lower variance compared to other methods. Additionally, the use of multiple processes in the A2C method significantly reduced training time, making it more efficient than traditional DQN approaches. However, the method also had notable limitations. One significant drawback was the high computational cost associated with using multiple processes for training, which might not have been feasible in resource-constrained environments. Furthermore, while the approach performed well in the tested ViZDoom scenarios, its generalizability to other, more diverse environments remained uncertain without further validation. The reliance on visual inputs also presented challenges in environments with varying lighting conditions or visual obstructions, which were not extensively tested in this study. Another limitation was the potential for over-fitting to specific task environments. The training setup in controlled ViZDoom scenarios might not have fully captured the complexities of real-world navigation tasks, where environmental dynamics were less predictable. Thus, while the A2C with GAE approach showed promise, its applicability to a broader range of visual navigation tasks would have benefited from additional research and testing in more varied and less controlled environments.


Table \ref{tab:A2C_Papers} summarizes the discussed papers in this section.
Over the next subsection, Deterministic Policy Gradient (DPG) algorithm, which addresses the challenges associated with continuous action spaces is discussed.

\begin{table}[t]
\centering
\renewcommand{\arraystretch}{1.2} 
\caption{A2C Papers Review }
\begin{tabular}{|>{\raggedright\arraybackslash}p{4cm}|>{\raggedright\arraybackslash}p{3cm}|}
\hline
\textbf{Application Domain} & \textbf{References} \\
\hline
Edge computing environments & \cite{yang2024ha}, \cite{sun2022a2c}
 \\
\hline 
Network Optimization & \cite{choppara2024reliability},\cite{kwon2022learning}
\\
\hline
Cloud-based Control and Encryption Systems &\cite{choppara2024reliability}
 \\
\hline
Energy and Power Management (IoT Networks, Smart Energy Systems) &\cite{dantas2023beam}
 \\
\hline
Financial Applications  & \cite{vishal2021trading}
\\
\hline
Autonomous UAVs & \cite{han2019multi} 

 \\
\hline
Visual Navigation & \cite{shao2018visual}  \\
\hline
\end{tabular}
\label{tab:A2C_Papers}
\end{table}

\subsection{Deterministic Policy Gradient (DPG)}

DPG addresses the challenges associated with continuous action spaces and offers significant improvements in sample efficiency over stochastic policy gradient methods. Traditional policy gradient methods in RL use stochastic policies, where the policy \(\pi_\theta(a|s)\) is a probability distribution over actions given a state, parameterized by \(\theta\). These methods rely on sampling actions from this distribution to compute the policy gradient, which can be computationally expensive and sample inefficient, especially in high-dimensional action spaces \cite{sutton2018reinforcement, silver2014deterministic}.
In contrast, the DPG algorithm uses a deterministic policy, denoted by \(\mu_\theta(s)\), which directly maps states to actions without involving any randomness. The policy gradient theorem for deterministic policies shows that the gradient of the expected return with respect to the policy parameters can be computed as \cite{silver2014deterministic}:

\begin{equation}
    \nabla_\theta J(\mu_\theta) = \mathbb{E}_{s \sim \rho^\mu} \left[ \nabla_\theta \mu_\theta(s) \nabla_a Q^\mu(s, a) \big|_{a = \mu_\theta(s)} \right]
\end{equation}
where \(Q^\mu(s, a)\) is the action-value function under the deterministic policy \(\mu_\theta\), and \(\rho^\mu\) is the discounted state visitation distribution under \(\mu_\theta\).

By employing an off-policy learning approach, DPG ensures adequate exploration while learning a deterministic target policy. To generate exploratory actions and gather experiences, a behavior policy, often a stochastic policy, is used. Gradients derived from these experiences are then used to update the deterministic policy. As the same experiences can be reused to improve policy in this non-policy setting, the data collected can be utilized more efficiently than in a policy setting \cite{degris2012off}.

The DPG algorithm is typically implemented within an Actor-Critic framework, where the actor represents the deterministic policy \(\mu_\theta\) and the critic estimates the action-value function \(Q^\mu(s, a)\). The critic is trained using TD learning to minimize the Bellman error:

\begin{equation}
    \delta_t = r_t + \gamma Q^\mu(s_{t+1}, \mu_\theta(s_{t+1})) - Q^\mu(s_t, a_t)
\end{equation}
where \(\delta_t\) is the TD error, \(r_t\) is the reward, \(\gamma\) is the discount factor, and \(s_t, a_t\) are the state and action at time step \(t\). The actor updates the policy parameters in the direction suggested by the critic:

\begin{equation}
    \theta \leftarrow \theta + \alpha \nabla_\theta \mu_\theta(s_t) \nabla_a Q^\mu(s_t, a_t) \big|_{a = \mu_\theta(s_t)}
\end{equation}
where \(\alpha\) is the learning rate for the actor.
A variant of DPG, which is designed to handle continuous action spaces with the help of DL, is analyzed in the next subsection in detail.

\subsubsection{Deep Deterministic Policy Gradient (DDPG)}
\label{Sec_DDPG}

The DDPG algorithm is an extension of the DPG method, designed to handle continuous action spaces effectively, introduced by \cite{lillicrap2015continuous}. DDPG leverages the power of DL to address the challenges associated with high-dimensional continuous control tasks \cite{lillicrap2015continuous}. The foundation of DDPG lies in the DPG algorithm. This approach contrasts with stochastic policy gradients, which sample actions from a probability distribution. The deterministic nature of DPG reduces the variance of gradient estimates and improves sample efficiency, making it suitable for continuous action spaces.
DDPG employs an Actor-Critic architecture, where the actor network represents the policy \(\mu(s|\theta^\mu)\) and the critic network estimates the action-value function \(Q(s, a|\theta^Q)\). The actor network outputs a specific action for a given state, while the critic network evaluates the action by estimating the expected return. The policy gradient is computed using the chain rule:

\begin{equation}
    \nabla_{\theta^\mu} J \approx \mathbb{E}_{s \sim \rho^\beta} \left[ \nabla_a Q(s, a|\theta^Q) \big|_{a = \mu(s|\theta^\mu)} \nabla_{\theta^\mu} \mu(s|\theta^\mu) \right]
\end{equation}
where \(\rho^\beta\) denotes the state distribution under a behavior policy \(\beta\).
To stabilize learning and address the challenges of training with large, non-linear function approximators, DDPG incorporates two key techniques from the DQN algorithm \cite{mnih2015human}:

\begin{enumerate}
    \item {Replay Buffer}: A replay buffer stores transitions \((s_t, a_t, r_t, s_{t+1})\) observed during training. By sampling mini-batches of transitions from this buffer, DDPG minimizes the correlations between consecutive samples, which stabilizes training and improves efficiency.
    \item {Target Networks}: DDPG uses target networks for both the actor and critic, which are periodically updated with a soft update mechanism. The target networks provide stable targets for the Q-learning updates, reducing the likelihood of divergence:
    \begin{equation}
        \theta^{Q'} \leftarrow \tau \theta^Q + (1 - \tau) \theta^{Q'} \\
    \end{equation}
    \begin{equation}
        \theta^{\mu'} \leftarrow \tau \theta^\mu + (1 - \tau) \theta^{\mu'}
    \end{equation}
where \(\tau < 1\) is the target update rate.
    
\end{enumerate}

Exploration in continuous action spaces is crucial for effective learning. DDPG employs an exploration policy by adding noise to the actor's deterministic policy. An Ornstein-Uhlenbeck process \cite{uhlenbeck1930theory} is typically used to generate temporally correlated noise, promoting exploration in environments with inertia.

The DDPG algorithm operates as follows:
As shown in Alg. \ref{alg:DDPG}, first, the parameters of the actor network \(\theta^\mu\) and the critic network \(\theta^Q\) are initialized \cite{kong2023edge} (line 1-2). Target networks for both the actor and critic are also initialized. Then, multiple agents interact with their respective environments, collecting transitions \((s_t, a_t, r_t, s_{t+1})\) which are stored in a replay buffer (lines 3-10). For each agent, the actor selects actions based on the current policy with added exploration noise. The critic network is updated using the Bellman equation to minimize the TD error (lines 11-13). The actor network is updated using the policy gradient derived from the critic (line 14). Periodically, the target networks are updated to slowly track the learned networks (line 15).
Over the next paragraphs, we will analyze some of the papers in the literature that used DDPG.

\begin{algorithm}[t]
\caption{DDPG}
\begin{algorithmic}[1]
\State Randomly initialize critic network \(Q(s, a|\theta^Q)\) and actor \(\mu(s|\theta^\mu)\) with weights \(\theta^Q\) and \(\theta^\mu\).
\State Initialize target network \(Q'\) and \(\mu'\) with weights \(\theta^{Q'} \gets \theta^Q\), \(\theta^{\mu'} \gets \theta^\mu\).
\State Initialize replay buffer \(R\).
\For{episode \(= 1\) to \(M\)}
    \State \parbox[t]{\dimexpr\linewidth-\algorithmicindent}{ Initialize a random process \(N\) for action exploration.\strut}
    \State Receive initial observation state \(s_1\).
    \For{t \(= 1\) to \(T\)}
        \State \parbox[t]{\dimexpr\linewidth-\algorithmicindent}{ Select action \(a_t = \mu(s_t|\theta^\mu) + N_t\) according \\
        to the current policy and exploration noise\strut}
        \State \parbox[t]{\dimexpr\linewidth-\algorithmicindent}{ Execute action \(a_t\) and observe reward \(r_t\) \\
        and observe new state \(s_{t+1}\)\strut}
        \State Store transition \((s_t, a_t, r_t, s_{t+1})\) in \(R\)
        \State \parbox[t]{\dimexpr\linewidth-\algorithmicindent}{ Sample a random minibatch of \(N\) \\
        transitions \((s_i, a_i, r_i, s_{i+1})\) from \(R\)\strut}
        \State Set \(y_i = r_i + \gamma Q'(s_{i+1}, \mu'(s_{i+1}|\theta^{\mu'})|\theta^{Q'})\).
        \State \parbox[t]{\dimexpr\linewidth-\algorithmicindent}{ Update critic by minimizing the loss: \\
        \(L = \frac{1}{N} \sum_i (y_i - Q(s_i, a_i|\theta^Q))^2\).\strut}
        \State \parbox[t]{\dimexpr\linewidth-\algorithmicindent}{ Update the actor \\
        policy using the sampled policy gradient:
        \begin{align}
            \nonumber \nabla_{\theta^\mu} J &\approx \frac{1}{N} \sum_i \nabla_a Q(s, a|\theta^Q)|_{s=s_i, a=\mu(s_i)} \\
            \nonumber &\times \nabla_{\theta^\mu} \mu(s|\theta^\mu)|_{s_i}
        \end{align}\strut}

        \State \parbox[t]{\dimexpr\linewidth-\algorithmicindent}{ Update the target networks:
        \[
        \theta^{Q'} \gets \tau \theta^Q + (1 - \tau) \theta^{Q'}
        \]
        \[
        \theta^{\mu'} \gets \tau \theta^\mu + (1 - \tau) \theta^{\mu'}
        \]\strut}
    \EndFor
\EndFor
\end{algorithmic}
\label{alg:DDPG}
\end{algorithm}

Authors in \cite{candeli2022deep} presented an innovative method for developing a missile lateral acceleration control system using the DDPG algorithm. This study reframed the autopilot control problem within the RL context, utilizing a 2-degrees-of-freedom nonlinear model of the missile’s longitudinal dynamics for training. One strength was the incorporation of performance metrics such as settling time, undershoot, and steady-state error into the reward function. By integrating these key performance indicators, the authors ensured that the trained agent not only learned to control the missile effectively but also adhered to desirable performance standards. This approach enhanced the practical applicability of the method, ensuring that the control system met operational requirements. The method's scalability to more complex scenarios and larger-scale implementations was another concern. The increased number of states and potential interactions in a real-world missile control system could have introduced additional complexities, making it challenging to maintain the same level of performance. Further research was needed to explore the scalability of the approach and develop mechanisms to manage the increased computational load.

A novel approach to fast charging lithium-ion batteries using a combination of a Model-based state observer and a DRL optimizer, specifically the DDPG algorithm, was proposed in \cite{wei2021deep}. This method aimed to balance charging rapidity with the enforcement of thermal safety and degradation constraints. One of the notable strengths of this paper was its innovative application of the DDPG algorithm to the complex problem of fast charging lithium-ion batteries. By formulating a multi-objective optimization problem that included penalties for over-temperature and degradation, the authors effectively addressed the crucial aspects of battery safety and longevity. This approach ensured that the charging strategy not only accelerated the charging process but also maintained the battery within safe operating limits, thus extending its life. The method's scalability to different types of batteries and charging environments was a concern. While the results were promising for the specific battery model used in the study, the ability to generalize the approach to other battery types and configurations remained uncertain. The increased number of variables and potential interactions in more complex systems could have introduced additional complexities, making it challenging to maintain the same level of performance. Further research was needed to explore the scalability of the approach and develop mechanisms to manage the increased computational load.

Authors explored the application of DDPG to the problem of obstacle avoidance in the trajectory planning of a robot arm in \cite{wen2018path}. The authors proposed using DDPG to plan the trajectory of a robot arm, ensuring smooth and continuous motion while avoiding obstacles. The rewards were specifically designed to overcome the convergence difficulties posed by multiple and potentially antagonistic rewards. One strength of this paper was the careful design of the reward function. By considering both the distance to the target and the proximity to obstacles, the authors ensured that the robot not only reached its goal efficiently but also avoided collisions. This multi-faceted reward structure addressed the convergence issues often encountered in RL tasks with multiple objectives. The empirical validation through simulations in the MuJoCo environment demonstrated the effectiveness of the proposed method, showing that the robot arm could successfully navigate to its target while avoiding obstacles. The method's scalability to more complex scenarios and larger-scale implementations might have impeded the achieved performance. The increased number of states and potential interactions in a real-world robotic system could have introduced additional complexities, making it challenging to maintain the same level of performance. Further research was needed to explore the scalability of the approach and develop mechanisms to manage the increased computational load. Investigating the impact of various hyperparameters and network architectures on the performance of the DDPG algorithm could have provided deeper insights into optimizing the method for different robotic applications.

An enhanced version of DDPG, integrating a Long Short-Term Memory (LSTM) network-based encoder to handle dynamic obstacle avoidance for mobile robots in stochastic environments is introduced in \cite{gao2023improved}. One of the notable strengths of this paper was the innovative combination of DDPG with LSTM. This hybrid approach allowed the robot to encode a variable number of obstacles into a fixed-length representation, which addressed the limitation of traditional DDPG that required a fixed number of inputs. By utilizing the LSTM network-based encoder, the proposed method could effectively process and integrate dynamic environmental information, enhancing the robot's adaptability to unpredictable scenarios. The reliance on accurate environmental sensing and real-time data processing was a potential limitation. The performance of the LSTM-based encoder and the overall DDPG framework heavily depended on the quality and accuracy of the sensor data. In real-world applications, factors such as sensor noise, varying environmental conditions, and communication delays could have affected the reliability and robustness of the system. Ensuring robust performance under diverse and unpredictable conditions remained a critical challenge.

DDPG combined with prioritized sampling to optimize power control in wireless communication systems, specifically targeting Multiple Sweep Interference (MSI) scenarios, was designed in \cite{zhou2020deep}. Prioritized sampling was another innovative aspect of the paper. By focusing on more valuable experiences during training, the algorithm accelerated the learning process and improved convergence speed. The empirical results showed that the DDPG scheme with prioritized sampling (DDPG-PS) outperformed the traditional DDPG scheme with uniform sampling and DQN scheme. This was evident in various MSI scenarios, where the DDPG-PS scheme achieved better reward performance and stability. The scalability of the proposed method to more complex scenarios and larger-scale implementations was another concern. While the results were promising in simulated environments, the ability to handle a broader range of interference patterns and larger numbers of channels remained uncertain. The increased number of states and potential interactions could have introduced additional complexities, making it challenging to maintain the same level of performance. Further research was needed to explore the scalability of the approach and develop mechanisms to manage the increased computational load.

Authors in \cite{liang2020agent} employed DDPG to model the bidding strategies of generation companies in electricity markets. This approach was aimed at overcoming the limitations of traditional game-theoretic methods and conventional RL algorithms, particularly in environments characterized by incomplete information and high-dimensional continuous state/action spaces. One significant strength was the ability of the proposed method to converge to the Nash Equilibrium even in an incomplete information environment. Traditional game-theoretic methods often required complete information and were limited to static games. In contrast, the DDPG-based approach could dynamically simulate repeated games and achieve stable convergence, demonstrating its robustness in modeling real-world market conditions. One limitation was the reliance on accurate modeling of market conditions and real-time data processing. The effectiveness of the proposed method depended heavily on the precision of the input data, such as nodal prices and load demands. In real-world applications, factors such as data inaccuracies, communication delays, and varying environmental conditions could have impacted the reliability and robustness of the system. Ensuring robust performance under diverse and unpredictable conditions remained a critical challenge that needed to be addressed. Additionally, the study assumed a specific structure for the neural networks used in the actor and critic models. The performance of the algorithm could have been sensitive to the choice of network architecture and hyperparameters. A more systematic exploration of different architectures and their impact on performance could have provided deeper insights into optimizing the DDPG algorithm for electricity market modeling.

An advanced approach for resource allocation in vehicular communications using a multi-agent DDPG algorithm was studied in \cite{xu2020deep}. This method was designed to handle the dynamic and high-mobility nature of vehicular environments, specifically targeting the optimization of the sum rate of Vehicle-to-Infrastructure (V2I) communications while ensuring the latency and reliability of Vehicle-to-Vehicle (V2V) communications. One of the significant strengths was the formulation of the resource allocation problem as a decentralized Discrete-time and Finite-state MDP. This approach allowed each V2V communication to act as an independent agent, making decisions based on local observations without requiring global network information. This decentralization was crucial for scalability and real-time adaptability in high-mobility vehicular environments. One potential limitation was the reliance on accurate and timely acquisition of CSI. In high-mobility vehicular environments, obtaining precise CSI could have been challenging due to fast-varying channel conditions. Any inaccuracies in CSI could have impacted the performance and robustness of the proposed resource allocation scheme. Ensuring robust performance under diverse and unpredictable conditions remained a critical challenge.
The last algorithm in the category of Actor-Critic methods to analyze is TD3, which is an enhancement of the DDPG algorithm, designed to address the issues of overestimation bias. This algorithm is analyzed in detail in the next subsection.


Table \ref{tab:DDPG_Papers} provides a summary of the analyzed papers with respect to their domain.

\begin{table}[t]
\centering
\renewcommand{\arraystretch}{1.2} 
\caption{DDPG Papers Review}
\begin{tabular}{|>{\raggedright\arraybackslash}p{4cm}|>{\raggedright\arraybackslash}p{3cm}|}
\hline
\textbf{Application Domain} & \textbf{References} \\
\hline
Theoretical Research (Convergence, stability) & \cite{lillicrap2015continuous}
 \\
\hline
Missile Control Systems & \cite{candeli2022deep}

\\
\hline
Battery Charging Optimization & \cite{wei2021deep} 

 \\
\hline
Robotics & \cite{wen2018path}, \cite{gao2023improved} 

 \\
\hline
Network Optimization & \cite{zhou2020deep}
\\
\hline
Financial Applications & \cite{liang2020agent}
 \\
\hline
Multi-agent Systems and Autonomous Behaviors & \cite{xu2020deep}
 \\
\hline
\end{tabular}
\label{tab:DDPG_Papers}
\end{table}

\subsubsection{Twin Delayed Deep Deterministic Policy Gradient (TD3)}

TD3 is an enhancement of the DDPG algorithm, designed to address the issues of overestimation bias in function approximation within Actor-Critic methods. introduced by \cite{fujimoto2018addressing}, TD3 incorporates several innovative techniques to improve the stability and performance of continuous control tasks in RL.

Overestimation bias occurs when the value estimates for certain actions are consistently higher than their true values due to function approximation errors. This issue is well-documented in Value-based Methods like Q-learning \cite{thrun1992efficient}. In Actor-Critic methods, this bias can lead to suboptimal policy updates and divergent behavior \cite{fujimoto2018addressing}.

TD3 builds on the Double Q-learning concept, which mitigates overestimation bias by maintaining two separate value estimators and using the minimum of the two estimates for the target update \cite{van2009theoretical}. This approach is adapted to the Actor-Critic setting by employing two critic networks, \(Q_{\theta_1}\) and \(Q_{\theta_2}\), which are independently trained. The target value is computed as:

\begin{equation}
    y = r + \gamma \min_{i=1,2} Q_{\theta_i'}(s', \mu_{\theta'}(s'))
\end{equation}

where \(Q_{\theta_i'}\) are the target critic networks and \(\mu_{\theta'}\) is the target actor network.

To further reduce the error propagation from the critic to the actor, TD3 delays the policy updates relative to the value updates. Specifically, the policy (actor) network is updated less frequently than the critic networks. This strategy ensures that the value estimates used to update the policy are more accurate and stable. The policy is updated every \(d\) iterations, where \(d\) is a hyperparameter typically set to 2 or more \cite{fujimoto2018addressing}.
To prevent over-fitting to narrow peaks in the value estimate, TD3 introduces target policy smoothing. This technique adds noise to the target policy, encouraging smoother value estimates and more robust policy learning. The target value is computed with added noise:

\begin{equation}
    y = r + \gamma \min_{i=1,2} Q_{\theta_i'}(s', \mu_{\theta'}(s') + \epsilon)
\end{equation}
where \(\epsilon\) is clipped noise sampled from a Gaussian distribution.
Based on Alg. \ref{alg:TD3}, the TD3 algorithm operates as follows:

First, the parameters of the actor network \(\theta\) and the critic networks \(\theta_1\) and \(\theta_2\) are initialized (line 1). Target networks for both the actor and critics are also initialized (line 2). Multiple agents interact with their respective environments, collecting transitions \((s_t, a_t, r_t, s_{t+1})\) which are stored in a replay buffer (line 6). The actor selects actions with added exploration noise (line 8). The critic networks are updated by minimizing the TD error using the clipped double Q-learning target (lines 9-10). The actor network is updated using the deterministic policy gradient, but only every \(d\) iterations (line 12). Periodically, the target networks are updated to slowly track the learned networks (line 13).

\begin{algorithm}[t]
\caption{TD3}
\begin{algorithmic}[1]
\State Initialize critic networks \(Q_{\theta_1}\), \(Q_{\theta_2}\), and actor network \(\pi_{\phi}\) with random parameters \(\theta_1\), \(\theta_2\), \(\phi\)
\State Initialize target networks \(\theta_1' \gets \theta_1\), \(\theta_2' \gets \theta_2\), \(\phi' \gets \phi\)
\State Initialize replay buffer \(\mathcal{B}\)
\For{\(t = 1\) to \(T\)}
    \State \parbox[t]{\dimexpr\linewidth-\algorithmicindent}{ Select action with exploration noise \(a \sim \pi_{\phi}(s) + \epsilon\), \(\epsilon \sim \mathcal{N}(0, \sigma)\) and observe reward \(r\) and new state \(s'\)\strut}
    \State Store transition tuple \((s, a, r, s')\) in \(\mathcal{B}\)
    \State \parbox[t]{\dimexpr\linewidth-\algorithmicindent}{ Sample mini-batch of \(N\) transitions \((s, a, r, s')\) from \(\mathcal{B}\)\strut}
    \State \(\tilde{a} \gets \pi_{\phi'}(s') + \epsilon\), \(\epsilon \sim \text{clip}(\mathcal{N}(0, \sigma), -c, c)\)
    \State \(y \gets r + \gamma \min_{i=1,2} Q_{\theta_i'}(s', \tilde{a})\)
    \State \parbox[t]{\dimexpr\linewidth-\algorithmicindent}{ Update critics \(Q_{\theta_i} \gets \arg\min_{\theta_i} N^{-1} \sum(y - Q_{\theta_i}(s, a))^2\)\strut}
    \If{\(t \mod d = 0\)}
        \State \parbox[t]{\dimexpr\linewidth-\algorithmicindent}{ Update \(\phi\) by deterministic policy gradient
        \resizebox{0.9\linewidth}{!}{$
        \nabla_{\phi} J(\phi) = N^{-1} \sum \nabla_a Q_{\theta_1}(s, a)|_{a=\pi_{\phi}(s)} \nabla_{\phi} \pi_{\phi}(s)
        $}\strut}
        \State Update target networks:
        \[
        \theta_i' \gets \tau \theta_i + (1 - \tau) \theta_i'
        \]
        \[
        \phi' \gets \tau \phi + (1 - \tau) \phi'
        \]
    \EndIf
\EndFor
\end{algorithmic}
\label{alg:TD3}
\end{algorithm}

Authors in \cite{yazar2023actor}, as the first analyzed paper in the literature, showcased both strengths and limitations in applying the TD3 algorithm for energy management in Hybrid EVs (HEVs). One of the primary strengths of the paper was its innovative application of the TD3 algorithm, which enhanced training efficiency and stability over the previously used methods like Q-learning, DQN, and DDPG. The TD3 algorithm's use of two critic networks helped provide more stable training by mitigating the overestimation bias. The paper also demonstrated significant improvements in fuel economy and battery state-of-charge sustainability, which are crucial metrics for HEVs. By comparing the performance under various driving cycles, the authors provided comprehensive evidence of the TD3 algorithm's effectiveness in real-world scenarios. However, there were notable limitations. The implementation of TD3, while improving stability, introduced complexity in the training process, requiring careful tuning of hyperparameters to achieve optimal performance. The algorithm's reliance on extensive computational resources for training might have limited its practical applicability in scenarios with constrained resources. Additionally, the paper focused primarily on the simulation results without providing sufficient real-world testing to validate the algorithm's performance under actual driving conditions. This gap raised questions about the robustness of the proposed method when deployed in a real-world environment.

The application of the TD3 algorithm for the target tracking of UAVs was proposed in \cite{mosali2022twin}. The authors integrated several enhancements into the TD3 framework to improve its performance in handling the high nonlinearity and dynamics of UAV control. A significant strength was the novel reward formulation that incorporated exponential functions to limit the effects of velocity and acceleration on the policy function approximation. This approach prevented deformation in the policy function, leading to more stable and robust learning outcomes. Additionally, the concept of multistage training, where the training process was divided into stages focusing on position, velocity, and acceleration sequentially, enhanced the learning efficiency and performance of the UAV in tracking tasks. However, the proposed method also had several limitations. The integration of a PD controller and the novel reward formulation added to the complexity of the training process. The scalability of the proposed method to more complex environments with a higher number of dynamic obstacles or more sophisticated UAV maneuvers was another concern. While the results were promising in the tested scenarios, the ability to handle a broader range of operational conditions and larger numbers of UAVs remained uncertain. The increased number of states and potential interactions could have introduced additional complexities, making it challenging to maintain the same level of performance.

A novel dynamic MsgA channel allocation strategy using TD3 to mitigate the issue of MsgA channel collisions in Low Earth Orbit (LEO) satellite communication systems is investigated in \cite{han2023two}. The paper's approach to dynamically pre-configuring the mapping relationship between PRACH occasions and PUSCH occasions based on historical access information was another strong point. This method allowed the system to adapt to changing access demands effectively, ensuring efficient use of available resources and reducing collision rates. The empirical results were impressive, demonstrating a 39.12\% increase in access success probability, which validated the effectiveness of the proposed strategy. The scalability of the proposed method to larger and more complex satellite networks was one of the concerns. While the results were promising in the tested scenarios, the ability to handle a broader range of interference patterns and a larger number of users remained uncertain. The increased number of states and potential interactions could have introduced additional complexities, making it challenging to maintain the same level of performance. Further research was needed to explore the scalability of the approach and develop mechanisms to manage the increased computational load.

A TD3-based method for optimizing Voltage and Reactive (VAR) power in distribution networks with high penetration of Distributed Energy Resources (DERs) such as battery energy storage and solar photovoltaic units was investigated in \cite{hossain2022volt}. The authors' approach of coordinating the reactive power outputs of fast-responding smart inverters and the active power of battery ESS enhanced the overall efficiency of the network. By carefully designing the reward function to ensure a proper voltage profile and effective scheduling of reactive power outputs, the method optimized both voltage regulation and power loss minimization. The results demonstrated that the TD3-based method outperformed traditional methods such as local droop control and DDPG-based approaches, showing significant improvements in reducing voltage fluctuations and minimizing power loss in the IEEE 34- and 123-bus test systems. The scalability of the proposed method to larger and more complex distribution networks was another concern. While the results were promising in the tested IEEE 34- and 123-bus systems, the ability to handle a broader range of network configurations and a larger number of DERs remained uncertain. The increased number of states and potential interactions could have introduced additional complexities, making it challenging to maintain the same level of performance. Further research was needed to explore the scalability of the approach and develop mechanisms to manage the increased computational load.

Authors in \cite{shehab2021low} presented an innovative approach to quadrotor control, leveraging the TD3 algorithm to address stabilization and position tracking tasks. This study was notable for its application to handle the complex, non-linear dynamics of quadrotor systems. The authors' method of integrating target policy smoothing, twin critic networks, and delayed updates of value networks enhanced the learning efficiency and reduced variance in the policy updates. This comprehensive approach ensured that the quadrotor could achieve precise control in both stabilization and position tracking tasks. The empirical results demonstrated the effectiveness of the TD3-based controllers, showcasing significant improvements in achieving and maintaining target positions under various initial conditions. The scalability of the proposed method to more complex environments with dynamic obstacles and more sophisticated maneuvers was another concern. While the results were promising in the tested scenarios, the ability to handle a broader range of operational conditions and larger-scale implementations remained uncertain. The increased number of states and potential interactions could have introduced additional complexities, making it challenging to maintain the same level of performance. Further research was needed to explore the scalability of the approach and develop mechanisms to manage the increased computational load.

A real-time charging navigation method for multi-autonomous underwater vehicles (AUVs) systems using the TD3 algorithm was designed in \cite{yu2023multi}. This method was designed to improve the efficiency of navigating AUVs to their respective charging stations by training a trajectory planning model in advance, eliminating the need for recalculating navigation paths for different initial positions and avoiding dependence on sensor feedback or pre-arranged landmarks. The primary strength of this paper lies in its application of the TD3 algorithm to the multi-AUV charging navigation problem. By training the trajectory planning model in advance, the method significantly improved the real-time performance of multi-AUV navigation. However, the paper also had some limitations. One major limitation was the reliance on the accuracy of the AUV motion model and the assumptions made during its formulation. For instance, the model assumed constant velocity and neglected factors like water resistance and system delays, which could have affected the real-world applicability of the results. Moreover, the simulation environment used for training and testing might not have fully captured the complexities and variabilities of real underwater environments. Another potential limitation was the need for extensive computational resources for training the TD3 model, especially given the high number of training rounds (up to 6000) and the large experience replay buffer size.

Authors in \cite{bishen2023adaptive} presented an advanced approach for Adaptive Cruise Control (ACC) using the TD3 algorithm. This method addressed the complexities of real-time decision-making and control in automotive applications. The authors carefully designed the reward function to consider the velocity error, control input, and additional terms to ensure stability and smooth driving behavior. This reward structure allowed the algorithm to learn an optimal policy that maintained safe distances between vehicles while adapting to changing traffic conditions. The empirical results demonstrated the effectiveness of the TD3-based ACC system in both normal and disturbance scenarios, highlighting its robustness and adaptability. The scalability of the proposed method to more complex driving scenarios and larger-scale implementations was another concern. While the results were promising in the tested scenarios, the ability to handle a broader range of operational conditions and larger numbers of vehicles remained uncertain. The increased number of states and potential interactions could have introduced additional complexities, making it challenging to maintain the same level of performance. Further research was needed to explore the scalability of the approach and develop mechanisms to manage the increased computational load. Additionally, the study assumed a specific structure for the neural networks used in the actor and critic models. The performance of the algorithm could have been sensitive to the choice of network architecture and hyperparameters. A more systematic exploration of different architectures and their impact on performance could have provided deeper insights into optimizing the TD3 algorithm for adaptive cruise control.
Table \ref{tab:TD3_Papers} gives a detailed summary of the discussed papers, and the domains of each paper.

\begin{table}[t]
\centering
\renewcommand{\arraystretch}{1.2} 
\caption{TD3 Papers Review }
\begin{tabular}{|>{\raggedright\arraybackslash}p{4cm}|>{\raggedright\arraybackslash}p{3cm}|}
\hline
\textbf{Application Domain} & \textbf{References} \\
\hline
Energy and Power Management & \cite{yazar2023actor}
 \\
\hline
Multi-agent Systems and Autonomous UAVs & \cite{mosali2022twin}, \cite{yu2023multi}
\\
\hline
Network Resilience and Optimization & \cite{han2023two}
 \\
\hline
Energy and Power Management & \cite{hossain2022volt}
 \\
\hline
Real-time Systems and Hardware Implementations & \cite{shehab2021low}, \cite{bishen2023adaptive} \\
\hline
\end{tabular}
\label{tab:TD3_Papers}
\end{table}


\begin{table*}[t]
\centering
\renewcommand{\arraystretch}{1.2} 
\caption{Overview of Algorithms Across Different Application Domains}
\begin{adjustbox}{max width=\textwidth} 
\begin{tabular}{|p{4cm}| p{9cm}| p{5cm}|}
\hline
\textbf{Application Domain} & \textbf{Algorithm(s)} & \textbf{References} \\
\hline
Energy Efficiency and Power Management & MC, Q Learning, Double Q-learning, SARSA, Dyna-Q, AMF, DDQN, TRPO, Dueling DQN, PPO, A2C, TD3 & \cite{liu2018intelligent}, \cite{paterova2022robustness}, \cite{paterova2021data}, \cite{fan2023double}, \cite{jaiswal2020green}, \cite{wang2020peer}, \cite{muduli2023application}, \cite{xu2015dyna}, \cite{del2023new}, \cite{iqbal2021double}, \cite{li2019partially}, \cite{xuan2021power}, \cite{liu2019green}, \cite{guo2023function}, \cite{thattai2023consumer}, \cite{zhang2020real}, \cite{dantas2023beam}, \cite{yazar2023actor}, \cite{hossain2022volt} \\
\hline
Cloud-based Systems & SARSA, A3C, A2C & \cite{suh2021sarsa}, \cite{mangalampalli2024multi}, \cite{du2021resource}, \cite{yang2024ha}, \cite{sun2022a2c}, \cite{choppara2024reliability} \\
\hline
Optimization & TD-Learning DQN, DDQN, Dueling DQN, DDPG & \cite{de2005adaptive}, \cite{talaat2022effective}, \cite{li2023double}, \cite{han2020research}, \cite{wei2021deep} \\
\hline
Multi-agent Systems & SARSA, MCTS, Prioritized Sweeping, Dyna-Q, DDQN, A3C, TD3 & \cite{go2016reinforcement}, \cite{zerbel2019multiagent}, \cite{bargiacchi2020model}, \cite{hwang2014model}, \cite{huang2022path}, \cite{xiaofei2022global}, \cite{yu2023multi} \\
\hline
Algorithmic RL & TD-Learning, MC, SARSA, Prioritized Sweeping & \cite{subramanian2019renewal}, \cite{peters2005monte}, \cite{wang2012monte}, \cite{wu2017monte}, \cite{bulteau2002new}, \cite{moradimaryamnegari2022model}, \cite{kuchibhotla2020n}, \cite{su2021prioritized}, \cite{baier2013monte}, \cite{lanctot2014monte}, \cite{li2020morphing} \\
\hline
General RL & TD-Learning, Q-learning, DQN, Dueling DQN, TRPO, PPO & \cite{altahhan2018td}, \cite{sutton1988learning}, \cite{bi2023comparative}, \cite{jin2018q}, \cite{stember2022reinforcement}, \cite{liu2020application}, \cite{mondal2023xss} \\
\hline
Robotics & MC, Q-learning, Dyna-Q, DDQN, Dueling DQN, REINFORCE, TRPO, DDPG & \cite{rana2023residual}, \cite{kober2013reinforcement}, \cite{balasubramanian2023intrinsically}, \cite{abeyruwan2023sim2real}, \cite{aghaei2022real}, \cite{wang2021taxi}, \cite{qu2011visual}, \cite{vitolo2018performance}, \cite{zhang2019double}, \cite{yu2023obstacle}, \cite{xue2019deep}, \cite{weerakoon2022htron}, \cite{erens2024universal}, \cite{wen2018path}, \cite{gao2023improved} \\
\hline
Financial Applications & Q-learning, A2C, DDPG & \cite{darapaneni2020automated}, \cite{vishal2021trading}, \cite{liang2020agent} \\
\hline
Games & TD-Learning, SARSA, MCTS, Dyna-Q, DDQN, DQN, A3C & \cite{souchleris2023reinforcement}, \cite{koyamada2024pgx},  \cite{xu2023language}, \cite{qu2023pursuit}, \cite{baykal2019reinforcement}, \cite{baxter1999knightcap}, \cite{daswani2013q}, \cite{wender2012applying}, \cite{coulom2006efficient}, \cite{wang2012multi}, \cite{perez2014multiobjective}, \cite{de2016monte}, \cite{winands2010monte}, \cite{santos2017monte}, \cite{cowling2012information}, \cite{ciancarini2010monte}, \cite{moerland2018monte}, \cite{santos2012dyna}, \cite{pei2021improved}, \cite{yoon2017deep}, \cite{lv2019path}, \cite{gao2019anti}, \cite{joypriyanka2023chess} \\
\hline
Signal Processing & TD-Learning & \cite{saha2002maximum} \\
\hline
Networks & TD-Learning, Q-learning, SARSA, Dyna-Q, Prioritized Sweeping, DQN, Dueling DQN, REINFORCE, A3C, A2C, DDPG, TD3 & \cite{lassila2001efficient}, \cite{xiao2022random}, \cite{nassar2019reinforcement}, \cite{gonzalez2023comparison}, \cite{oh2020reinforcement}, \cite{zhang2022anti}, \cite{dearden2001structured}, \cite{guo2022deep}, \cite{song2018power}, \cite{ban2020autonomous}, \cite{tadele2024optimization}, \cite{tao2024reinforcement}, \cite{shi2019content}, \cite{choppara2024reliability}, \cite{kwon2022learning}, \cite{zhou2020deep}, \cite{han2023two} \\
\hline
Intelligent Transportation Systems (ITS) & Q-learning, SARSA, Prioritized Sweeping, Dyna-Q, DDQN, Dueling DQN, TRPO, PPO, A2C, DDPG & \cite{zhu2023multi}, \cite{yazdani2023intelligent}, \cite{liu2023swarm}, \cite{chen2023deep}, \cite{ghosh2024maximizing}, \cite{shou2023design}, \cite{wang2023q}, \cite{aljohani2022real}, \cite{kekuda2021reinforcement}, \cite{desai2017prioritized}, \cite{zhang2018human}, \cite{zhang2021tactical}, \cite{mo2019decision}, \cite{huang2022usv}, \cite{wang2020autonomous}, \cite{liu2022improved}, \cite{jiang2021research}, \cite{kuba2021trust}, \cite{santoso2020multiagent}, \cite{wei2019mixed}, \cite{ying2021adaptive}, \cite{guan2020centralized}, \cite{tiong2022autonomous}, \cite{han2019multi}, \cite{shao2018visual}, \cite{candeli2022deep}, \cite{xu2020deep}, \cite{mosali2022twin} \\
\hline
Theoretical Research & Q-learning, TD-Learning, Prioritized Sweeping, REINFORCE, TRPO, DDPG & \cite{dayan1992convergence}, \cite{dayan1994td}, \cite{chen2021lyapunov}, \cite{lee2024analysis}, \cite{de2018unified}, \cite{wang2020greedy}, \cite{watkins1989learning}, \cite{peng1993efficient}, \cite{van2013planning}, \cite{zhang2021sample}, \cite{yuan2019monotonic}, \cite{lillicrap2015continuous} \\
\hline
Dynamic Environments & TD-Learning, Q-learning, Dyna-Q & \cite{de2018per}, \cite{wiering2007two}, \cite{chai2021multi}, \cite{li2023motion}, \cite{pei2021improved} \\
\hline
Partially Observable Environments & TD-Learning, SARSA & \cite{sutton2004temporal}, \cite{zuters2010realizing}, \cite{iima2008swarm} \\
\hline
Real-time Systems and Hardware Implementations & Q-learning, DQN, DDQN, PPO, TD3 & \cite{spano2019efficient}, \cite{nasreen2022overview}, \cite{da2018parallel}, \cite{ben2023efficient}, \cite{huang2017double}, \cite{lee2022intelligent}, \cite{jin2021optimal}, \cite{bohn2019deep}, \cite{lopes2018intelligent}, \cite{shehab2021low}, \cite{bishen2023adaptive} \\
\hline
Benchmark Tasks & TD-Learning & \cite{watkins1992q}, \cite{wu2023bias} \\
\hline
Data Management and Processing & Q-learning, DQN, PPO & \cite{akshay2023enhancing}, \cite{yao2022improving}, \cite{konecny2022double}, \cite{zhang2021image} \\
\hline
\end{tabular}
\end{adjustbox}
\label{tab:application_domains}
\end{table*}

\section{Discussion}\label{sec:Final_Notes}

Throughout this survey, we examined various algorithms in RL and their applications in a variety of domains, including but not limited to Robotics, ITS, Games, Wireless Networks, and many more. There is, however, more to discover both in terms of the number of analyzed papers and in terms of the different algorithms. There are several algorithms and methods that were not analyzed in this survey for a variety of reasons. To begin with, the considered algorithms are those that have been applied to a variety of domains and are more widely used by researchers. In addition, time, resources, and page limitations render it impossible to analyze all the algorithms and methods in one paper. Thirdly, understanding these algorithms enables one to understand different variations being introduced by the community on a regular basis. The purpose of this survey is not to identify which algorithm is better than the others, and as we know, there is no one-fit-all solution to RL, so one cannot state "for problem X, algorithm Y performs better than other algorithms" as it needs implementation of new algorithms for the same problem, and reproducing of the original work. Also, results achieved with RL and specifically DRL may vary since different extrinsic and intrinsic factors change, as stated in \cite{henderson2018deep}, making it tough to compare and analyze.


Lastly, we strongly recommend reading the chapters listed in this paper one by one, reading the introductions to the algorithms, and if necessary, consolidating the knowledge of each algorithm by reviewing the reference papers. As a result, you will be able to read the analysis of the papers that have used that particular algorithm. The provided tables are valuable to readers who are not interested in reading the entire article. By providing various tables at the end of the survey, we summarized helpful information gathered throughout the survey.

We tried our best to shed light on RL, in terms of theory and applications, to give a thorough understanding of various broad categories of algorithms. This survey is a helpful resource to readers who would like to expand their knowledge in RL (theory), as well as readers who desire to take a look at the applications of these algorithms in the literature.


In the final part of our survey, we present a comprehensive table that highlights the application of RL algorithms across various domains in Table \ref{tab:application_domains}. Given the wide-ranging impact of RL in numerous fields, we have categorized these domains into broader categories to provide a more organized and concise overview. This categorization allows us to succinctly illustrate the relevance and utilization of specific RL algorithms in different research areas while effectively managing the limited space available.

The \textbf{Energy Efficiency and Power Management} category encompasses research areas such as train control, IoTs, WBAN, PID Controllers, and Smart Energy Systems, all of which focus on optimizing energy usage and improving power management. \textbf{Cloud-based Systems} includes works focused on cloud-based control and encryption systems, as well as edge computing environments, reflecting the growing importance of RL in managing and optimizing cloud resources.

The \textbf{Optimization} category captures studies that leverage RL for solving complex optimization problems across various applications. \textbf{Multi-agent Systems} is a category that emphasizes RL's role in enabling Autonomous behaviors, covering research involving Shepherding, Virtual Agents, and other Multi-agent Systems.

\textbf{Algorithmic RL} covers advanced RL methodologies and hybrid approaches, including Renewal Theory, Rough Set Theory, Bayesian RL, and MPC tuning. The General RL category encompasses broad RL applications, including policy learning, cybersecurity, and learning from raw experience. \textbf{Robotics} research, focusing on the application of RL in robotics, includes trajectory control, learning, routing, and more.

In the \textbf{Financial Applications} category, studies on portfolio re-balancing and other financial strategies using RL are included. The \textbf{Games} category features research on game strategies in Chess, StarCraft, Video Games, and Card Games, illustrating RL's success in complex strategic environments. Signal Processing research, which uses RL for signal processing and parameter estimation, is grouped under its own category.

The \textbf{Networks} category covers studies focused on network reliability, blocking probabilities, Optical Transport Networks, Fog RAN, and Network Resilience and Optimization. \textbf{ITS} includes RL applications in Railway Systems, EVs, Intelligent Traffic Signal Control, UAVs, and other transportation-related technologies.

The \textbf{Theoretical Research} category includes studies focused on the theoretical aspects of RL, such as convergence and stability. \textbf{Dynamic Environments} research involves RL in environments like mazes, the Mountain Car problem, and Atari games. \textbf{Partially Observable Environments} includes studies on predictions, POMDPs, and Swarm Intelligence in optimization problems.

Research on applying RL in FPGA, Real-time Systems, and other hardware implementations is grouped under \textbf{Real-time Systems and Hardware Implementations}. Studies using benchmark tasks like Mountain Car, and Acrobot to test RL algorithms are included in the \textbf{Benchmark Tasks} category. \textbf{Data Management and Processing} involves research applying RL in data management and processing environments, such as Hadoop and Pathological Image Analysis. Finally, the Object Recognition category encompasses studies focusing on using RL for object recognition tasks.

Table \ref{tab:application_domains} serves as a quick reference for researchers to identify relevant work in their specific area of interest, showcasing the diversity and adaptability of RL algorithms across various domains. The categorization helps streamline the information, making it easier to navigate and understand the various applications of RL.

\section{Conclusion} \label{sec:conclusion}

In this survey, we presented a comprehensive analysis of Reinforcement Learning (RL) algorithms, categorizing them into Value-based, Policy-based, and Actor-Critical Methods. By reviewing numerous research papers, it highlights the strengths, weaknesses, and applications of each algorithm, offering valuable insights into various domains. From classical approaches such as Q-learning to advanced Deep RL (DRL), along with algorithmic variations tailored to specific domains, the paper provided a comprehensive overview. Besides classifying RL algorithms according to Model-free/based approaches, scalability, and sample efficiency, it also provided a practical guide for researchers and practitioners about the type(s) of algorithms used in various domains. Furthermore, this survey examines the practical implementation and performance of RL algorithms across several fields, including Games, Robotics, Autonomous systems, and many more. It also provided a balanced assessment of their usefulness. 



\bibliographystyle{IEEEtran}   
\bibliography{references}      

\end{document}